%% file: main.tex
\documentclass[10pt]{article} 
\usepackage[accepted]{tmlr}

\input{math_commands.tex}

\usepackage{url}

\usepackage{algorithm}
\usepackage[noend]{algorithmic}

\let\classAND\AND
\let\AND\relax
\usepackage{algorithmic}

\let\AND\classAND
\AtBeginEnvironment{algorithmic}{\let\AND\algoAND}

\floatname{algorithm}{Algorithm}
\renewcommand{\algorithmicrequire}{\textbf{Input:}}
\renewcommand{\algorithmicensure}{\textbf{Output:}}
\renewcommand{\algorithmicendif}{\textbf{end}}

\usepackage{amsmath}
\usepackage{amssymb,mathrsfs}
\usepackage{mathtools}
\usepackage{amsthm}
\usepackage{xargs}
\usepackage{bm}
\usepackage{bbm}
\usepackage{float}
\usepackage{siunitx}
\usepackage{adjustbox}

\theoremstyle{plain}
\newtheorem{theorem}{Theorem}[section]
\newtheorem{proposition}[theorem]{Proposition}
\newtheorem{lemma}[theorem]{Lemma}

\theoremstyle{definition}
\newtheorem{definition}[theorem]{Definition}
\theoremstyle{remark}

\usepackage[utf8]{inputenc}   
\usepackage[T1]{fontenc}      
\usepackage{microtype}
\usepackage{graphicx}
\usepackage{subcaption}
\usepackage{wrapfig}
\usepackage{booktabs} 

\usepackage{hyperref}
\hypersetup{
    colorlinks=true,
    linkcolor=blue,
    filecolor=blue,      
    urlcolor=blue,
    citecolor=blue,
}
\definecolor{RedOrange}{HTML}{F26035}

\usepackage[commands,oxfordcolors]{MJH}
\usepackage{layouts}

\usepackage[inline]{enumitem}

\usepackage{comment}
\usepackage{amssymb,mathrsfs}
\usepackage{amsfonts}

\usepackage{nccmath}
\usepackage{upgreek}
\usepackage{xspace}

\usepackage{graphicx}
\usepackage{color}
\usepackage{bbm}
\usepackage[textwidth=1.8cm]{todonotes}

\usepackage{aliascnt}
\usepackage[capitalise,noabbrev]{cleveref}
\usepackage[toc,page,header]{appendix}
\usepackage{multirow}
\hypersetup{
  colorlinks,
  linkcolor={red!50!black},
  citecolor={blue!50!black},
  urlcolor={blue!50!black}
}

\newcommand{\appendixhead}{
  \centerline{\textbf{\LARGE Supplementary to: }\vspace{0.15in}}
  \centerline{\textbf{\LARGE Diffusion Models for Constrained Domains}\vspace{0.25in}}
}

\input{preamble/def}
\DeclareMathOperator{\intersect}{intersect}
\DeclareMathOperator{\pt}{parallel transport}
\DeclareMathOperator{\reflect}{reflect}

\title{Diffusion Models for Constrained Domains}

\author{\name Nic Fishman \email njwfish@gmail.com\\
    \addr Department of Statistics, University of Oxford
    \AND
    \name Leo Klarner \email leo.klarner@stats.ox.ac.uk\\
    \addr Department of Statistics, University of Oxford
    \AND
    \name Valentin De Bortoli \email valentin.debortoli@gmail.com\\
    \addr CNRS, ENS Ulm
    \AND
    \name Emile Mathieu \email ebm32@cam.ac.uk\\
    \addr Department of Engineering, University of Cambridge
    \AND
    \name Michael Hutchinson \email michael.hutchinson@stats.ox.ac.uk\\
    \addr Department of Statistics, University of Oxford
}

\def\month{07}  
\def\year{2023} 
\def\openreview{\url{https://openreview.net/forum?id=xuWTFQ4VGO}} 

\newtheorem{assumption}{\textbf{A}\hspace{-3pt}}
\crefformat{assumption}{{\textbf{A}}#2#1#3}
\def \Pker {\mathrm{P}}

\begin{document}

\maketitle

\begin{abstract}
Denoising diffusion models are a novel class of generative algorithms that achieve state-of-the-art performance across a range of domains, including image generation and text-to-image tasks. Building on this success, diffusion models have recently been extended to the Riemannian manifold setting, broadening their applicability to a range of problems from the natural and engineering sciences.
However, these Riemannian diffusion models are built on the assumption that their forward and backward processes are well-defined for all times, preventing them from being applied to an important set of tasks that consider manifolds defined via a set of inequality constraints.
%
%
In this work, we introduce a principled framework to bridge this gap. 
We present two distinct noising processes based on
\begin{enumerate*}[label=(\roman*)]
    \item the \emph{logarithmic barrier} metric and
    \item the \emph{reflected} Brownian motion
\end{enumerate*} induced by the constraints. As existing diffusion model techniques cannot be applied in this setting, we derive new tools to define such models in our framework.
We then demonstrate the practical utility of our methods on a number of synthetic and real-world tasks, including applications from robotics and protein design.
\end{abstract}

\input{sections/intro}
\input{sections/background}

\input{sections/method}

\input{sections/related_work}
\input{sections/experiments}
\input{sections/conclusions}


\bibliography{bibliography.bib}
\bibliographystyle{tmlr}

\newpage
\onecolumn
\appendix
\appendixhead
\input{appendices/intro_appendix.tex}
\input{appendices/manifold_concepts.tex}
\vfill
\pagebreak
\input{appendices/brownian.tex}

\vfill
\pagebreak
\input{appendices/log_barrier.tex}
\vfill
\pagebreak
\input{appendices/reflected_bm.tex}
\vfill
\pagebreak
\input{appendices/proof_ism.tex}
\vfill
\pagebreak
\input{appendices/likelihood.tex}
\vfill
\pagebreak
\input{appendices/proof_time_reversal.tex}
\vfill
\pagebreak
\input{appendices/application_psd.tex}

\input{appendices/application_loop.tex}
\input{appendices/exp_details}
\end{document}

%% file: math_commands.tex
\usepackage{amsmath,amsfonts,bm}

\def\eqref#1{equation~\ref{#1}}

\def\1{\bm{1}}

\def\rmA{{\mathbf{A}}}

\def\rmL{{\mathbf{L}}}
\def\rmM{{\mathbf{M}}}

\DeclareMathAlphabet{\mathsfit}{\encodingdefault}{\sfdefault}{m}{sl}
\SetMathAlphabet{\mathsfit}{bold}{\encodingdefault}{\sfdefault}{bx}{n}

\newcommand{\E}{\mathbb{E}}

\newcommand{\R}{\mathbb{R}}

\DeclareMathOperator*{\argmin}{arg\,min}

\newcommand{\pms}[1]{\ensuremath{{\scriptstyle\pm #1}}}

%% file: preamble/def.tex
 \newcommandx{\normLigne}[2][1=]{\ifthenelse{\equal{#1}{}}{\Vert #2 \Vert}{\Vert #2\Vert^{#1}}}

\def\bfn{\mathbf{n}}
\def\bfk{\mathbf{k}}
\def\bfj{\mathbf{j}}

\def\bfY{\mathbf{Y}}

\def\bfX{\mathbf{X}}

\def\bfB{\mathbf{B}}

\def\msa{\mathsf{A}}

\def\msu{\mathsf{U}}

\newcommand{\mcb}[1]{\mathcal{B}(#1)}

\def\rset{\mathbb{R}}

\def\nset{\mathbb{N}}

\def\rmA{\mathrm{A}}
\def\rmM{\mathrm{M}}
\def\rmL{\mathrm{L}}

\def\rmd{\mathrm{d}}

\def\rmc{\mathrm{C}}

\newcommand{\M}{\mathcal M}
 \newcommand{\PE}{\mathbb{E}}
 \newcommand{\expeLigne}[1]{\PE [ #1 ]}

\def\eqsp{\;}
\newcommand{\coint}[1]{\left[#1\right)}
\newcommand{\ocint}[1]{\left(#1\right]}
\newcommand{\ooint}[1]{\left(#1\right)}
\newcommand{\ccint}[1]{\left[#1\right]}

\newcommand{\metric}{\mathfrak{g}}
\newcommand{\metricN}{\mathfrak{h}}

\def\Id{\operatorname{Id}}

 \newcommandx{\CPELigne}[3][1=]{{\mathbb E}_{#1}[#2  \vert #3  ]} 
 \newcommand{\ensembleLigne}[2]{\{#1\,:\eqsp #2\}}

%% file: sections/intro.tex
\section{Introduction}
Diffusion models
\citep{sohl2015deep,song2019generative,song2020score,ho2020denoising} have
recently been introduced as a powerful new paradigm for generative modelling.
They work as follows:
noise is progressively added to data
following a Stochastic Differential Equation (SDE)---the forward \emph{noising}
process---until it is approximately Gaussian. The generative model is given by
an approximation of the associated \emph{time-reversed} process called the backward
\emph{denoising} process. This is also an SDE whose drift depends on the
gradient of the logarithmic densities of the forward process, referred to as the Stein score.
This score is approximated by leveraging techniques from deep learning and score matching
\citep{hyvarinen2005estimation,vincent2011connection}.
Building on the success of diffusion models in domains such as images and text, this framework has recently been extended to a wide range of Riemannian manifolds \citep{debortoli2022riemannian,huang2022Riemannian}, broadening their applicability to various important modelling domains from the natural and engineering sciences---including Lie groups such as the group of rotations $\mathrm{SO}(3)$, the group of rigid body motions $\mathrm{SE}(3)$, and many others \citep[see e.g.][]{trippe2022Diffusion,corso2022DiffDock, watson2022Broadly,leach2022denoising,urain2022se,yim2023se}.

However, a key assumption of the Riemannian diffusion models introduced in \citet{debortoli2022riemannian} and \citet{huang2022Riemannian} is that the stochastic processes they consider are defined \emph{for all times}. 
While this holds for a large class of stochastic processes, it is not the case for most manifolds defined via a set of inequality constraints. For instance, in the case of the hypercube $\ooint{-1,1}^d$ equipped with the Euclidean metric, the Riemannian Brownian motion coincides with the Euclidean $d$-dimensional Brownian motion $(\bfB_t)_{t \in \ccint{0,T}}$ as long as $\bfB_t \in \ooint{-1,1}$. With probability one, $(\bfB_t)_{t \geq 0}$ escapes from $\ooint{-1,1}$, meaning that the Riemannian Brownian motion is not defined for all times and the frameworks introduced in \citet{debortoli2022riemannian} and \citet{huang2022Riemannian} do not apply.
Such constrained manifolds comprise a wide variety of settings---including polytopes and convex sets of Euclidean spaces---and are studied across a large number of disciplines, ranging from computational statistics \citep{Morris2002}, over robotics \citep{han2006inverse} and quantum physics \citep{lukens2020practical}, to computational biology \citep{Thiele2013}. Deriving principled diffusion models that are able to operate directly on these manifolds is thus of significant practical importance, as they enable generative modelling in data-scarce and safety-critical settings in which constraints on the modelled domain may reduce the number of degrees of freedom or prevent unwanted behaviour. 





As sampling problems on such manifolds are important \citep{kook2022sampling,heirendt2019creation}, 
a flurry of Markov chain based methods have been developed to sample from unnormalised densities.
Successful algorithms include the reflected Brownian motion
\citep{williams1987Reflected,petit1997Time,shkolnikov2013Timereversal},
log-barrier methods
\citep{Kannan2009,lee2017Geodesic,noble2022Barrier,kook2022sampling,gatmiry2022convergence,lee2018convergence}
and hit-and-run approaches in the case of polytopes
\citep{Smith1984,Lovasz2006}.
In this work, we study the generative modelling counterparts of these algorithms
through the lens of diffusion models. Among existing methods for statistical
sampling on constrained manifolds, the geodesic Brownian motion
\citep{lee2017Geodesic} and the reflected Brownian motion
\citep{williams1987Reflected} are continuous stochastic processes, and thus well suited for extending the continuous Riemannian diffusion framework developed by \citet{debortoli2022riemannian} and  \citet{huang2022Riemannian}. In particular, we introduce two principled diffusion models for generative modelling on constrained domains based on \begin{enumerate*}[label=(\roman*)]
    \item the geodesic Brownian motion, leveraging tools from the log-barrier methods, and 
    \item the reflected Brownian motion
\end{enumerate*}. In both cases, we show how one can extend the ideas of
time-reversal and score matching to these settings.
We demonstrate the practical utility of these methods on a range of tasks defined on convex polytopes and the space of symmetric positive definite matrices, including the constrained conformational modelling of proteins and robotic arms.
The code for all of our experiments is available \href{https://github.com/oxcsml/constrained-diffusion}{here}. 

\begin{figure}[t]
    \centering
    \includegraphics[width=\textwidth]{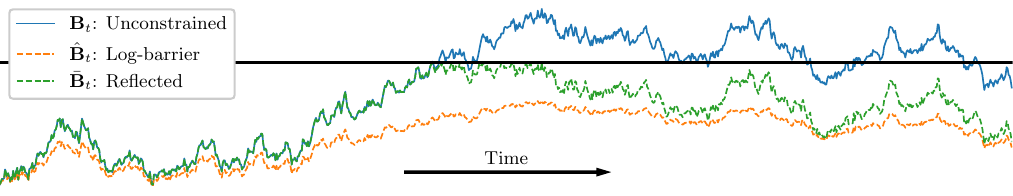}
    \caption{The behaviour of different types of noising processes considered in this work defined on the unit interval. \(\mathbf{B}_t\): Unconstrained (Euclidean) Brownian motion.
            \(\hat{\mathbf{B}}_t\): Log-barrier forward noising process. \(\bar{\mathbf{B}}_t\): Reflected Brownian motion. All sampled with the same initial point and driving noise. Black line indicates the boundary.
        }
    \label{fig:reflected_vs_brownian}
\end{figure}

%% file: sections/background.tex
\section{Background}
\paragraph{Riemannian manifolds.} A Riemannian manifold is a tuple
$(\M, \metric)$ with $\M$ a smooth manifold and $\metric$ a metric which defines
an inner product on tangent spaces.  The metric $\metric$ induces key quantities
on the manifold, such as an exponential map
$\exp_x: \mathrm{T}_x \M \rightarrow \M$, defining the notion of following
straight lines on manifolds,
a gradient operator $\nabla$\footnote{The (Riemannian) gradient $\nabla$ is
  defined s.t.\ for any smooth $f\in \rmc^\infty(\M)$,
  $x \in \M, v \in \mathrm{T}_x\M$, $\metric(x)(f(x), v) = \rmd f(x) (v)$.} and
a divergence operator $\mathrm{div}$\footnote{The Riemannian divergence acts
  vector fields and can be defined using the volume form of $\M$.}.  It also
induces the Laplace-Beltrami operator $\Delta$ and consequently a Brownian
motion with density (w.r.t.\ the volume form\footnote{We assume that $\M$ is
  orientable and therefore that a volume form exists.}), whose density is given
by the heat equation ${\partial_t } p_t = \Delta p_t$.  We refer the reader to
\cref{sec:riemannain_intro} for a brief introduction to differential geometry,
to \citet{lee2013smooth} for a thorough treatment and to
\citet{hsu2002stochastic} for details on stochastic analysis on manifolds.
\begin{figure}
    \centering
    \begin{minipage}{0.5\textwidth}
        \includegraphics[width=\textwidth]{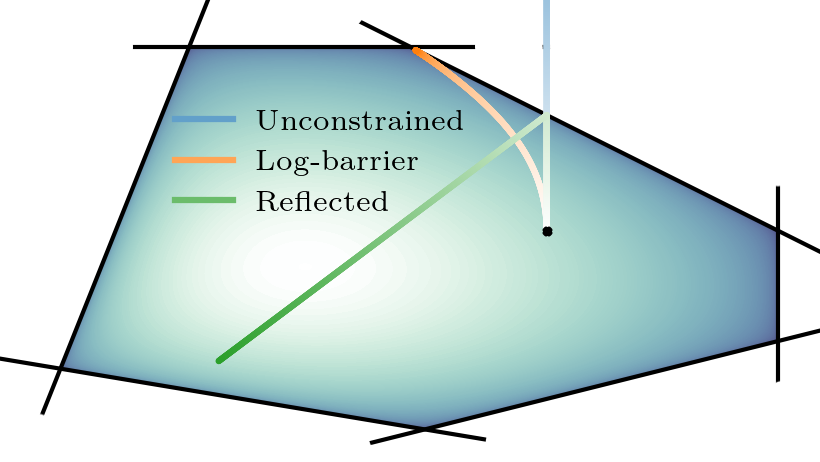}
    \end{minipage}
    \hfill
    \begin{minipage}{0.45\textwidth}
        \caption{
            A convex polytope defined by six constraints $\{f_i\}_{i \in \c{I}}$, along with the log barrier potential, and `straight trajectories' under the log-barrier metric and under the Euclidean metric with and without   reflection at the boundary.
        }
        \label{fig:polytope}
    \end{minipage}
\end{figure}

\input{sections/manifolds_with_boundary}


%% file: sections/manifolds_with_boundary.tex
\paragraph{Constrained manifolds.}
\label{sec:manifold_w_boundary}

%
In this work, we are concerned with \emph{constrained} manifolds. More
precisely, given a Riemannian manifold $(\c{N}, \metricN)$,
we consider a family of real functions 
$\{f_i: \c{N} \to \R\}_{i \in \mathcal{I}}$ indexed by $\mathcal{I}$. We then define
\begin{equation}
  \label{eq:constrained}
  \c{M} = \ensembleLigne{x \in \c{N}}{f_i(x) < 0 , \ i \in \mathcal{I}}.
  \end{equation}
  In this scenario, $\{f_i\}_{i \in \mathcal{I}}$ is interpreted as a set of constraints on \(\c{N}\).
  For example, choosing $\c{N} = \rset^d$ and affine constraints
  $f_i(x) = \langle a_i, x\rangle - b_i$, $x\in \rset^d$, we get that $\c{M}$ is
  an open polytope as illustrated in \cref{fig:polytope}. This setting naturally
  appears in many areas of engineering, biology, and physics
  \citep{boyd2004convex, han2006inverse, lukens2020practical}.

%

  
  While the two methods we introduce in \Cref{sec:constrained_diffusion_models}
  can be applied to the general framework (\ref{eq:constrained})
  , in our
  applications, we focus on two specific settings:
  \begin{enumerate*}[label=(\alph*)]
    \item Polytopes---$\c{N} = \rset^d$ and  
      \item symmetric positive-definite (SPD)
        matrices under trace constraints---$\c{N}=\mathcal{S}_{++}^{d}$.
   \end{enumerate*}
  

\paragraph{Continuous diffusion models.}
\label{sec:sgm}

We briefly recall the framework for constructing continuous diffusion processes
introduced by \citet{song2020score} in the context of generative modelling over
$\rset^d$. At minimum diffusion models need four things:
\begin{enumerate*}[label=(\roman*)]
    \item A forward noising process converging to an invariant distribution.
    \item A time reversal for the reverse process.
    \item A discretization of the continuous-time process for the forward/reverse process.
    \item A score matching loss --- in this paper we will focus on the implicit score matching loss.
\end{enumerate*}
\citet{song2020score} consider a forward \textit{noising process}
$(\bfX_{t})_{t \in \ccint{0,T}}$ which progressively noises a data distribution
$p_0$ into a Gaussian $\mathrm{N}(0,\Id)$.
More precisely $(\bfX_{t})_{t \in \ccint{0,T}}$ is an Ornstein–Uhlenbeck (OU)
process which is given  by the following stochastic differential equation (SDE)
\begin{equation}
  \label{eq:ornstein_uhlenbeck}
 \rmd \bfX_t = - \tfrac{1}{2}\bfX_t \rmd t + \rmd \bfB_t, \qquad \bfX_0 \sim p_0.
\end{equation}
Under mild conditions on $p_0$, the time-reversed process $(\overleftarrow{\bfX}_t)_{t \in \ccint{0,T}} = (\bfX_{T-t})_{t \in \ccint{0,T}}$ also satisﬁes an SDE \citep{cattiaux2021time,haussmann1986time} given by
\begin{equation}
  \label{eq:time_reversal}
\rmd \overleftarrow{\bfX}_t = \{\tfrac{1}{2}\overleftarrow{\bfX}_t + \nabla
\log p_{T-t}(\overleftarrow{\bfX}_t)\} \rmd t + \rmd \bfB_t, \qquad \overleftarrow{\bfX}_0 \sim p_T, 
\end{equation}
where $p_t$ denotes the density of $\bfX_t$.  This construction allows direct
sampling of the forward process and leverages the Euler-Maruyama
discretisation to facilitate sampling of the reverse process.
Finally, the quantity $\nabla \log p_t $ is referred as the
Stein score and is unavailable in practice.  It can be approximated with a score
network $s_\theta(t,\cdot)$ trained by minimising 
a denoising score matching ($\mathrm{dsm}$) loss
\begin{equation}
\label{eq:denoising_sm}
\textstyle 
\mathcal{L}(\theta) = \E [ \lambda_t \norm{ \nabla \log p_{t|0}(\bfX_t|\bfX_0) - 
  s_\theta(t, \bfX_t) }^2],
\end{equation}
or an equivalent implicit score matching ($\mathrm{ism}$) loss
\begin{equation}
\label{eq:implicit_sm}
\textstyle{
\mathcal{L}(\theta) = \E[  \lambda_t \{ \tfrac{1}{2} \norm{
  s_\theta(t, \bfX_t) }^2 + \mathrm{div}(s_\theta)(t, \bfX_t)}\}] + C,
\end{equation}
where $C \geq 0$ and $\lambda_t >0$ is a weighting function, and the expectation
is taken over $t \sim \mathcal{U}([0, T])$ and $(\bfX_0, \bfX_t)$.  For an
arbitrarily flexible score network, the global minimiser
$\theta^\star = \mathrm{argmin}_{\theta}\mathcal{L}(\theta)$ satisfies
$s_{\theta^\star}(t, \cdot)=\nabla \log p_t$.



%% file: sections/method.tex
\section{Inequality-Constrained Diffusion Models}
\label{sec:constrained_diffusion_models}

We are now ready to introduce our methodology to deal with manifolds defined via \emph{inequality constraints} (\ref{eq:constrained}). 
In \Cref{sec:bsgm}, we propose a Riemannian diffusion model endowed with a metric induced by a log-barrier potential.
In \Cref{sec:rbm}, we introduce a \emph{reflected} diffusion model.  While both
models extend classical diffusion models to inequality-constrained settings,
they exhibit very different behaviours. 
We discuss their practical differences in \Cref{sec:diff_methods}.

\input{sections/barrier_method}
 \input{sections/reflected_method}

\subsection{Loss and score network parameterisation}\label{sec:score-netw-param}

In order to train log-barrier and reflected diffusion models we prove that we can use a \emph{tractable} score matching loss in constrained manifolds. We will prove that the implicit score-matching loss leads to the recovery of the correct score when we have a boundary, so long as we enforce that the score is zero on the boundary (see~\Cref{app:proof_for_ism}). This proof holds for both the log-barrier and the reflected process.

\begin{proposition}
  Let $s \in \rmc^\infty(\ccint{0,T} \times \rset^d, \rset^d)$ such
  that for any $x \in \partial \M$ and $t \geq 0$, $s_t(x) =0$. Then, there
  exists $C >0$ such that
  \begin{equation}
    \textstyle{\expeLigne{\normLigne{\nabla \log p_t - s_t}^2} = \expeLigne{\normLigne{s_t}^2 + 2~ \mathrm{div}(s_t)}} + C ,
  \end{equation}
  where $\mathbb{E}$ is taken over $\bfX_t \sim p_t$ and
  $t \sim \mathcal{U}([0,T])$.
\end{proposition}



This result immediately implies we can optimise the score network using the ism
loss function so long as we enforce a Neumann boundary condition. The estimation
of the score term $\nabla \log p_t$ is also done by minimising the
$\mathrm{ism}$ loss function (\ref{eq:implicit_sm})
.

The score term \(\nabla \log p_{T-t}\) appearing in both time-reversal processes 
(\ref{eq:reversal_langevin_hessian_manifold}) and (\ref{eq:evolution_Y})
is intractable.
  It is thus approximated with a \emph{score network} $s_\theta(t, \cdot) \approx \nabla \log p_t$.
 We use a multi-layer perceptron architecture with $\sin$ activation functions, see \cref{sec:exp_details} for more details on the experimental setup.
 Due to boundary condition (\ref{eq:heat_eq_rbm})
 , the normal component of the score is zero at the boundary in the reflected case. A similar result holds in the log-barrier setting. This is additionally required for our proof of the ism loss.
 
 Following \cite{liu2022estimating}, we can accommodate this in our score parameterisation by additionally scaling the score output of the neural network by a monotone function $h(d(x, \partial \M))$ where $d$ is the distance from $x$ to the boundary and $h(0) = 0$. In particular, we use a clipped ReLU function:
 $s_\theta(t, x) = \min(1, \textrm{ReLU}(d(x, \partial \M) - \delta)) \cdot \textrm{NN}_\theta(t, x)$ with $\delta>0$ a threshold so the model is forced to be zero ``close'' to the boundary as well as exactly on the boundary, see \Cref{sec:exp_details} for an illustration. The inclusion of this scaling function is necessary to produce reasonable results as we show in \Cref{sec:boundary_score}. The weighting function in (\ref{eq:implicit_sm}) is set to $\lambda_t = (t + 1)$.

%

The forward processes (\ref{eq:langevin_hessian_manifold}) and (\ref{eq:rbm}) 
for the barrier and reflected methods cannot be sampled in closed form, so at training time samples from the conditional marginals $p(\bfX_t|\bfX_0)$ are obtained by discretising these processes.
As to take the most of this computational overhead, we use several samples from the discretised forward trajectory $(\bfX_{t_1}, \cdots, \bfX_{t_k}|\bfX_0)$ instead of only using the last sample. 


%% file: sections/barrier_method.tex
\subsection{Log-barrier diffusion models}
\label{sec:bsgm}

\begin{figure}[h]
    \begin{minipage}[b]{0.49\linewidth}
        \centering
        \includegraphics{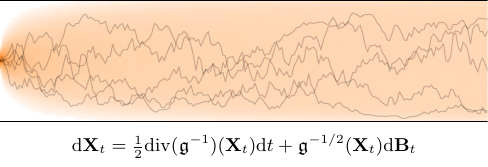}
        \caption{Convergence of the Barrier Langevin dynamics on the unit interval to the uniform distribution.}
        \label{fig:density}
    \end{minipage}
    \hfill
    \begin{minipage}[b]{0.49\linewidth}
      \centering
        \begin{tikzpicture}
        \tikzset{
            punkt/.style={
                   rectangle,
                   rounded corners,
                   draw=black, very thick,
                   text width=9em,
                   minimum height=2em,
                   text centered},
            pil/.style={
                   <->,
                   thick,
                   shorten <=2pt,
                   shorten >=2pt,}
        }
          \node[punkt, text width=6em] (a) at (0,0) {$(\mathcal{N}, \metricN )$};
          \node[punkt, text width=6em, right=5em of a] (b) {$(\mathcal{M}, \nabla^2 \phi)$}
            edge[pil] (a.east);
        \node[draw, above=1em of a] (e) {Original space};
        \node[draw, above=1em of b] (f) {Constrained space};
        \end{tikzpicture}
        \vspace{2.7em}
        \caption{
        Illustrative diagram of the barrier method and the change of metric.
        }
    \end{minipage}
\end{figure}

\paragraph{Barrier Langevin dynamics.}
Barrier methods work by constructing a smooth potential $\phi: \M \rightarrow
\rset$ such that it blows up on the boundary of a desired set, see
\citet{nesterov2018lectures}. Such potentials form the basis of interior point
methods in convex optimisation \citep{boyd2004convex}. Of these functions, the \emph{logarithmic
  barrier} is the most popular among practitioners ~\citep{lee2017Geodesic}.
For a convex polytope $\M$ defined by the constraints $\rmA x < b$, with $\rmA
\in \rset^{m \times d}$ and $b \in \rset^m$, the logarithmic barrier $\phi: \ \M
\to \rset_+$ is given for any $x \in \M$ by
\begin{equation}
  \label{eq:log_barrier_polytope}
  \textstyle{
    \phi(x) = - \sum_{i=1}^m \log(\langle \rmA_i, x\rangle-b_i).
    }
  \end{equation}
  Assuming that $\| \rmA_i \| =1$, we have that for any $x \in \M$, $\phi(x) = -
  \sum_{i=1}^m \log(d(x, \partial \M_i))$, where $\partial \M_i =
  \ensembleLigne{x \in \rset^d}{\langle \rmA_i, x\rangle =
    b}$.  More generally we can define for any $x \in \M$
\begin{equation}
  \label{eq:log_barrier_manifold}
  \textstyle{
    \phi(x) = - \sum_{i=1}^m \log(d(x, \partial \M_i))
    }
  \end{equation}

  where $d(x, \partial \M_i)$ computes the minimum geodesic distance from $x$ to the boundary $\partial \M$. In general this is a highly non-trivial optimization problem, contrary to the polytope case which admits a simple closed form.
  
%
%
  While developed and most commonly used in optimisation, barrier methods can
  also be used for sampling \citep{lee2017Geodesic}.  The core idea of barrier
  methods is to `warp the geometry' of the constrained space, stretching it
  as the process approaches the boundary so that it never hits it, hence bypassing the
  need to explicitly deal with the boundary. 
%
  Assuming $\phi$ to be strictly convex and smooth, its Hessian $\nabla^2
  \phi$ is positive definite and thus defines a valid Riemannian metric on
  $\M$.  The formal approach to 'warping the geometry' of the convex space with
  the boundary is to endow $\M$ with the Hessian as a Riemannian metric $\metric
  = \nabla^2
  \phi$, making it into a Hessian Manifold, see \cite{shima1997geometry}.
  In the special case where the barrier is given by
  (\ref{eq:log_barrier_polytope})
  , we get for any $x \in \M$
\begin{equation}
    \metric(x) = \rmA^\top S^{-2}(x) \rmA \text{ with } S(x) = \text{diag}(b_i - \langle \rmA_i, x \rangle)_i.
\end{equation}
%

Equipped with this Riemannian metric, we consider the following Langevin
dynamics as a forward process
\begin{equation}
  \label{eq:langevin_hessian_manifold}
  \textstyle{
    \rmd \bfX_t = \tfrac{1}{2} \mathrm{div}(\metric^{-1})(\bfX_t)  \rmd t + \metric(\bfX_t)^{-\frac{1}{2}} \rmd \bfB_t,
    }
\end{equation}
with $\mathrm{div}(F)(x) \triangleq \left(\mathrm{div}(F_1)(x), \dots,
  \mathrm{div}(F_d)(x) \right)^\top $, for any smooth $F: \ \rset^d \to
\rset^d$. Under mild assumptions on $\M$,
$\mathrm{div}(\metric^{-1})$ and $\metric^{-1}$ we get that $(\bfX_t)_{t \geq
  0}$ is well-defined and for any $t \geq 0$, $\bfX_t \in
\M$. In particular, for any $t \geq 0$, 
$\bfX_t$ does not reach the boundary.  This stochastic process was first
proposed by \cite{lee2017Geodesic} in the context of efficient sampling from the
uniform distribution over a polytope.  Under similar conditions,
$\bfX_t$ admits a density
$p_t$ w.r.t.\ the Lebesgue measure and we have that $\partial_t p_t =
\tfrac{1}{2} \mathrm{Tr}(\metric^{-1} \nabla^2p_t)$. In addition, $(\bfX_t)_{t
  \geq 0}$ is irreducible. Hence, assuming that
$\M$ is compact, the uniform distribution on
$\M$ is the unique invariant measure of the process $(\bfX_t)_{t \geq
  0}$ and $(\bfX_t)_{t \geq
  0}$ converges to the uniform distribution in some sense. We refer the reader
to \Cref{sec:geod-brown-moti} for a proof of these results.



\paragraph{Time-reversal.}
Assuming that $\metric^{-1}$ and its derivative are bounded on $\M$, the
time-reversal of (\ref{eq:langevin_hessian_manifold}) 
is given by
\citet{cattiaux2021time}, in particular we have
\begin{align}
  \rmd \overleftarrow{\bfX}_t &= \textstyle{[-\frac{1}{2} \mathrm{div}(\metric^{-1}) + \mathrm{div}(\metric^{-1})}   \textstyle{ + \metric^{-1} \nabla \log p_{T-t} ](\overleftarrow{\bfX_t}) \rmd t + \metric(\overleftarrow{\bfX}_t)^{-\frac{1}{2}} \rmd \bfB_t,}    \\
                               &= \textstyle{ \left[\frac{1}{2} \mathrm{div}( \metric^{-1}) + \metric^{-1} \nabla \log p_{T-t}\right](\overleftarrow{\bfX}_t) \rmd t  + \metric(\overleftarrow{\bfX}_t)^{-\frac{1}{2}} \rmd \bfB_t.}  \label{eq:reversal_langevin_hessian_manifold}
\end{align}
%
%
$\overleftarrow{\bfX}_0$ is initialised with the uniform distribution on $\M$ (which is close to the one of $\bfX_T$ for large $T$). 

The estimation of the score
term $\nabla \log p_t$ is done by minimising the $\mathrm{ism}$ loss function
(\ref{eq:implicit_sm})
. We refer to \Cref{sec:score-netw-param} for details on
the training and parameterisation.

\paragraph{Sampling.}
Sampling from the forward 
(\ref{eq:langevin_hessian_manifold}) and backward (\ref{eq:reversal_langevin_hessian_manifold}) processes, once the score is
learnt, requires a discretisation scheme.
We use Geodesic Random Walks (GRW) \citep{jorgensen1975central} for this purpose,
see \Cref{alg:grw}.  This discretisation is a generalisation of the
Euler-Maruyama discretisation of SDE in Euclidean spaces, where the $+$ operator
is replaced by the exponential mapping on the manifold, computing the geodesics.

\begin{algorithm}[]
  \caption{\emph{Geodesic Random Walk.} Discretisation of the SDE
    $\rmd \bfX_t = d(t, \bfX_t) \rmd t +  \rmd \bfB_t$.}
   \label{alg:grw}
\begin{algorithmic}
    \STATE {\bfseries Require:} $T$ (simulation time), $N$ (number of steps), $X_0^\gamma$ (initial point), $d$ (drift function)
  \STATE $\gamma = T / N$ 
  \FOR{$k \in \{0, \dots, N-1\}$}
  \STATE $Z_{k+1} \sim \mathrm{N}(0, \Id)$
  \STATE $W_{k+1} = \gamma d(k \gamma, X_k) + \sqrt{\gamma} Z_{k+1}$ 
  \STATE $X_{k+1}^\gamma = \exp_{X_k}[W_{k+1}] \approx \mathrm{proj}_\M(X_k + W_{k+1})$
  \ENDFOR
  \STATE {\bfseries return} $\{ X_k\}_{k=0}^{N}$
\end{algorithmic}
\end{algorithm}



However, contrary to \citet{debortoli2022riemannian}, we will not have access
explicitly to the exponential mapping of the Hessian manifold. Instead, we rely on an approximation,
using a \emph{retraction} (see \citet{absil2012projection, boumal2023introduction} for a definition and alternative schemes). 

%



%% file: sections/reflected_method.tex
\subsection{Reflected diffusion models}
\label{sec:rbm}
\begin{figure}[h]
  \label{fig:rbm}
        \includegraphics{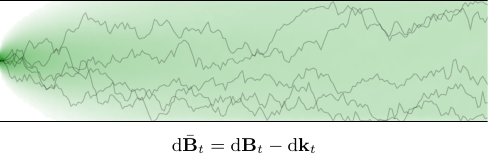}
    \hfill
        \includegraphics{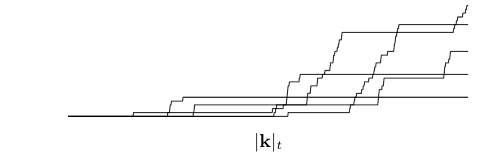}
    \caption{\emph{Left:} Convergence of the reflected Brownian motion on the unit interval to the uniform distribution. \emph{Right:} Value of \(\abs{\bfk}_t\) for the trajectory samples on the left through time. 
    }
\end{figure}

Another approach to deal with the geometry of $\M$ is to use the standard metric $\mathfrak{h}$ and forward dynamics of $\c{N}$ and constraining it to $\M$ by \emph{reflecting} the process whenever it would encounter a boundary.
We will first assume that $\M$ is compact and convex.
To simplify the presentation, we focus on the Euclidean case $\c{N} = \rset^d$ with smooth boundary $\partial M$\footnote{We refer to \Cref{sec:time-revers-refl} for a definition of smooth boundary.}.

The key difference between this approach and the barrier approach is that in the
reflected case we leave the geometry unchanged, so all we need to do is show
that the dynamics induced by reflecting the forward process whenever it hits the
boundary leads to an invariant distribution and admits a time-reversal.

It is worth noting that while barrier approaches have received considerable
theoretical attention in the sampling literature \citep{lee2017Geodesic,
  noble2022Barrier}, reflected methods have remained comparatively undeveloped
from the methodological and practical point of view. In the next section, we
recall the basics of reflected stochastic processes.

\paragraph{Skorokhod problem.} The reflected Brownian motion is defined as the
solution to the Skorokhod problem. Roughly speaking a solution to the Skorokhod
problem consists of two coupled processes, $(\bar{\bfB}_t, \bfk_t)_{t \geq 0}$,
such that $(\bar{\bfB}_t)_{t \geq 0}$ acts \emph{locally} as a Euclidean
Brownian motion $(\bfB_t)_{t \geq 0}$ and $\bfk_t$ compensates for the excursion
of $(\bfB_t)_{t \geq 0}$ so that $(\bar{\bfB}_t)_{t \geq 0}$ remains in $\M$. We
say that $(\bar{\bfB}_t, \bfk_t)_{t \geq 0}$ is a solution to the
\emph{Skorokhod problem} \citep{skorokhod1961stochastic} if
$(\bfk_t)_{t \geq 0}$ and $(\bar{\bfB}_t)_{t \geq 0}$ are two processes
satisfying mild conditions, see \Cref{sec:refl-brown-moti} for a rigorous
introduction, such that for any $t \geq 0$,
\begin{equation} \label{eq:rbm}
    \bar{\bfB}_t = \bar{\bfB}_0 + \bfB_t - \bfk_t \in \M,
\end{equation}
and
$\textstyle{\abs{\bfk}_t = \int_0^t \mathbf{1}_{\bar{\bfB}_s \in \partial \M}
  \rmd \abs{\bfk}_s}$,
$\textstyle{\bfk_t = \int_0^t \bfn(\bar{\bfB}_s) \rmd \abs{\bfk}_s ,}$ where
$(\abs{\bfk}_t)_{t \geq 0}$ is the total variation of $(\bfk_t)_{t \geq 0}$ and
we recall that $\bfn$ is the outward normal to $\M$\footnote{We extend the
  normal $\bfn$ to the whole space by letting $\bfn(x)=0$ if
  $x \not \in \partial \M$.}. When
$(\bar{\bfB}_t)_{t \geq 0}$ hits the boundary, the condition
$\textstyle{\bfk_t = \int_0^t \bfn(\bar{\bfB}_s) \rmd \abs{\bfk}_s ,}$ tells us
that $-\bfk_t$ ``compensates'' for $\bar{\bfB}_t$ by pushing the process back
into $\M$ along the inward normal $-\bfn$, while the condition
$\textstyle{\abs{\bfk}_t = \int_0^t \mathbf{1}_{\bar{\bfB}_s \in \partial \M}
  \rmd \abs{\bfk}_s}$ can be interpreted as $\bfk_t$ being constant when
$(\bar{\bfB}_t)_{t \geq 0}$ does not hit the boundary.
As a result $(\bar{\bfB}_t)_{t \geq 0}$ can be understood as the continuous-time
counterpart to the reflected Gaussian random walk. The process
$(\bfk_t)_{t \geq 0}$ can be related to the notion of \emph{local time}
\citep{revuz2013continuous} and quantifies the amount of time
$(\bar{\bfB}_t)_{t \geq 0}$ spends at the boundary $\partial \M$. \citet[Theorem
2.1]{lions1984stochastic} ensure the existence and uniqueness of a solution to
the Skorokhod problem. One key observation is that the event
$\{ \bar{\bfB}_t \in \partial \M \}$ has probability zero \citep[Section 7,
Lemma 7]{harrison1987brownian}.  As in the \emph{unconstrained} setting, one can
describe the dynamics of the density of $\bar{\bfB}_t$.
\begin{proposition}[\citet{burdzy2004heat}]
  For any $t > 0$, the distribution of $\bar{\bfB}_t$ admits a density w.r.t.\
  the Lebesgue measure denoted $p_t$. In addition, we have for any
  $x \in \mathrm{int}(\M)$ and $x_0 \in \partial \M$
  \begin{equation}
  \label{eq:heat_eq_rbm}
    \partial_t p_t(x) = \tfrac{1}{2} \Delta p_t(x) , \quad \partial_{\bfn} p_t(x_0) = 0 ,
  \end{equation}
  where we recall that $\mathbf{n}$ is the outward normal to $\M$.
\end{proposition}

Note that contrary to the unconstrained setting, the heat equation has
\emph{Neumann} boundary conditions. Similarly to the compact Riemannian setting
\citep{saloff1994precise} it can be shown that the reflected Brownian motion
converges to the uniform distribution on $\M$ exponentially fast
\citep{loper2020uniform,burdzy2006traps}, see \cref{fig:rbm}. Hence,
$(\bar{\bfB}_t)_{t \geq 0}$ is a candidate for a forward noising process in the
context of diffusion models.

\paragraph{Time-reversal.} In order to extend the diffusion model approach to
the reflected setting, we need to derive a \emph{time-reversal} for
$(\bar{\bfB}_t)_{t \in \ccint{0,T}}$. Namely, we need to characterise the
evolution of
$(\overleftarrow{\bfX}_t)_{t \in \ccint{0,T}} = (\bar{\bfB}_{T-t})_{t \in
  \ccint{0,T}}$.  It can be shown that the time-reversal of
$(\bar{\bfB}_t)_{t \in \ccint{0,T}}$ is also the solution to a Skorokhod
problem.

\begin{algorithm}[t]  
    \begin{algorithmic}
        \REQUIRE \(x \in \c{M}\), \(\v{v} \in \mathrm{T}_x\c{M}\), \(\{f_i\}_{i \in \mathcal{I}}\)
        \STATE \(\ell \gets \norm{\v{v}}_\metric\)
        \STATE \(\v{s} \gets \v{v} / \norm{\v{v}}_\metric\)
        \WHILE{\(\ell \geq 0\)}
            \STATE \(d_i = \arg \intersect_t\sbr{\exp_\metric(x, t\v{s}), f_i}\)
            \STATE \(i \gets \argmin_i\; d_i \; s.t. \; d_i > 0\)
            \STATE \(\alpha \gets \min\del{d_i, \ell} \)
            \STATE \(x' \gets \exp_\metric\del{x, \alpha\v{s}}\)
            \STATE \(\v{s} \gets \pt_\metric\del{x, \v{s}, x'} \)
            \STATE \(\v{s} \gets \reflect\del{\v{s}, f_i}\)
            \STATE \(\ell \gets \ell - \alpha\)
            \STATE \(x \gets x'\)
        \ENDWHILE
        \STATE \textbf{return} \(x\)
    \end{algorithmic}
    \caption{\emph{Reflected step algorithm}. The algorithm operates by
      repeatedly taking geodesic steps until one of the constraints is violated, or
      the step is fully taken. Upon hitting the boundary we parallel transport
      the tangent vector to the boundary and then reflect it against the boundary. We
      then start a new geodesic from this point in the new direction. The \({\arg} \intersect_t\) function computes the distance one must travel along a geodesic in direction \(\v{s}\) til constraint \(f_i\) is intersected. For a discussion of \(\pt\), \(\exp_\metric\) and \(\reflect\) please see \cref{sec:riemannain_intro}.}
    \label{alg:reflection}
\end{algorithm}
\begin{algorithm}[H]
  \caption{\emph{Reflected Random Walk}. Discretisation of the SDE
    $\rmd \bfX_t = b(t, \bfX_t) \rmd t + \rmd \bfB_t - \rmd \bfk_t$.}
   \label{alg:rrw}
\begin{algorithmic}
    \STATE {\bfseries Require:} $T$ (simulation time), $N$ (number of steps), $X_0^\gamma$ (initial point), $\{f_i\}_{i \in \mathcal{I}}$ (boundary functions)
  \STATE $\gamma = T / N$
  \FOR{$k \in \{0, \dots, N-1\}$}
  \STATE $Z_{k+1} \sim \mathrm{N}(0, \Id)$
  \STATE $X_{k+1}^\gamma = \text{ReflectedStep}[X_{k}^\gamma, \sqrt{\gamma} Z_{k+1}, \{f_i\}_{i \in \mathcal{I}}]$
  \ENDFOR
  \STATE {\bfseries return} $\{ X_k^\gamma\}_{k=0}^{N}$
\end{algorithmic}
\end{algorithm}

\begin{theorem}
  \label{sec:time-reversal-reflected}
  There exist $(\overleftarrow{\bfk}_t)_{t \geq 0}$ a bounded variation process and
  a Brownian motion $(\bfB_t)_{t \geq0}$ such that 
  \begin{equation}
    \textstyle{
      \overleftarrow{\bfX}_t = \overleftarrow{\bfX}_0 + \bfB_t + \int_0^t \nabla \log p_{T-s}(\overleftarrow{\bfX}_s) \rmd s - \overleftarrow{\bfk}_t .
      }
    \end{equation}
    In addition, for any $t \in \ccint{0,T}$ we have 
\begin{equation}
  \label{eq:skorokhod_reversal}
  \textstyle{\overleftarrow{\abs{\bfk}}_t = \int_0^t \mathbf{1}_{\overleftarrow{\bfX}_s \in \partial \M} \rmd \overleftarrow{\abs{\bfk}}_s , \quad \overleftarrow{\bfk}_t = \int_0^t \bfn(\overleftarrow{\bfX}_s) \rmd \overleftarrow{\abs{\bfk}}_s .}
\end{equation}
\end{theorem}

The proof, see \Cref{sec:time-revers-refl}, follows \citet{petit1997Time} which
provides a time-reversal in the case where $\M$ is the positive orthant. It is
based on an extension of \citet{haussmann1986time} to the reflected setting,
with a careful handling of the boundary conditions. In particular, contrary to
\citet{petit1997Time}, we do not rely on an explicit expression of $p_t$ but
instead use the intrinsic properties of $(\bfk_t)_{t \geq 0}$.  Informally,
\Cref{sec:time-reversal-reflected} means that the process
$(\overleftarrow{\bfX}_t)_{t \in \ccint{0,T}}$ satisfies
\begin{equation}
  \label{eq:evolution_Y}
  \rmd \overleftarrow{\bfX}_t = \nabla \log p_{T-t}(\overleftarrow{\bfX}_t) \rmd t + \rmd \bfB_t - \rmd \overleftarrow{\bfk}_t , 
\end{equation}
which echoes the usual time-reversal formula (\ref{eq:time_reversal}). 
In practice, in order to sample from
$(\overleftarrow{\bfX}_t)_{t \in \ccint{0,T}}$, one needs to consider the
\emph{reflected} version of the \emph{unconstrained} dynamics
$\rmd \overleftarrow{\bfX}_t = \nabla \log p_{T-t}(\overleftarrow{\bfX}_t) \rmd
t + \rmd \bfB_t$.  

\paragraph{Sampling.}
In practice, we approximately sample the reflected dynamics by considering
the Markov chain given by \Cref{alg:rrw}. We refer to
\citet{pacchiarotti1998numerical} and \citet{bossy2004symmetrized} for weak convergence
results on this numerical scheme in the Euclidean setting.

\paragraph{Likelihood evaluation.} In the case of a reflected Brownian motion,
it is possible to compute an equivalent ODE in order to perform likelihood
evaluation. The associated ODE was first derived in \cite{lou2023reflected}. The
form of the ODE and the proof that it remains in $\M$ are postponed to
\Cref{sec:likel-eval}.

\begin{figure}[t]
    \centering
    \begin{minipage}[0.25\textwidth]{0.25\textwidth}
        \includegraphics{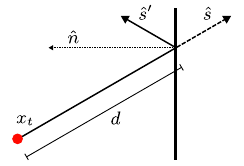}
    \end{minipage}
    \hfill
    \begin{minipage}[[0.25\textwidth]{0.7\textwidth}
        \caption{Reflection against a linear boundary. For a step $s$ with magnitude $|s|$ and direction $\hat{s}$ the distance to the boundary described by the normal $\hat{n}$ and offset $b$ is $d=\frac{\innerprod{\hat{s}}{x_t} - b}{\innerprod{\hat{s}}{\hat{n}}}$. The reflected direction is given by \(\hat{s}' = \hat{s} - 2\innerprod{\hat{s}}{\hat{n}}\hat{n}\).}
    \label{fig:reflection_diagrams_lin}
    \end{minipage}
    \vspace{1em}
    \begin{minipage}{0.25\textwidth}
        \includegraphics{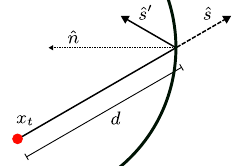}
    \end{minipage}
    \hfill
    \begin{minipage}{0.7\textwidth}
        \caption{Reflection against a spherical boundary. For a sphere of radius \(r\), the distance to the boundary from \(x_t\) in direction \(s\) is given by \(d= \frac{1}{2}(\innerprod{\hat{s}}{x_t}^2 + 4 (r^2 - \norm{x_t}^2))^{1/2} -\frac{1}{2}\innerprod{\hat{s}}{x_t}\). The normal at the intersection can be computed as the unit vector in the direction \(-2(d\hat{s} + x_t)\), and then \(\hat{s}'\) as above.}
        \label{fig:reflection_diagrams_sph}
    \end{minipage}
\end{figure}


%% file: sections/related_work.tex
\section{Related Work}
\label{sec:related_work}
%
\paragraph{Sampling on constrained manifolds.}
Sampling from a distribution on a space defined by a set of constraints is an
important ingredient in several computational tasks, such as computing the
volume of a polytope \citep{lee2017Geodesic}.  Incorporating such constraints
within MCMC algorithms while preserving fast convergence properties is an active
field of research \citep{kook2022sampling,lee2017Geodesic,noble2022Barrier}. In
this work, we are interested in sampling from the uniform distribution defined
on the constrained set in order to define a proper \emph{forward process} for
our diffusion model. Log-barrier methods such as the Dikin walk or Riemannian
Hamiltonian Monte Carlo \citep{Kannan2009,lee2017Geodesic,noble2022Barrier}
change the geometry of the underlying space and define stochastic processes
which never violate the constraints, see \cite{Kannan2009} and \cite{noble2022Barrier} for
instance. If we keep the Euclidean metric, then \emph{unconstrained} stochastic
processes might not be well-defined for all times.  To counter this effect, it
has been proposed to \emph{reflect} the Brownian motion
\citep{williams1987Reflected,petit1997Time,shkolnikov2013Timereversal}. Finally,
we also highlight hit-and-run approaches \citep{Smith1984,Lovasz2006}, which
generalise Gibbs' algorithm and enjoy fast convergence properties provided that
one knows how to sample from the one-dimensional marginals.


\paragraph{Diffusion models on manifolds.}
\citet{debortoli2022riemannian} extended the work of \citet{song2020score} to
Riemannian manifolds by defining forward and backward stochastic processes in
this setting. 
Concurrently, a similar framework was introduced by \citet{huang2022Riemannian},
extending the maximum likelihood approach of \citet{huang2021variational}.
Existing applications of denoising diffusion models on Riemannian manifolds have
been focused on well-known manifolds for which one can find metrics so that the
framework of \citet{debortoli2022riemannian} applies. In particular, on compact
Lie groups, geodesics and Brownian motions can be defined in a canonical
manner. Their specific structure can be leveraged to define efficient diffusion
models \citep{yim2023se}. \citet{leach2022denoising} define diffusion models on
$\mathrm{SO}(3)$ for rotational alignment, while \citet{jing2022torsional}
consider the product of tori for molecular conformer
generation. \citet{corso2022DiffDock} use diffusion models on
$\rset^3 \times \mathrm{SO}(3) \times \mathrm{SO}(2)$ for protein docking
applications. RFDiffusion \citep{watson2022Broadly} and FrameDiff also
incorporate $\mathrm{SE}(3)$ diffusion models. Finally, \citet{urain2022se}
introduce a methodology for $\mathrm{SE}(3)$ diffusion models with applications
to robotics.

\paragraph{Comparison with \cite{lou2023reflected}.}
We now discuss a few key differences between our work and the reflected diffusion models presented in \cite{lou2023reflected}.
First, in the hypercube setting, our methodologies are identical from a theoretical viewpoint.
The main
difference is that \cite{lou2023reflected} use an approxiamted version of the
DSM loss (\ref{eq:denoising_sm}), whereas we rely on the ISM loss
(\ref{eq:implicit_sm}). \citet{lou2023reflected} exploit the specific factorised structure of the hypercube to make the DSM loss tractable, leading to significant practical advantages. These can however be directly employed in our framework for reflected models, and also in the log-barrier setting.  Second, in our work, we target scientific applications
where the underlying geometry is not Euclidean, whereas \cite{lou2023reflected}
focus on the case where the constrained domain of interest is the hypercube (a
subset of Euclidean space) with image applications, or where the domain can be
easily projected into the hypercube, such as the simplex. Our approach is designed to handle a wider range of settings, such as non-convex polytopes in Euclidean space, or non-Euclidean geometries.


%% file: sections/experiments.tex
\section{Experimental results}
\label{sec:experiments}

\setlength{\tabcolsep}{13pt}
\begin{table}[t]
    \centering
    \caption{
    MMD metrics between samples from synthetic distributions and trained constrained and unconstrained (Euclidean) diffusion models.
    Means and confidence intervals are computed over 5 different runs.
    }
    \label{tab:synthetic_polytope}
    
    \begin{adjustbox}{max width=\textwidth}
    \begin{tabular}{cccccccc}
    \toprule
    \multirow{2}{*}{Space} & \multirow{2}{*}{$d$} & \multicolumn{2}{c}{Log-barrier} & \multicolumn{2}{c}{Reflected} & \multicolumn{2}{c}{Euclidean}\\
    & & \textsc{MMD} & \% in $\mathcal{M}$ & \textsc{MMD} & \% in $\mathcal{M}$ & \textsc{MMD} & \% in $\mathcal{M}$ \\
    \midrule
    \multirow{3}{*}{$[-1,1]^d$} & 2 &  $.066 \pms{.006}$ &100.0 &  $.055 \pms{.015}$ & 100.0&  $.062 \pms{.011}$ & 98.8 \\
     & 3 &  $.209 \pms{.077}$ &100.0 & $.080 \pms{.004}$ &100.0 &  $.076 \pms{.004}$ &98.5\\
     & 10 &  $.330 \pms{.004}$& 100.0&  $.313 \pms{.048}$ & 100.0&  $.081 \pms{.005}$ & 96.4\\
     \midrule
    \multirow{3}{*}{$\Delta^d$} & 2 &  $.050 \pms{.012}$ & 100.0&$.043 \pms{.002}$ &100.0 & $.055 \pms{.013}$ & 96.4\\
     & 3 & $.238 \pms{.009}$& 100.0& $.181 \pms{.007}$ & 100.0& $.068\pms{.014}$ &96.3\\
     & 10 & $.275 \pms{.015}$&100.0 &$.290 \pms{.009}$ &100.0 & $.060	\pms{.003}$ &92.6\\
    \bottomrule
    \end{tabular}
    \end{adjustbox}
\end{table}

\begin{figure}[t!]
\begin{subfigure}{0.49\linewidth}
    \includegraphics[width=\textwidth]{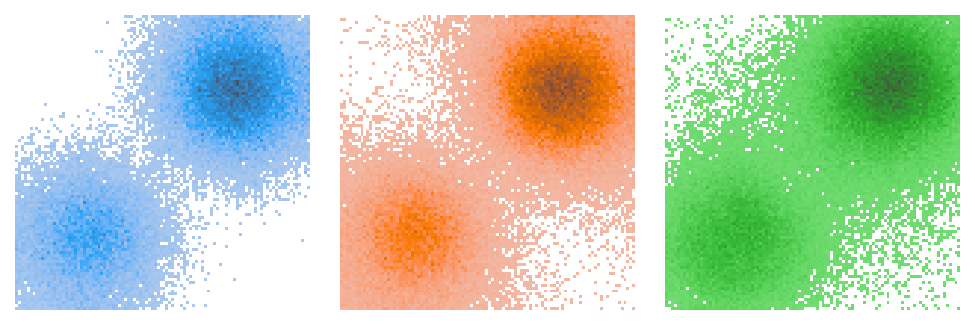}
    \put(-200,-10){Data}
    \put(-137,-10){Log-barrier}
    \put(-60,-10){Reflected}
    \caption{2D square data.}
\end{subfigure}
\hfill
\begin{subfigure}{0.49\linewidth}
    \includegraphics[width=\textwidth]{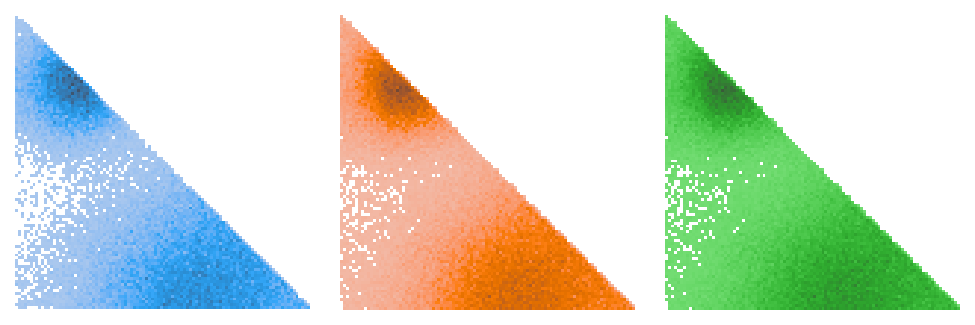}
    \put(-200,-10){Data}
    \put(-137,-10){Log-barrier}
    \put(-60,-10){Reflected}
    \caption{2D Dirichlet data.}
\end{subfigure}
\caption{
Histograms of samples from the data distribution and from trained constrained diffusion models.
}
\label{fig:polytope_synthetic}
\end{figure}
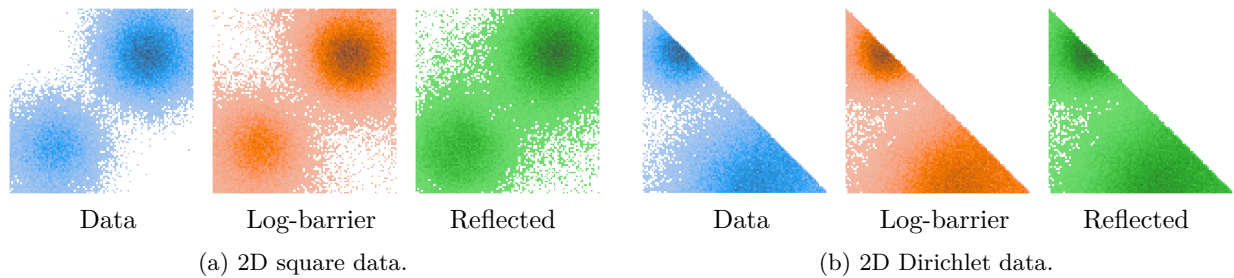

To demonstrate the practical utility of the constrained diffusion models introduced in \cref{sec:constrained_diffusion_models}, we evaluate them on a series of increasingly difficult synthetic tasks on convex polytopes, including the hypercube and the simplex, in \cref{subsec:synthetic_data}.
We then highlight their applicability to real-world settings by considering two problems from robotics and protein design.
In particular, we show that our models are able to learn distributions over the space of $d\times d$ symmetric positive definite (SPD) matrices $S_{++}^d$ under trace constraints in \cref{subsec:psd_matrices}---a setting that is essential to describing and controlling the motions and exerted forces of robotic platforms \citep{jaquier2021geometry}.
In \cref{subsec:protein_loops}, we use the parametrisation introduced in \cite{han2006inverse} to map the problem of modelling the conformational ensembles of proteins under positional constraints on their endpoints to the product manifold of a convex polytope and a torus.
The code is available \href{https://github.com/oxcsml/constrained-diffusion}{here}. We refer the reader to \cref{sec:exp_details} for more details on the experimental setup.


\subsection{Method characterisation on convex polytopes}
\label{subsec:synthetic_data}
\label{sec:diff_methods}

First, we aim to assess the empirical performance of our methods on constrained manifolds of increasing dimensionality. For this, we focus on polytopes and construct synthetic datasets that represent bimodal distributions. In particular, we investigate two specific instances of polytopes: hypercubes and simplices. 
In \cref{sec:app_synthetc_data} we also present results on the Birkhoff polytope. 
We quantify the performance of each model via the Maximum Mean Discrepancy (\textsc{MMD}) \citep{gretton2012kernel}, which is a kernel-based metric between distributions. 
%
We present a qualitative comparison of the logarithmic barrier and reflected Brownian motion models in \cref{fig:polytope_synthetic}, and observe that both methods recover the data distribution on the two-dimensional hypercube and simplex, although the reflected method seems to produce a better fit.
In \cref{tab:synthetic_polytope}, we report the \textsc{MMD} between the data distribution and the learnt diffusion models.
Here, we similarly observe that the reflected method consistently yields better results than the log-barrier one.

We additionally compare both of these models to a set of unconstrained Euclidean diffusion models, noting that they are outperformed by the constrained models in lower dimensions, but generate better results in higher dimensions. There are a number of potential explanations for this: First, we note that the mixture of Normal distributions we use as a synthetic data-generating process places significantly less probability mass near the boundary as its dimensionality increases, more closely resembling an unconstrained mixture distribution that is easier for the Euclidean diffusion models to learn, while posing a challenge to the log-barrier and reflected diffusion models that initialise at the uniform distribution within the constraints. Additionally, we note that the design space and hyperparameters used for all experiments were informed by best practices for Euclidean models that may be suboptimal for the more complex dynamics of constrained diffusion models.

\subsection{Modelling robotic arms under force and velocity constraints}
\label{subsec:psd_matrices}

\begin{figure}[t]
\centering
\begin{subfigure}{0.22\columnwidth}
    \includegraphics[width=\linewidth]{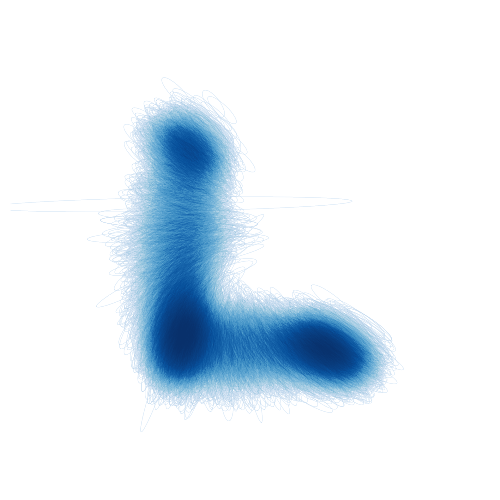}
    \caption{Samples from the data distribution.}
\end{subfigure}
\hfill
\begin{subfigure}{0.22\columnwidth}
    \includegraphics[width=\linewidth]{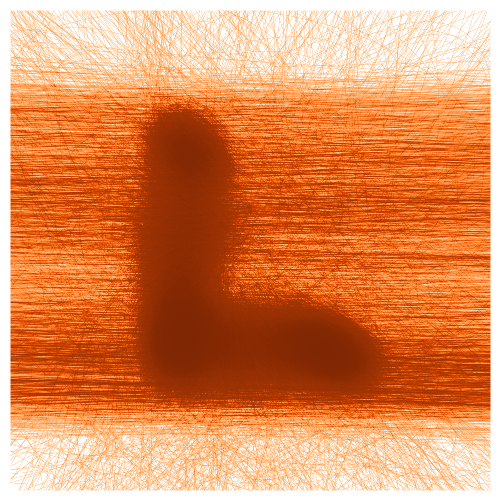}
    \caption{Samples from our log-barrier diffusion model.}
\end{subfigure}
\hfill
\begin{subfigure}{0.22\columnwidth}
    \includegraphics[width=\linewidth]{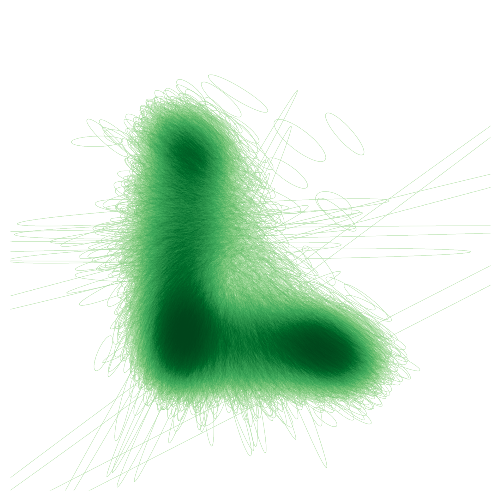}
    \caption{Samples from our reflected diffusion model.}
\end{subfigure}
\hfill
\begin{subfigure}{0.22\columnwidth}
    \includegraphics[width=\linewidth]{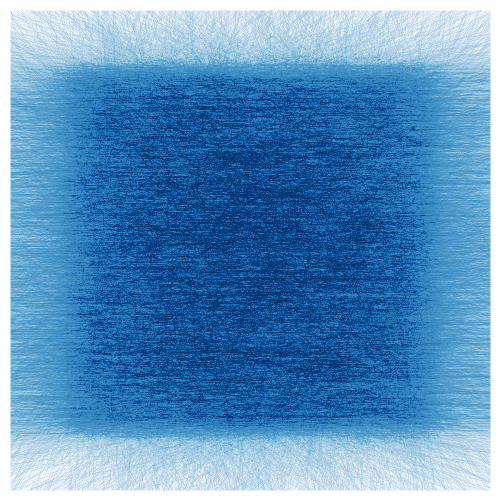}
    \caption{Samples from the uniform distribution.}
\end{subfigure}
\caption{Samples in $S_{++}^2\times\mathbb{R}^2$ from (a) the data distribution, (b) our log-barrier diffusion model, (c) our reflected diffusion model and (d) the uniform distribution. Each sample is visualised as the manipulability ellipsoid encoded by the SPD matrix $M\in S_{++}^2$ placed at the corresponding location in $\mathbb{R}^2$. Additional results and full correlation plots are postponed to~\Cref{sec:app_spd_exp}.
 }
\label{fig:spd_data}
\end{figure}

Accurately determining and controlling the movement and exerted forces of robotic platforms is a fundamental problem in many real-world robotics applications. A kinetostatic descriptor that is commonly used to quantify the ability of a robotic arm to move and apply forces along certain dimensions is the so-called manipulability ellipsoid $E\in\mathbb{R}^d$ \citep{yoshikawa1985manipulability}, which is naturally described as a symmetric positive definite (SPD) matrix $\mathrm{M} \in\mathbb{R}^{d\times d}$ \citep{jaquier2021geometry}. The manifold of such $d\times d$ SPD matrices, denoted as $S_{++}^d$, is defined as the set of matrices \( \{ x^\top \rmM x \geq 0 , \  x\in \mathbb{R}^d : \rmM \in \rset^{d \times d} \} \).
In many practical settings, it may be desirable to constrain the volume of $E$ to retain flexibility or limit the magnitude of an exerted force, which can be expressed as an upper bound on the trace of $\rmM$, i.e.\ as the inequality constraint \(\sum_{i=1}^d \rmM_{ii} < C\) with $C >0$. 
Constraining the rest of the entries of the matrix to ensure it is SPD is non-trivial.
Alternatively, we can parameterise the SPD matrices via their Cholesky decomposition.
Each SPD matrix has a unique decomposition of the form \(\rmM = \rmL \rmL^\top\), where \( \rmL\) is a lower triangular matrix with strictly positive diagonal \citep[p.143]{golub2013matrix}.
Constraining the entries of this matrix simply requires ensuring the diagonal is positive.
The trace of the SPD matrix is given by \(\mathrm{Tr}(\rmM) = \mathrm{Tr}(\rmL \rmL^\top) = \sum_{ij}\rmL_{ij}^2\), and results in the constraint on the entries of $\rmL$ to live in a ball of radius $C$.
We additionally model the two-dimensional position of the arm.
In summary, the space over which we parameterise the diffusion models is defined as $\{ \mathrm{L} \in \R^{d(d+1)/2} : \mathrm{L}_{i,i} > 0, \sum_{i,j}  \mathrm{L}_{i,j}^2 < C \} \times \R^2$.
Under the Euclidean metric, we can apply both our log-barrier and reflected approaches.
The positive diagonal (linear) constraint is handled similarly to the polytope setting.
The reflection on the ball boundary is defined and illustrated on \cref{fig:reflection_diagrams_lin,fig:reflection_diagrams_sph}.

Using this framework, we model the datasets presented in \cite{jaquier2021geometry} (see \Cref{sec:robotic_arm} for full experimental details). 
The joint distribution over the SPD matrices (represented as ellipsoids) and their positions is presented in~\Cref{fig:spd_data}. 
We qualitatively observe that the reflected method is able to model the joint data distribution better than the log-barrier one. This is reflected by an MMD of 0.161 and 0.247, respectively.
\subsection{Modelling protein loops with anchored endpoints}
\label{subsec:protein_loops}

\begin{figure}[t]
\centering
\vspace{-0.5cm}
\begin{subfigure}[t]{0.45\textwidth}
  \centering
  \includegraphics[height=4cm]{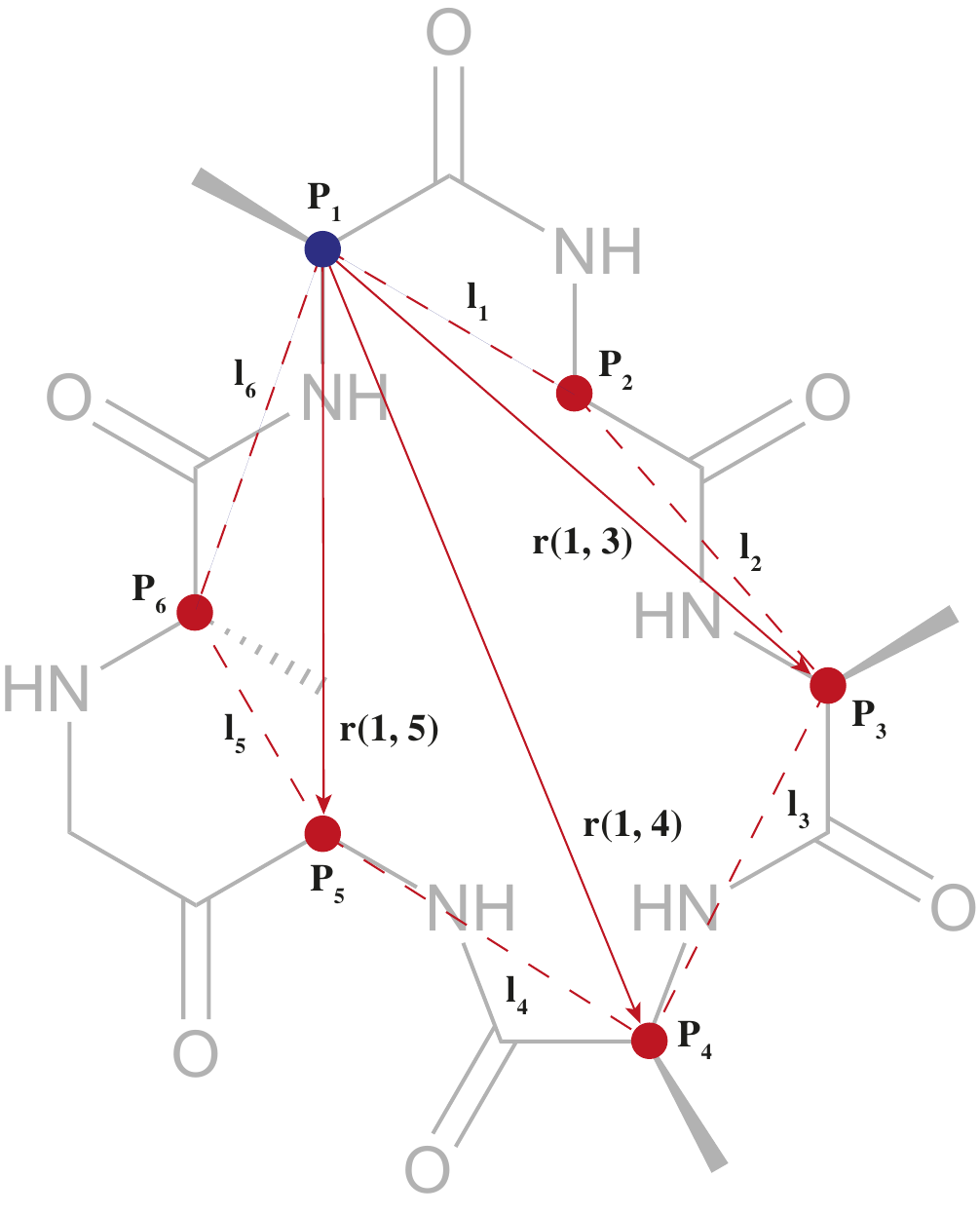}
  \caption{An illustrative diagram of the parameterisation used for the conformational modelling of the $C_\alpha$ trace of a cyclic peptide, introduced in \cite{han2006inverse}.}
  \label{fig:cyclic_peptide_main}
\end{subfigure}
\hfill
\begin{subfigure}[t]{0.45\textwidth}
  \centering
  \includegraphics[height=4cm]{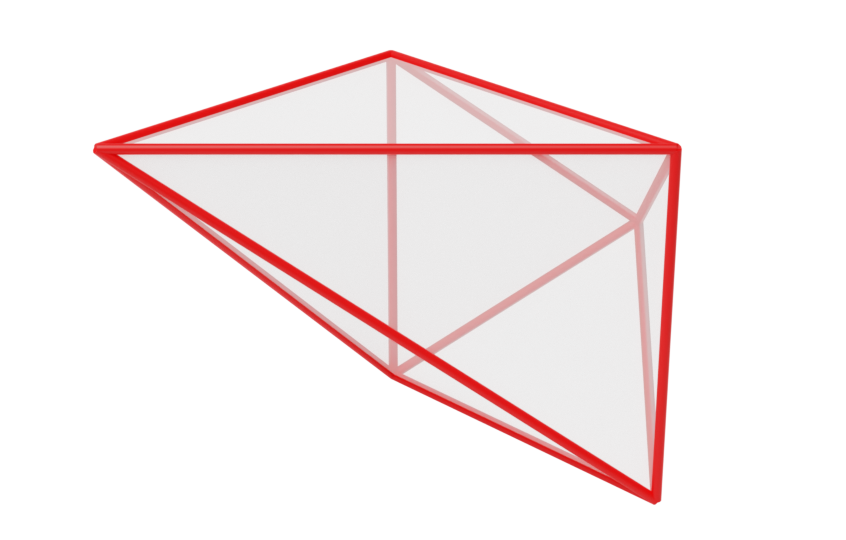}
  \caption{The convex polytope constraining the diagonals of the triangles for the given bond lengths in the illustrated molecule. The total design space is the product of this polytope with the 4D flat torus.}
  \label{fig:6s_random_walk_main}
\end{subfigure}
\caption{Illustrations of the parametrisation used to model distribution of polypeptide backbone conformations under anchor point distance constraints as the product of a convex polytope $\mathbb{P}$ and torus $\mathbb{T}$: $\mathbb{P}^{3} \times \mathbb{T}^{4}$.}
\label{fig:han_and_rudolph_main}
\end{figure}



Modelling the conformational ensembles of proteins is an important task in the field of molecular biology, particularly in the context of bioengineering and drug discovery.
In many data-scarce practical settings such as antibody or enzyme design, it is often unnecessary or even undesirable to model the structure of an entire protein, as researchers are primarily interested in specific functional sites with distinct biochemical properties.
However, generating conformational ensembles for such substructural elements necessitates positional constraints on their endpoints to ensure that they can be accommodated by the remaining scaffold.

This problem can be reformulated as modelling a spatial chain with spherical joints and fixed end points. Following the framework outlined in \citet{han2006inverse}, we parametrise the conformations of such chains with $d$ fixed-length links and arbitrary end-point constraints as the product of a convex polytope $\mathbb{P}$ and torus $\mathbb{T}$: $\mathbb{P}^{(d-3)} \times \mathbb{T}^{(d-2)}$.
The essential idea of this parameterisation is to fix one end point as an ``anchor'' and model the chain as the series of triangles formed by the anchor and each pair of adjacent joints in the chain.
A point in the polytope corresponds to the lengths of the diagonals of these triangles, and a point in the torus to the angles between each pair of subsequent triangles.
See~\Cref{fig:han_and_rudolph_main} for an illustration and \Cref{sec:protein_param} for a full description.

Using this framework, we model the conformational landscape of the cyclic peptide \textsc{c-AAGAGG}, consisting of a circular polypeptide chain with coinciding endpoints. We generate $10^6$ backbone conformations using tools from molecular dynamics \citep{eastman2017openmm, hornak2006comparison} and divide them into training and evaluation datasets, (see~\Cref{sec:protein_data} for full experimental details).
Drawing on the definition above, the space on which we effectively parameterise our constrained diffusion models for a circular polypeptide chain of length $d=6$ is given by the product manifold $\mathbb{P}^3 \times \mathbb{T}^4$. 

\begin{figure}[tb]
\centering
\begin{subfigure}{0.23\textwidth}
  \centering
  \includegraphics[width=\linewidth]{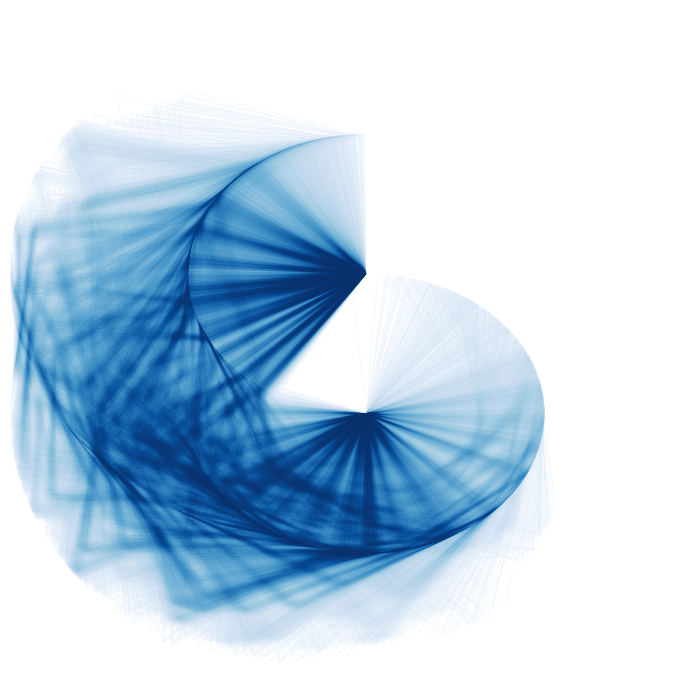}
  \caption{Samples from the data distribution.}
  \label{fig:loop_plot_smoothened}
\end{subfigure}
\hfill
\begin{subfigure}{0.23\textwidth}
  \centering
  \includegraphics[width=\linewidth]{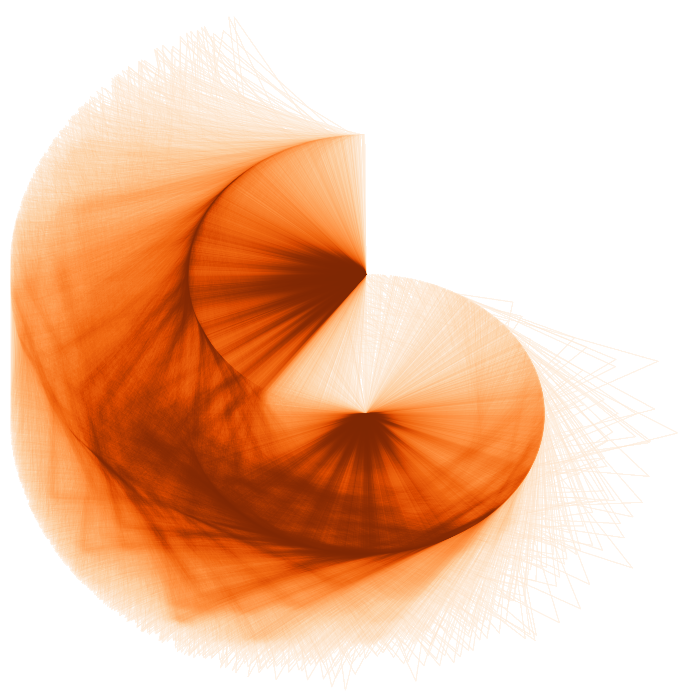}
  \caption{Samples from our log-barrier diffusion model.}
  \label{fig:loop_plot_logbarrier}
\end{subfigure}
\hfill
\begin{subfigure}{0.23\textwidth}
  \centering
  \includegraphics[width=\linewidth]{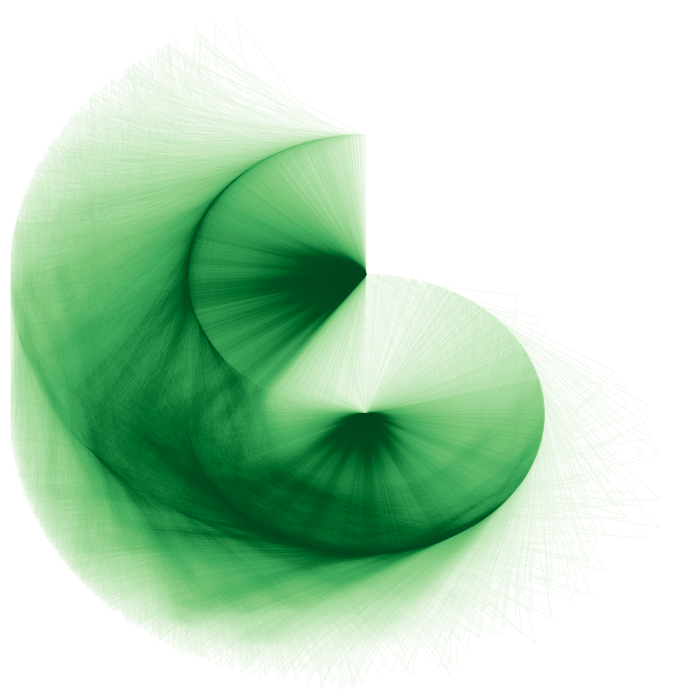}
  \caption{Samples from our reflected diffusion model.}
  \label{fig:loop_plot_sampled}
\end{subfigure}
\hfill
\begin{subfigure}{0.23\textwidth}
  \centering
  \includegraphics[width=\linewidth]{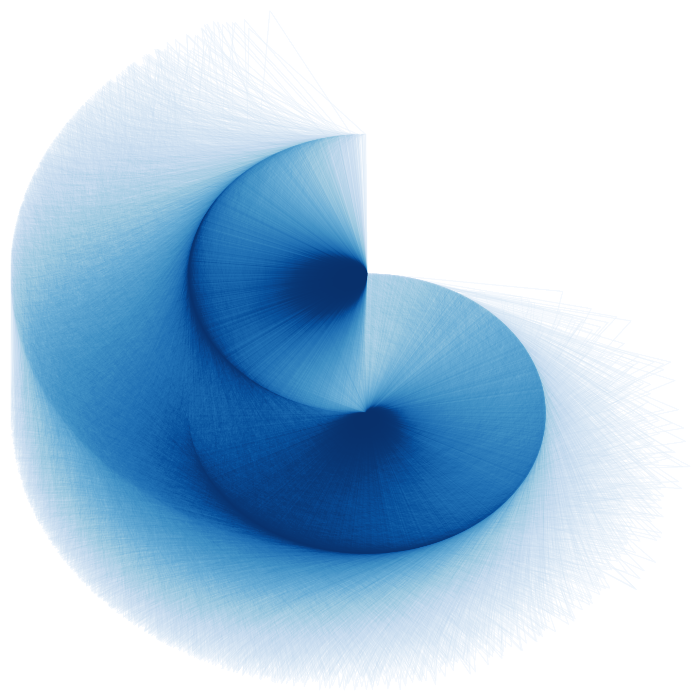}
  \caption{Samples from the uniform distribution.}
  \label{fig:loop_plot_uniform}
\end{subfigure}
\caption{Planar projection of the modelled $C_\alpha$ chains from (a) the training dataset, (b) our log-barrier diffusion model, (c) our reflected diffusion model and (d) the uniform distribution. Additional results and full correlation plots are postponed to~\Cref{sec:app_protein_exp}.}
\label{fig:loop_plot}
\end{figure}

To learn a distribution over this space, we leverage the methodology introduced in \cref{sec:constrained_diffusion_models} for the polytope component $\mathbb{P}^3$ and in \citet{debortoli2022riemannian} for the torus component $\mathbb{T}^4$.
A qualitative comparison of samples from the data distribution, our log-barrier and reflected diffusion models, and the uniform distribution is presented in \Cref{fig:loop_plot_smoothened,fig:loop_plot_logbarrier,fig:loop_plot_sampled,fig:loop_plot_uniform}. For enhanced visual clarity, we project the modelled spatial chain onto the 2D plane by removing the (unconstrained) torus component of the product manifold and only plotting the planar chains encoded by the (constrained) polytope component (a correlation plot of the full product manifold is presented in \Cref{fig:loops_pairs}).

It is apparent that the data distribution is highly multimodal, encompassing a large number of locally optimal conformational clusters. Nevertheless, our reflected and log-barrier diffusion models are able to robustly approximate this energetic landscape, producing samples that reflect key conformational states and producing comparable \textsc{MMD} metrics of $0.032\pms{0.021}$ and $0.032\pms{0.001}$, respectively. As a point of comparison, the uniform distribution on the polytope-torus product has an \textsc{MMD} of $0.112 \pms{0.001}$. 



%% file: sections/conclusions.tex
\section{Discussion}

Learning complex distributions whose support is confined to constrained spaces is a crucial task with many applications in the natural and engineering sciences, including computational statistics \citep{Morris2002}, robotics \citep{han2006inverse},  quantum physics \citep{lukens2020practical} and computational biology \citep{Thiele2013}.
In this work we extend continuous diffusion models to this setting, proposing two complementary approaches---one based on log-barrier methods and the other on the reflected Brownian motion.
For both methods, we derive the time-reversal formula, propose discretisation schemes and extend the score-matching toolbox. We demonstrate the utility of our methods on a range of synthetic and real-world tasks, including the constrained conformational modelling of proteins and robotic arms, and find that reflected methods, while enjoying fewer theoretical guarantees than their log-barrier counterparts, often yield preferable results.

We conclude by highlighting important directions of future research.  First, the
computational cost of performing the reflection when discretising the
reflected Brownian motion is high. 
Finding numerically efficient approximations of the reflected process is
therefore necessary to extend this methodology to very high dimensional
settings.  Second, the retraction used in place of the exponential map for the
barrier method leads to a high number of discretisation steps to ensure a good
approximation. Designing a faster forward process for the log-barrier method is key to
targeting more complex distributions. 


\newpage

%% file: appendices/intro_appendix.tex
\section*{Introduction}
\label{sec:introduction}


In this supplementary, we first recall some key concepts from Riemannian
geometry in \cref{sec:riemannain_intro}.  In \cref{sec:brownian} we remind the
expression of the Brownian motion in local coordinates. Details about the
geodesic Brownian motion are given in \Cref{sec:geod-brown-moti}.  Background on
the Skorokhod problem is given in \Cref{sec:refl-brown-moti}.  In
\cref{sec:impl-score-match} we derive the implicit score matching loss. We give
details about the likelihood evaluation in \Cref{sec:likel-eval}.  Then in
\cref{sec:time-revers-refl} we prove the time-reversal formula of reflected
Brownian motion.
In \cref{sec:protein_background} we give some background on the conformational modelling of proteins backbone for the experiment in \cref{subsec:protein_loops}.
In \cref{sec:robotic_arm} we detail the geometrical constraint arising from the configurational robotics arm modelling experiment from \cref{subsec:psd_matrices}. Additional results, training and miscalleneous experimental details are reported in \Cref{sec:exp_details}.


%% file: appendices/manifold_concepts.tex
\section{Manifold concepts}
\label{sec:riemannain_intro}
For readers unfamiliar with Riemannian geometry here we give a brief overview of some of the key concepts. This is not a technical introduction, but a conceptual one for the understanding of terms. For a technical introduction, we refer readers to \citet{lee2013smooth}.
A Riemannian manifold is a tuple $(\M, \metric)$ with $\M$ a smooth manifold and $\metric$ a metric which defines an inner product on tangent spaces.

\begin{wrapfigure}{r}{2in}
    \includegraphics[width=2in]{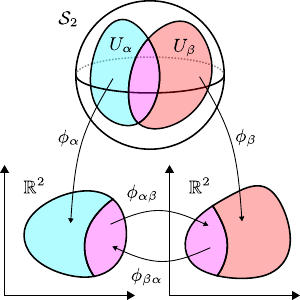}
    \caption{Example charts of the 2D manifold \(\c{S}^2\).}
\end{wrapfigure}
A \emph{smooth manifold} is a topological space which locally can be described by Euclidean space. It is characterised by a family of \emph{charts} \(\{U \subset \c{M}, \phi:  U \to \R^d\}\), homeomorphic mappings between subsets of the manifold and Euclidean space. The collection of charts must cover the whole manifold. For the manifolds to be smooth the charts must be smooth, and the transition between charts where their domains overlap must also be smooth.

The \emph{metric} on a Riemannian manifold gives the manifold a notion of distance and curvature. The same underlying smooth manifold with different metrics can look wildly different. The metric is defined as a smooth choice of positive definite inner product on each of the tangent spaces of the manifold. That is we have at every point a symmetric bilinear map 
\[\metric(p): \mathrm{T}_p\c{M} \times \mathrm{T}_p\c{M} \to \R\]
The tangent space of a point on a manifold is the generalisation of the notion of tangent planes and can be thought of as the space of derivatives of scalar functions on the manifold at that point. The collection of all tangent spaces is written as \(\mathrm{T}\c{M} = \bigcup_{p\in \c{M}} \mathrm{T}_p \c{M}\) and is called the tangent bundle. It is also manifold. Vector fields on manifolds are defined by making a choice of tangent vector at every point on the manifold in a smooth fashion. The space of vector fields is written at \(\Gamma\del{\mathrm{T}\c{M}}\) and is more technically the space of \emph{sections} of the tangent bundle.

One thing that the metric itself does not immediately define is how different tangent spaces at points on the manifold relate to one another. For this, we need additionally the concept of a \emph{connection}. A connection is a map that takes two vector fields and produces a derivative of the first with respect to the second, that is a function \(\nabla: \Gamma(\mathrm{T}\c{M}) \times \Gamma(\mathrm{T}\c{M}) \to \Gamma(\mathrm{T}\c{M})\) and it typically written as \(\nabla(X, Y) = \nabla_X Y\). Such a connection must for \(X, Y,Z\in\Gamma(\mathrm{T}\c{M})\) and smooth functions on the manifold \(a,b: \c{M}\to\R\) obey the following conditions:
\begin{enumerate}[label=(\alph*)]
    \item \(\nabla_{aX + bY}Z = a\nabla_X Z + b\nabla_Y Z\),
    \item \(\nabla_X(Y+Z) = \nabla_X Y + \nabla_X Z\),
    \item \(\nabla_X(aY) = \partial_X aY + a \nabla_X Y\),
\end{enumerate}
where \( \partial_X aY \) is the regular \emph{directional derivative} of \(aY\) in the direction \(X\). These conditions ensure the connection is a well-defined derivative. 

\begin{wrapfigure}[14]{l}{2in}
    \includegraphics[width=2in]{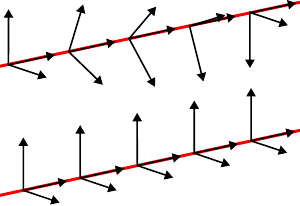}
    \caption{\emph{Top:} Parallel transport of vectors along the red path with non-zero torsion. \emph{Bottom:} Parallel transport of vectors along the red path with zero torsion. Both under the Euclidean metric.}
\end{wrapfigure}
On a given manifold, there are infinitely many connections. Fortunately, there is a natural choice, called the \emph{Levi-Cevita} connection if we impose two additional conditions:
\begin{enumerate}[label=(\alph*)]
    \item \(X \cdot (\metric(Y,Z)) = \metric(\nabla_X Y, Z) + \metric(Y, \nabla_X Z)\),
    \item \(\sbr{X, Y} =  \nabla_X Y - \nabla_Y X\),
\end{enumerate}
where \(\sbr{\cdot, \cdot}\) is the Lie bracket. The first condition ensures that the metric is preserved by the connection. That is to say, the \emph{parallel transport} (to be defined shortly) using the connection leaves inner products unchanged on the manifold. The second ensures the connection is \emph{torsion-free}. 
The change in tangent space along a geodesic (again to be defined shortly) can be described in two parts, the \emph{curvature}, how the tangent space rotates perpendicular to the direction of travel, and the torsion, how the tangent space rotates around the axis of the direction of travel. The curvature of the connection is uniquely fixed by the other 4 conditions (the well-defined derivative and preservation of the metric). The torsion however is not fixed. By requiring it to be zero we ensure a unique connection. The requirement of zero torsion also has implications for ensuring integrability on the manifold.

With the metric and the Levi-Cevita connection in hand, we can define a number of key concepts. 

\begin{wrapfigure}[18]{r}{2in}
    \includegraphics[width=2in]{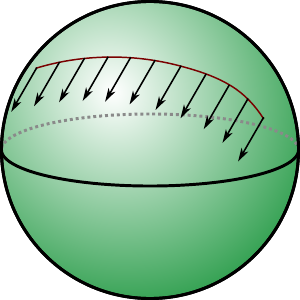}
    \caption{Parallel transport of vectors along a path on the sphere.}
\end{wrapfigure}
We say that a vector field \(X\) is \emph{parallel} to a curve \(\gamma: (0,1)\to\c{M}\) if
\[\nabla_{\gamma'} X = 0\]
where \(\gamma': (0,1)\to \mathrm{T}_{\gamma(t)}\c{M}\) is the derivative of the path.
For two points on the manifold \(p,q\in\c{M}\) and a curve between them, \(\gamma\), \(\gamma(0)=p\), \(\gamma(1)=q\), for an initial vector \(X_0 \in \mathrm{T}_p\c{M}\) there is a unique vector field \(X\) that is parallel to \(\gamma\) such that \(X(p) = X_0\). This induces a map between the tangent spaces at \(p\) and \(q\)
\[\tau_\gamma: \mathrm{T}_p\c{M} \to \mathrm{T}_q\c{M}\]
This map is referred as the \emph{parallel transport} of tangent vectors between \(p\) and \(q\), and this satisfies the condition that for \(\v{v}, \v{u} \in \mathrm{T}_p\c{M}\)
\[\metric(p)(\v{v}, \v{u}) = \metric(q)(\tau_\gamma(\v{v}), \tau_\gamma(\v{u})).\]

A \emph{geodesic} on a manifold is the unique path on the manifold \(\gamma: (0,1) \to \c{M}\) such that $\nabla_{\gamma'}\gamma' = 0$.
It is also the shortest path between two points on a manifold in the sense that 
\[\textstyle{L(\gamma) = \int_0^1 \sqrt{\metric(\gamma(t))(\gamma'(t), \gamma(t))}, }\]
is minimal out of any path between the start and end of the geodesic. Geodesics give the notion of `straight lines' on manifolds.
We define the \emph{exponential map} on a manifold as the mapping between an element of the tangent space at point \(p\), \(\v{v}\in T_p\c{M}\) and the endpoint of the unique geodesic \(\gamma\) with \(\gamma(0) = p\) and \(\gamma'(0) = \v{v}\).
In the tangent space of a manifold, we require the notion of a reflection in order to reflect geodesics off of boundary constraints.
If at a point in the manifold with have \(\v{v}\in T_p\c{M}\) and a unit vector normal to the constraint \(\v{n}\in T_p\c{M}\), then the reflection of this vector is given by $\v{v}' = \v{v} - 2\metric(\v{v},\v{n}) \v{n}$.


%% file: appendices/brownian.tex

\section{Brownian motion in local coordinates}
\label{sec:brownian}

We consider a smooth function $f \in \rmc^\infty(\M)$. The Laplace-Beltrami operator $\Delta_\M$ is given by
$\Delta_{\M}(f) = \mathrm{div}(\mathrm{grad}(f)))$. In local coordinates we have
\begin{equation}
\textstyle{
  \mathrm{div}(X) = (\det(\metric)^{-1/2}) \sum_{i=1}^d \partial_i (\det(\metric)^{1/2} X_i)  ,
  }
\end{equation}
as well as
\begin{equation}  
  \mathrm{grad}(f) = \metric^{-1} \nabla f  .
\end{equation}
Therefore, the Laplace-Beltrami operator is given by
\begin{equation}
\textstyle{
  \Delta_\M(f) = \sum_{i,j=1}^d \metric_{i,j}^{-1} \partial_{i,j} f + (\det(\metric)^{-1/2}) \sum_{i,j=1}^d \partial_i (\det(\metric)^{1/2} \metric_{i,j}^{-1}) \partial_j f .}
\end{equation}
Therefore, in local coordinates the infinitesimal generator associated with the
Laplace-Beltrami operator is given by
\begin{equation}
\textstyle{
  \mathcal{A}(f) = \sum_{i,j=1}^d \metric_{i,j}^{-1} \partial_{i,j} f + \langle b^i, \nabla f \rangle  ,}
\end{equation}
with
\begin{equation}
  \label{eq:def_drift}
  \textstyle{
  b^i = (\det(\metric)^{-1/2}) \sum_{j=1}^d \partial_j (\det(\metric)^{1/2} \metric_{i,j}^{-1}) .} 
\end{equation}
Therefore, the dual operator associated with $\mathcal{A}$ is given by
\begin{equation}
\textstyle{
  \mathcal{A}^\star(f) = \sum_{i,j=1}^d \partial_{i,j} (\metric_{i,j}^{-1} f) - \sum_{i=1}^d \partial_i (b^i f) .}
\end{equation}
Note that by letting $f = \det(\metric)^{1/2}$ we get that $\mathcal{A}^\star(f) = 0$
and therefore we recover that $p \propto \det(\metric)$ is the invariant
distribution of the Brownian motion.

\paragraph{Langevin dynamics on $\M$.}
We know that the Brownian motion targets $ \det(\metric)^{1/2}$. Therefore in order
to correct and sample from the uniform distribution we consider the Langevin dynamics
\begin{equation}
  \rmd \bfX_t = -\mathrm{grad} \log (\det(\metric)^{1/2})(\bfX_t) \rmd t + \sqrt{2} \rmd \bfB_t^\M .
\end{equation}
Note that in the previous equation $\mathrm{grad}$ and $\bfB_t^{\M}$ are defined
w.r.t. the metric of the manifold. In local coordinates we have
\begin{equation}
  \label{eq:forward_process}
    \rmd \bfX_t = \{b -\mathrm{grad} \log (\det(\metric)^{1/2})\}(\bfX_t) \rmd t + \sqrt{2} \metric(\bfX_t)^{-1/2} \rmd \bfB_t .
\end{equation}
where $b = \{b^i\}_{i=1}^d$ is given in \eqref{eq:def_drift}. In addition, we have
\begin{equation}
  \label{eq:expansion_grad}
  \mathrm{grad} \log (\det(\metric)^{1/2}) = \det(\metric)^{-1/2} \metric^{-1}  \nabla \det(\metric) .
\end{equation}
Using \eqref{eq:def_drift} we have
\begin{equation}
  b^i = \textstyle{(\det(\metric)^{-1/2}) \sum_{j=1}^d \partial_j (\det(\metric)^{1/2} \metric_{i,j}^{-1})} 
  = \textstyle{\sum_{j=1}^d \partial_j \metric_{i,j}^{-1} + \mathrm{grad} \log (\det(\metric)^{1/2})_i}
\end{equation}
This can also be rewritten as
\begin{equation}
  \mathrm{div}_\M(\metric^{-1}) = \textstyle{\mu + \mathrm{grad} \log (\det(\metric)^{1/2})  ,} 
\end{equation}
with
\begin{equation}
  \textstyle{\mu_i = \sum_{j=1}^d \partial_j \metric_{i,j}^{-1} .}
\end{equation}
Combining this result and \eqref{eq:expansion_grad} we get that
\eqref{eq:forward_process} can be rewritten as
\begin{equation}
  \rmd \bfX_t = \mu(\bfX_t) + \sqrt{2} \rmd \bfB_t .
\end{equation}
    Note that (up to a factor $2$) this is the same SDE as the one considered in  \citet{lee2017Geodesic}.

%% file: appendices/log_barrier.tex
\section{Geodesic Brownian Motion}
\label{sec:geod-brown-moti}

In this section, we provide some details on the geodesics Brownian motion
introduced in \Cref{sec:bsgm}. In the rest of the section, we make the following
assumption.

\begin{assumption}
  \label{assum:log_barrier_method}
  $\cl{\M} \subset \rset^d$ is compact and
  $\metric^{-1}: \ \M \to \mathcal{S}_d^{++}$ can be
  $\rmc^\infty(\rset^d, \rset^{d \times d})$. 
\end{assumption}

First, we start by showing that the process $(\bfX_t)_{t \geq 0}$ defined in
\eqref{eq:langevin_hessian_manifold} exists and that we have for any $t \geq 0$,
$\bfX_t \in \M$. We recall that $\metric = \nabla^2 \phi$ and
$\lim_{x \to \partial \M} \Phi(x) = +\infty$. 

\begin{proposition}
  Assume \textup{\Cref{assum:log_barrier_method}}.  For any $x_0 \in \M$, there
  exists a unique strong solution to \eqref{eq:langevin_hessian_manifold}
  denoted $(\bfX_t)_{t \geq 0}$. In addition, we have that for any $t \geq 0$,
  $\bfX_t \in \M$ almost surely. More precisely, we have
  $\expeLigne{\phi(\bfX_t)} \leq \phi(x_0) + t$.
\end{proposition}

\begin{proof}
  A unique strong solution $(\bfX_t)_{t \geq 0}$ of
  \eqref{eq:langevin_hessian_manifold} with starting point $x_0\in \M$ exists
  since the coefficients are smooth, see \cite[Theorem 3.1,
  p.165]{ikeda2014stochastic}. For any $A \geq 0$, we define
  $\tau_A = \inf \ensembleLigne{t \geq 0}{\Phi(\bfX_t) \geq A}$. Note that for
  any $t \in \ccint{0, \tau_A}$, $\Phi(\bfX_t) \in \M$. Using It\^o formula, we
  have for any $t \geq 0$
  \begin{equation}
    \textstyle \expeLigne{\Phi(\bfX_{t \wedge \tau_A})} = \Phi(x_0) + \expeLigne{\int_0^{t \wedge \tau_A} \mathrm{Tr}(\metric^{-1}(\bfX_s) \nabla^2 \Phi(\bfX_s) )\rmd s} = \Phi(x_0) + \expeLigne{t \wedge \tau_A} . 
  \end{equation}
  Using Fatou's lemma, and letting $A \to +\infty$, we conclude the proof.
\end{proof}

In the next result, we show that the uniform distribution is the \emph{unique}
invariant probability distribution for $(\bfX_t)_{t \geq 0}$ and that
$(\bfX_t)_{t \geq 0}$ converges to this invariant distribution. We refer to
\cite[Section 2, p.490]{meyn1993stability} for a definition of
irreducibility. We recall that the total variation of a finite (not necessarily
non-negative) measure $\mu$ over $\rset^d$ is given by
$\| \mu \|_{\mathrm{TV}} = \sup \ensembleLigne{\mu(\msa)}{\msa \in
  \mcb(\rset^d)}$.

\begin{proposition}
  Assume \textup{\Cref{assum:log_barrier_method}}. $(\bfX_t)_{t \geq 0}$ is
  $\pi$-irreducible, the uniform distribution over $\M$ is the only invariant
  probability distribution and
  $\lim_{t \to +\infty} \|\Pker_t - \pi\|_{\mathrm{TV}} = 0$, where $\Pker_t$ is
  the distribution of $\bfX_t$ for any $t \geq 0$ and $\pi$ is the uniform
  distribution over $\M$.
\end{proposition}

\begin{proof}
  Since $x \mapsto \mathrm{div}(\metric^{-1})(x)$ and $x \mapsto \metric^{-1}$
  are smooth and $\metric^{-1}(x)$ is positive definite for any $x \in \M$, we
  have that $(\bfX_t)_{t \geq 0}$ is $\pi$-irreducible, extending \cite[Lemma
  1.4]{bhattacharya1978criteria} to $\M$ and using \cite[Proposition
  2.1]{meyn1993stability}. In addition, $(\bfX_t)_{t \geq 0}$ is
  $\mathrm{T}$-Feller using \cite[Proposition 3.3]{meyn1993stability}. Combining
  these results and the fact that $\M$ is bounde, we get that
  $(\bfX_t)_{t \geq 0}$ is positive Harris recurrent \cite[Theorem
  3.2]{meyn1993stability}. The uniform distribution $\pi$ is an invariant
  distribution for \eqref{eq:langevin_hessian_manifold}. Since
  $(\bfX_t)_{t \geq 0}$ is $\pi$-irreducible, we get that this invariant measure
  is unique. Hence, we conclude using \cite[Theorem 6.1]{meyn1993stability}.
\end{proof}

Note that the convergence result in total variation could be improved. In
particular, quantitative geometric results could be derived. We finish this
section, by applying results from the Malliavin calculus to show that for any
$t > 0$, $\bfX_t$ admits a density w.r.t. the Lebesgue measure.

\begin{proposition}
  Assume \textup{\Cref{assum:log_barrier_method}}. Then, for any $t \geq 0$,
  $\bfX_t$ admits a smooth density $p_t$ w.r.t. the Lebesgue measure.
\end{proposition}

\begin{proof}
  This is a direct consequence of \cite[Theorem 2.3.3]{nualart2006malliavin}.
\end{proof}


%% file: appendices/reflected_bm.tex
\section{Reflected Brownian Motion and Skorokhod problems}
\label{sec:refl-brown-moti}

In this section, we provide the basic definitions and results to derive the
time-reversal of the reflected Brownian motion in \Cref{sec:time-revers-refl}.
We follow closely the presentation of \cite{lions1984stochastic} and
\cite{burdzy2004heat}. We first define the \emph{Skorokhod problem} for
deterministic problems. We consider $\M$ to be a smooth open bounded domain.  We
recall that the normal vector $n$ is defined on $\partial \M$ and we set
$n(x) = 0$ for any $x \notin \partial \M$.

Before giving the definition of the \emph{Skorokhod problem}, we recall what is
the space of functions of \emph{bounded variations}.

\begin{definition}
  Let $a, b \in \ooint{-\infty, +\infty}$ and
  $f: \in \rmc(\ccint{a,b}, \rset)$. We define the \emph{total variation} of
  $f$ as
  \begin{equation}
    \textstyle \mathrm{V}_{a,b}(f) = \sup \ensembleLigne{\sum_{i=0}^{n-1} \normLigne{f(x_{i+1}) - f(x_i)}}{(x_i)_{i=0}^{n-1}, \ a=x_0 \leq x_1 \leq \dots \leq x_{n-1} \leq x_n = b, \ n \in \nset} . 
  \end{equation}
  $f$ has bounded variations over $\ccint{a,b}$ if
  $\mathrm{V}_{a,b}(f) < +\infty$. Let $f \in \rmc(\coint{0,+\infty},
  \rset)$. $f$ has bounded variations over $\coint{0,+\infty}$ if for any
  $b>0$, $f$ has bounded variations over $\ccint{0,b}$. 
\end{definition}

The notion of bounded variation is a relaxation of the differentiability
requirement. In particular, if $f \in \rmc^1(\ccint{a,b}, \rset)$, we have
$\mathrm{V}_{a,b}(f) = \int_a^b \normLigne{f'(t)} \rmd t$.  In the definition of
the \emph{Skorokhod problem}, we will see that this relaxation is necessary,
even in the deterministic setting.

For any function of bounded variation $f \in \rmc(\ccint{a,b}, \rset)$ on
$\ccint{a,b}$, we define $\abs{f}: \ \ccint{a,b} \to \coint{0,+\infty}$ given
for any $t \in \ccint{a,b}$ by $\abs{f}_t = \mathrm{V}_{a,t}(f)$. Note that
$\abs{f}$ is non-decreasing and right-continuous. Therefore, we can define the
measure $\mu_{\abs{f}}$ on $\ccint{a,b}$, given for any $s, t \in \ccint{a,b}$
with $t \geq s$ by $\mu_{\abs{f}}([s,t]) = \abs{f}(t) - \abs{f}(s)$. In
particular, for any $\varphi: \ccint{a,b} \to \rset_+$, we define
\begin{equation}
  \textstyle \int_a^b \varphi(t) \rmd \abs{f}_t  =  \int_a^b \varphi(t) \rmd \mu_{\abs{f}}(t) . 
\end{equation}
In addition, $f$ can be decomposed in a difference of two non-decreasing
processes right continuous processes $g_1$, $g_2$, where for any
$t \in \ccint{a,b}$, $f(t) = g_1(t) - g_2(t)$, $g_1(t) = \abs{f}_t$ and
$g_2(t) = \abs{f}_t - f(t)$. Hence, for every $\varphi$ bounded on
$\ccint{a,b}$, we can define
\begin{equation}
  \textstyle \int_a^b \varphi(t) \rmd f(t) = \int_a^b \varphi(t) \rmd g_1(t) - \int_a^b \varphi(t) \rmd g_2(t)  .
\end{equation}
Note that these definitions can be extended to the setting where
$f: \ \ccint{a,b} \to \rset^d$.

We begin with the following result, see
\cite{lions1984stochastic}.

\begin{theorem}
  \label{thm:existence_reflected_det}
  Let $(x_t)_{t \geq 0} \in \rmc(\coint{0, +\infty}, \rset)$. Then, there
  exists a unique couple $(\bar{x}_t, k_t)_{t \geq 0}$ such that
  \begin{enumerate}[label=(\alph*)]
  \item $(k_t)_{t \geq 0}$ has bounded variation over $\coint{0,+\infty}$.
  \item $(\bar{x}_t)_{t \geq 0} \in \rmc(\coint{0,+\infty}, \cl{\M})$.
  \item For any $t \geq 0$, $x_t + k_t = \bar{x}_t$. \label{item:decomp}
  \item For any $t \geq 0$,
    $\abs{k}_t = \int_0^t \mathbf{1}_{\bar{x}_s \in \partial \M}(\bar{x}_s) \rmd \abs{k}_s$ and 
    $k_t = \int_0^t n(\bar{x}_s) \rmd \abs{k}_s$.
  \end{enumerate}
\end{theorem}

Let us discuss \Cref{thm:existence_reflected_det}. First,
\Cref{thm:existence_reflected_det}-\ref{item:decomp} states the original
(unconstrained) process $(x_t)_{t \geq 0}$ can be decomposed into a constrained
version $(\bar{x}_t)_{t \geq 0}$ and a bounded variation process
$(k_t)_{t \geq 0}$. The process $(\abs{k}_t)_{t \geq 0}$ counts the number of
times the constrained process $(\bar{x}_t)_{t \geq 0}$ hits the boundary. More
formally, we have
$\abs{k}_t = \int_0^t \mathbf{1}_{x \in \partial \M}(\bar{x}_s) \rmd \abs{k}_s$. When,
we hit the boundary, we reflect the process. This condition is expressed in
$k_t = \int_0^t n(\bar{x}_s) \rmd \abs{k}_s$.

We now consider the extension to stochastic processes. We are given
$(\bfX_t)_{t \geq 0}$ such that
\begin{equation}
  \label{eq:sde}
  \rmd \bfX_t = b(\bfX_t) \rmd t + \sigma(t) \rmd \bfB_t , 
\end{equation}
where $(\bfB_t)_{t \geq 0}$ is a $d$-dimensional Brownian motion. We also assume
that $b$ and $\sigma$ are Lipschitz which implies the existence and strong
uniqueness of $(\bfX_t)_{t \geq 0}$. We have the following result
\cite{lions1984stochastic}.

\begin{theorem}
  \label{thm:existence_reflected_sto}
  There
  exists a unique process $(\bar{\bfX}_t, \bfk_t)_{t \geq 0}$ such that
  \begin{enumerate}[label=(\alph*)]
  \item $(\bfk_t)_{t \geq 0}$ has bounded variation over $\coint{0,+\infty}$ almost surely.
  \item $(\bar{\bfX}_t)_{t \geq 0} \in \rmc(\coint{0,+\infty}, \cl{\M})$.
  \item For any $t \geq 0$, $\bar{\bfX}_t = \bar{\bfX}_0 + \int_0^t b(\bar{\bfX}_s) \rmd s + \int_0^t \sigma(\bar{\bfX}_s) \rmd \bfB_s - \bfk_t$. \label{item:sde_ref}
  \item For any $t \geq 0$,
    $\abs{\bfk}_t = \int_0^t \mathbf{1}_{\bar{x}_s \in \partial \M}(\bar{x}_s) \rmd \abs{\bfk}_s$ and
    $\bfk_t = \int_0^t n(\bar{x}_s) \rmd \abs{\bfk}_s$.
  \end{enumerate}
\end{theorem}

The process $(\bfX_t)_{t \geq 0}$ is almost surely continuous, so we could 
apply the previous theorem almost surely for all the realizations of the
process,. However, this does not tell us if the obtained solutions
$(\bar{\bfX}_t, \bfk_t)_{t \geq 0}$ form themselves a process. The main
difference with \Cref{thm:existence_reflected_det} is in
\Cref{thm:existence_reflected_sto}-\ref{item:sde_ref} which differs from
\Cref{thm:existence_reflected_sto}-\ref{item:decomp}. Note that in the case
where $b=0$ and $\sigma =\Id$ we recover
\Cref{thm:existence_reflected_sto}-\ref{item:decomp}. This is not true in the
general case. However, it can be seen that for any realization of the process
$(\bar{\bfX}_t)_{t \geq 0}$, we have that $(\bar{\bfX}_t, \bfk_t)_{t \geq 0}$ is
solution of the \emph{deterministic} Skorokhod problem by letting
$x_t = \bar{\bfX}_0 + \int_0^t b(\bar{\bfX}_s) \rmd s + \int_0^t
\sigma(\bar{\bfX}_s) \rmd \bfB_s$. The backward and forward Kolmogorov equations
can be found in \cite{burdzy2004heat}. Note that the presence of the process
$(\bfk_t)_{t \geq 0}$ incurs notable complications compared to unconstrained
processes. In particular, there is no martingale problem associated with weak
solutions of reflected SDEs but only sub-martingale problems, see
\cite{kang2017submartingale} for instance.


%% file: appendices/proof_ism.tex
\section{Implicit Score Matching Loss}
\label{sec:impl-score-match}

\subsection{Proof of ISM}
\label{app:proof_for_ism}

Using the divergence theorem, we have 
\begin{align}
  \label{eq:score_matching_app}
  &\textstyle{(1/2) \int_\M \| \mathbf{s}_\theta(x) - \nabla \log p_t(x) \|^2 p_t(x) \rmd \mu(x)} \\
  &= \textstyle{(1/2) \int_\M \| \mathbf{s}_\theta(x) \|^2 p_t(x) \rmd \mu(x)  - \int_\M \langle \mathbf{s}_\theta(x), \nabla \log p_t(x) \rangle p_t(x) \rmd \mu(x) +   (1/2) \int_\M \| \nabla \log p_t(x) \|^2 p_t(x) \rmd \mu(x)} \\
  &= \textstyle{(1/2) \int_\M \| \mathbf{s}_\theta(x) \|^2  p_t(x) \rmd \mu(x) -\int_{\partial \M} \langle \mathbf{s}_\theta(x), \mathbf{n} \rangle  p_t(x) \rmd \nu(x)} \\
  & \qquad \textstyle{ + \int_\M \mathrm{div}(\mathbf{s}_\theta)(x) p_t(x) \rmd \mu(x) +   (1/2) \int_\M \| \nabla \log p_t(x) \|^2 p_t(x) \rmd \mu(x)}. 
\end{align}
Using that $\mathbf{s}_\theta(x) = 0$ for all $x \in \partial\M$, we get that
\begin{align}
  &\textstyle{(1/2) \int_\M \| \mathbf{s}_\theta(x) - \nabla \log p_t(x) \|^2 p_t(x) \rmd \mu(x)} \\
  & \qquad \qquad = \textstyle{(1/2) \int_\M \| \mathbf{s}_\theta(x) \|^2  p_t(x) \rmd \mu(x)} 
   \textstyle{ + \int_\M \mathrm{div}(\mathbf{s}_\theta)(x) p_t(x) \rmd \mu(x) +   (1/2) \int_\M \| \nabla \log p_t(x) \|^2 p_t(x) \rmd \mu(x)} ,
\end{align}
which concludes the proof.

\subsection{Importance of scaling function}\label{sec:boundary_score}

\begin{figure}[H]
    \centering
    \includegraphics[width=0.4\textwidth]{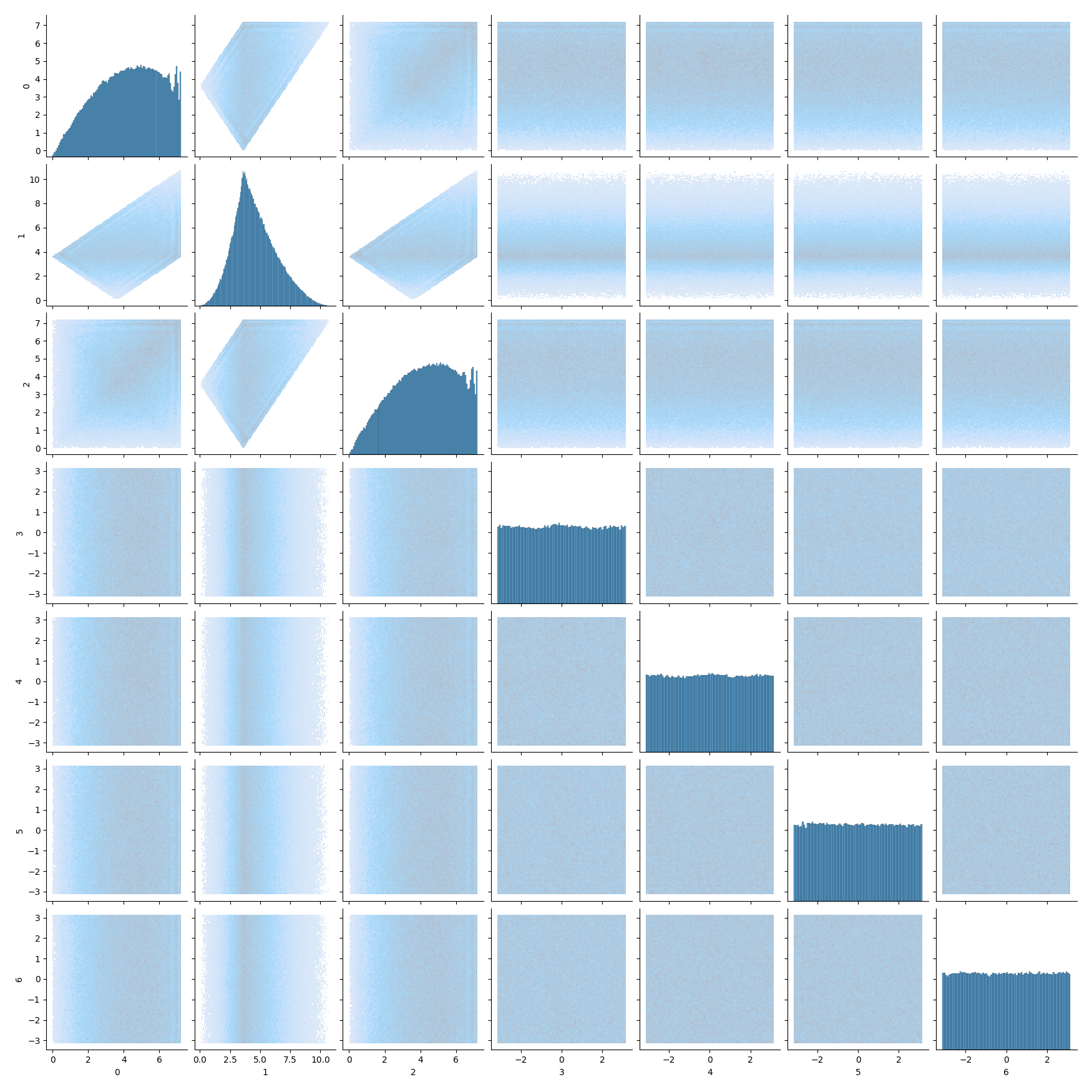}
    \includegraphics[width=0.4\textwidth]{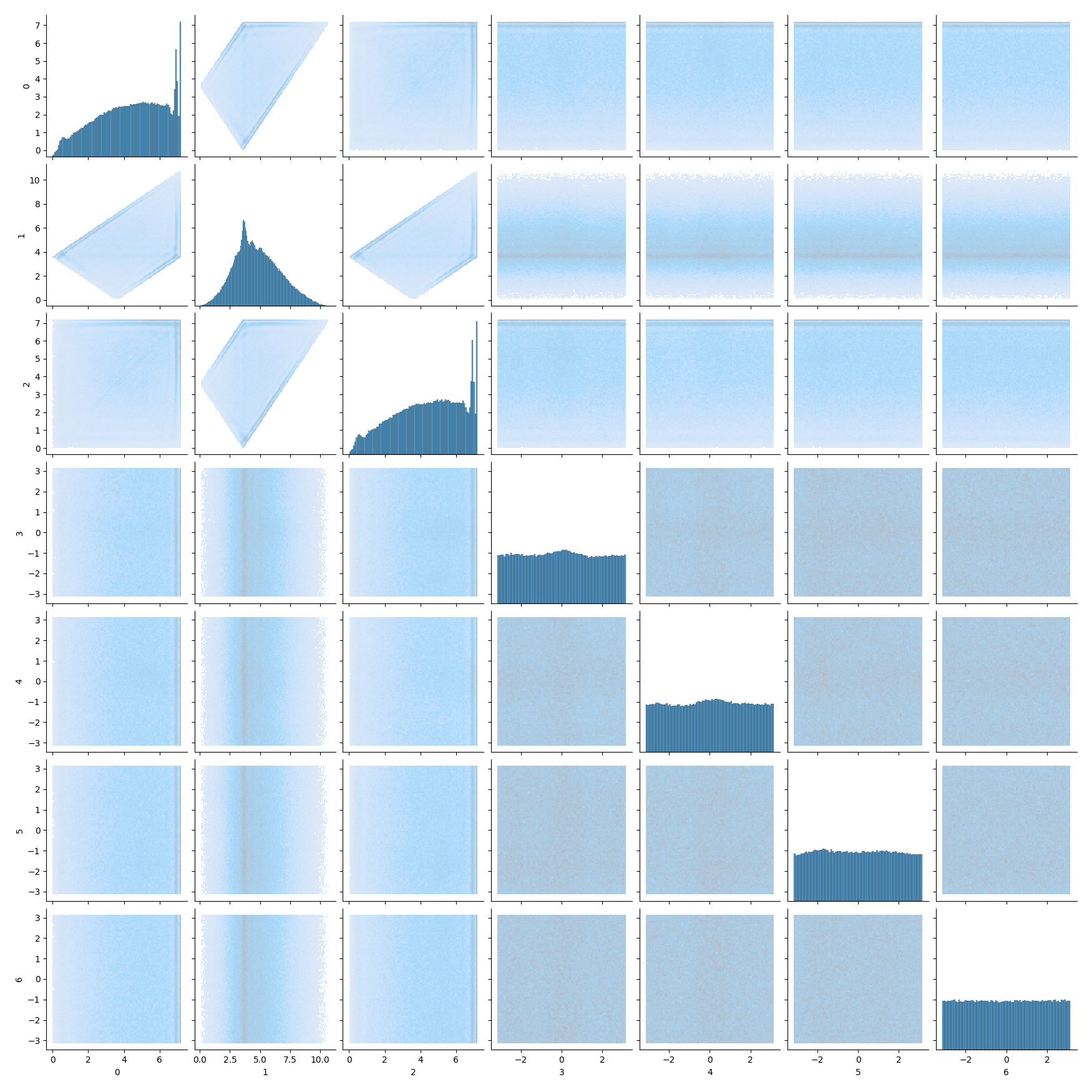}
    \caption{Reverse process samples for the cyclic peptide dataset from~\Cref{subsec:protein_loops} at $t=1.0,0.9$ (left and right respectively) trained without the scaling function.}
    \label{fig:no_scaling}
\end{figure}

\begin{figure}[H]
    \centering
    \includegraphics[width=0.4\textwidth]{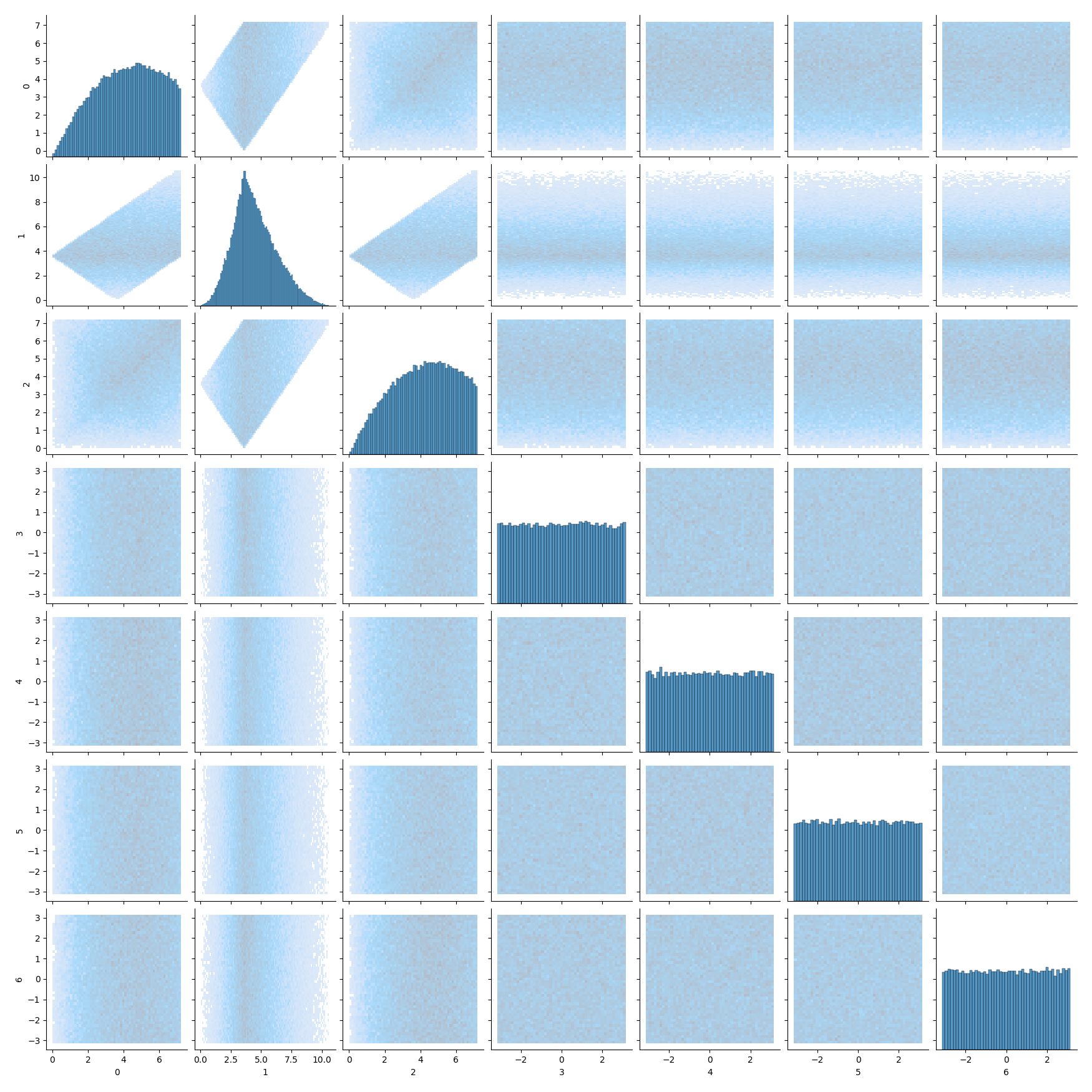}
    \includegraphics[width=0.4\textwidth]{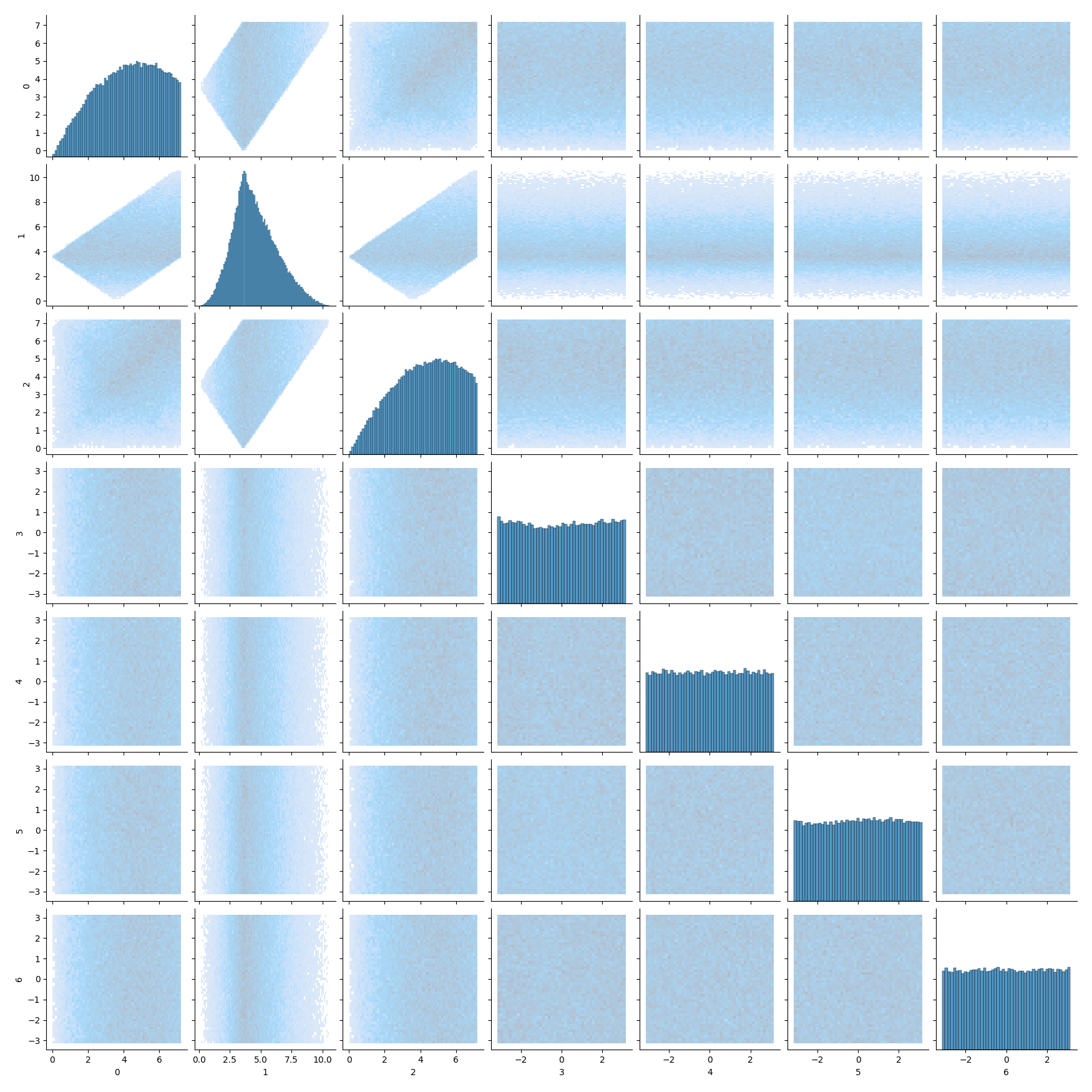}
    \caption{Reverse process samples for the cyclic peptide dataset from~\Cref{subsec:protein_loops} at $t=1.0,0.9$ (left and right respectively) trained with the scaling function.}
    \label{fig:scaling}
\end{figure}

As discussed in \Cref{sec:score-netw-param}, we include a monotone scaling function $h$ which is zero close to the boundary to ensure the relevant conditions are met for the score matching loss and the boundary conditions. This may seem like a technical detail, but is of significant practical importance. Upon removal of the scaling function, we observe that the learned score functions behave strangely around the boundary in the reverse process, leading to samples that do not match the forward process.
The problems are apparent when comparing the top three plots of Fig. \ref{fig:no_scaling} and Fig. \ref{fig:scaling}. Interestingly, we found that these issues early on in the sampling are smoothed out by the end of the reverse process, but still lead to a failure to recover the target density.

%% file: appendices/likelihood.tex
\section{Likelihood evaluation}
\label{sec:likel-eval}

One key advantage of constructing a continuous noising process is that, similarly to \citet{song2020score}, 
we can evaluate the model's likelihood via the following \emph{probability flow} Ordinary Differential Equation (ODE).
In particular, for the Langevin dynamics (\ref{eq:langevin_hessian_manifold}) 
which we recall
\begin{equation}
  \textstyle{
    \rmd \bfX_t = \tfrac{1}{2} \mathrm{div}(\metric^{-1})(\bfX_t)  \rmd t + \metric(\bfX_t)^{-\frac{1}{2}} \rmd \bfB_t,
    }
\end{equation}
the following ODE has the same marginal density
\begin{equation}
 \label{eq:probability_flow_hessian_manifold}
 \textstyle{
    \rmd {\bfY}_t = \left[ \tfrac{1}{2} \nabla \cdot \metric^{-1}(\bfY_t) - \tfrac{1}{2} \metric^{-1}(\bfY_t) \nabla \log p_t(\bfY_t) \right] \rmd t .
    }
\end{equation}


We conclude this section with a derivation of the equivalent ODE. We highlight
that the ODE representation for reflected diffusion models was first derived in
\cite{lou2023reflected}. We recall that if $\M \subset \rset^d$ is a bounded
open set with smooth boundary then \cite[Theorem 2.2]{burdzy2004heat} ensures
that the reflected Brownian motion admits a density w.r.t. the Lebesgue
measure. We denote $p_t$ this smooth density.

  \begin{proposition}
    Assume that $\M \subset \rset^d$ is a bounded open set with smooth boundary.
    Assume that $(t,x) \mapsto \nabla \log p_t(x)$ is smooth on
    $\coint{0,+\infty} \times \partial \M$.  Let $(\bar{\bfB}_t)_{t \geq 0}$ be
    the reflected Brownian motion with $\bar{\bfB}_0 \sim p_0$ smooth and
    supported in $\M$. Let $(\bfX_t)_{t \geq 0}$ be given for any $t \geq 0$ by
    $\rmd \bfX_t = \tfrac{1}{2} \nabla \log p_t(\bfX_t) \rmd t$ and
    $\bfX_0 \sim p_0$, where $p_t$ denotes the density of $\bar{\bfB}_t$ w.r.t. the
    Lebesgue measure for any $t > 0$. Then for any $t \in \ccint{0,T}$, $\bar{\mathbf{B}}_t$ and $\mathbf{Y}_t$ have the same distribution.
  \end{proposition}

  \begin{proof}
  Since the distributions of $(\bar{\mathbf{B}}_t)_{t \in \ccint{0,T}}$ and $(\mathbf{Y}_t)_{t \in \ccint{0,T}}$ satisfy the same Fokker-Planck equation whenever these processes are well-defined. Therefore, we first show that the process $(\mathbf{Y}_t)_{t \in \ccint{0,T}}$ is well-defined and stay in $\mathcal{M}$ at all times. 
    Using \cite[Theorem 2.2]{burdzy2004heat}, we have that
    $\partial p_t(x) = \tfrac{1}{2} \mathrm{div}(\nabla \log p_t)(x)$, for any
    $t > 0$ and $x \in \M$. Next, we define
    $\rmd \bfX_t = \tfrac{1}{2} \nabla \log p_t(\bfX_t) \rmd t$. Note that
    $(\bfX_t)_{t \geq 0}$ is defined up to an explosion time $T_\infty$ after
    which we fix $\bfX_t = \infty$. Denote $T_0$ the first time such that
    $\bfX_t \in \partial \M$. Note that since $p_0$ is supported on $\M$ we have
    $T_0 > 0$. We denote
    $(\bfY_t)_{t \in \ccint{0,T_0}} = (\bfX_{T_0-t})_{t \in \ccint{0,T_0}}$. We
    have that for any $t \in \ccint{0,T_0}$,
    $\rmd \bfY_t = -\tfrac{1}{2} \nabla \log p_{T_0-t}(\bfY_t) \rmd t$. Since
    $(t,x) \mapsto \nabla \log p_t(x)$ is smooth on
    $\coint{0,+\infty} \times \partial \M$, we get that for any
    $t \in \ccint{0,T_0}$, $\bfY_t \in \partial \M$. In particular, we have that
    $\bfY_{T_0/2} = \bfX_{T_0/2} \in \partial \M$ which is absurd. Therefore
    $T_0 = +\infty$ (which also implies that $T_\infty = +\infty$). Hence,
    $(\bfX_t)_{t \geq 0}$ is a flow on $\M$ and therefore for any $t \geq 0$,
    the density $q_t$ of $\bfX_t$ is smooth and satisfies
    $\partial_t q_t(x) = -\tfrac{1}{2} \mathrm{div}(q_t \nabla \log p_t
    )(x)$. We conclude using the uniqueness of the solutions to the transport
    equation for smooth initialisation and coefficients on $\rset^d$.
  \end{proof}


%% file: appendices/proof_time_reversal.tex
\section{Time-reversal for reflected Brownian motion}
\label{sec:time-revers-refl}

We start with the following definition.

\begin{definition}
  Let $\M \subset \rset^d$ be an open set. $\M$ has a smooth boundary if for any
  $x \in \partial \M$, there exists $\msu \subset \rset^d$ open and
  $f \in \rmc^\infty(\msu, \rset)$ such that $x \in \msu$ and
  \begin{enumerate*}[label=(\alph*)]  
  \item $\mathrm{cl}(\M) \cap \msu = \ensembleLigne{x \in \msu}{f(x) \leq 0}$,
  \item $\nabla f(x) \neq 0$ for any $x \in \msu$ 
  \end{enumerate*}
  where $\mathrm{cl}(\M)$ is the closure of $\M$.
\end{definition}

We will make the use of the following lemma which is a straightforward extension
of \citet[Theorem 2.6]{burdzy2004heat}. The surface measure is defined in \cite[Proposition 2.43]{lee2006riemannian}. Under mild regularity assumptions, it corresponds to the Hausdorff measure of $\partial \M$, see \cite{evans2015measure}.

\begin{lemma}
  \label{lemma:formula_integration}
  Let $u$ such that $s \mapsto u(s,x) \in \rmc^1(\ooint{0,T}, \rset)$, for any
  $s \in \ooint{0,T}$, $x \mapsto u(s,x) \in \rmc^2(\mathrm{\M}, \rset)$ and
  $u \in \rmc^1(\mathrm{cl}(\mathrm{\M}), \rset)$. Then for any $T \geq 0$, $s, t \in \ccint{0,T}$,
  we have
  \begin{equation}
    \textstyle{\expeLigne{\int_s^t u(w, \bar{\bfB}_w) \rmd \bfk_w } = \tfrac{1}{2} \int_s^t \int_{\partial \M} u(x) p_w(x) \rmd \sigma(x) \rmd w .}
  \end{equation}
\end{lemma}

Note that we recover \citet[Theorem 2.6]{burdzy2004heat} if we set $u=1$.  We
also emphasize that the result of \citet[Theorem 2.6]{burdzy2004heat} is stronger than \Cref{lemma:formula_integration} as it
holds not only in expectation but also in $\mathrm{L}^2$ and almost surely.

We are now ready to
prove \Cref{sec:time-reversal-reflected}. We follow the approach of
\cite{petit1997Time} which itself is based on an extension of
\cite{haussmann1986time}. We refer to \citet{cattiaux2021time} for recent
entropic approaches of time-reversal.  Recall that
$(\bar{\bfB}_t, \bfk_t)_{t \geq 0}$ is a solution to the \emph{Skorokhod
  problem} \citep{skorokhod1961stochastic} if $(\bfk_t)_{t \geq 0}$ a bounded
variation process and $(\bar{\bfB}_t)_{t \geq 0}$ a continuous adapted process
such that for any $t \geq 0$, $\bfB_t = \bar{\bfB}_t + \bfk_t \in \M$,
$(\bar{\bfB}_t)_{t \geq 0}$ and
\begin{equation}
  \label{eq:skorokhod_app}
  \textstyle{\abs{\bfk}_t = \int_0^t \mathbf{1}_{\bar{\bfB}_s \in \partial \M} \rmd \abs{\bfk}_s , \quad \bfk_t = \int_0^t \bfn(\bar{\bfB}_s) \rmd \abs{\bfk}_s ,}
\end{equation}
In what follows, we define $(\bfY_t)_{t \in \ccint{0,T}}$ such that for any
$t \in \ccint{0,T}$, $\bfY_t = \bar{\bfB}_{T-t}$. Let us consider the process
$(\tilde{\bfB}_t)_{t \in \ccint{0,T}}$ defined for any $t \in \ccint{0,T}$ by
\begin{equation}
  \textstyle{\tilde{\bfB}_t = -\bar{\bfB}_T + \bar{\bfB}_{T-t} + \bfk_T - \bfk_{T-t} - \int_{T-t}^T \nabla \log p_s(\bar{\bfB}_s) \rmd s .}
\end{equation}
First, note that $t \mapsto \tilde{\bfB}_t$ is continuous. Denote by
$\mathcal{F}$, the filtration associated with
$(\bar{\bfB}_{T-t})_{t \in \ccint{0,T}}$. We have that
$(\tilde{\bfB}_t)_{t \in \ccint{0,T}}$ is adapted to
$(\bar{\bfB}_{T-t})_{t \in \ccint{0,T}}$. Even more so, we have that
$(\tilde{\bfB}_t)_{t \in \ccint{0,T}}$ satisfies the strong Markov property
since $(\bar{\bfB}_t)_{t \in \ccint{0,T}}$ also satisfies the strong Markov
property. Let $g \in \rmc_c^\infty(\mathrm{cl}(\M))$ and consider for any $0 \leq s \leq t \leq T$,
$\expeLigne{(\tilde{\bfB}_t - \tilde{\bfB}_s)g(\bar{\bfB}_{T-t})}$.  For any
$0 \leq s \leq t \leq T$ we have
\begin{equation}
  \label{eq:first_equality}
  \expeLigne{(\tilde{\bfB}_t - \tilde{\bfB}_s)g(\bar{\bfB}_{T-t})} = \textstyle{\expeLigne{(-\bar{\bfB}_{T-s} + \bar{\bfB}_{T-t} + \bfk_{T-s} - \bfk_{T-t} - \int_{T-t}^{T-s} \nabla \log p_u(\bar{\bfB}_u) \rmd u) g(\bar{\bfB}_{T-t})} .}
\end{equation}
In what follows, we prove that for any
$0 \leq s \leq t \leq T$ we have $\expeLigne{(\tilde{\bfB}_t  - \tilde{\bfB}_s)g(\bar{\bfB}_{T-t})} = 0$. Therefore, we only need to prove that for any
$0 \leq s \leq t \leq T$ we have
\begin{equation}
  \label{eq:second_equality}
  \textstyle{\expeLigne{(-\bar{\bfB}_{t} + \bar{\bfB}_{s} + \bfk_t - \bfk_{s} - \int_{s}^{t} \nabla \log p_u(\bar{\bfB}_u) \rmd u) g(\bar{\bfB}_{t})}=0 .}
\end{equation}
Let $t \in \ocint{0,T}$. We introduce $u: \ \ccint{0,t} \times \M$ such that for
any $s \in \ccint{0,t}$ and $x \in \M$,
$u(s, x) = \CPELigne{g(\bar{\bfB}_t)}{\bar{\bfB}_s=x}$. Using \citet[Theorem
2.8]{burdzy2004heat} we get that for any $x \in \M$,
$s \mapsto u(s,x) \in \rmc^1(\ooint{0,t}, \rset)$ and for any
$s \in \ooint{0,t}$, $x \mapsto u(s,x) \in \rmc^2(\mathrm{\M}, \rset)$ and
$x \mapsto u(s,x) \in \rmc^1(\mathrm{cl}(\M), \rset)$. In
addition, we have that for any $s \in \ooint{0,t}$ and for any $x \in \M$ and
$x_0 \in \partial \M$
\begin{equation}
  \label{eq:backward_kolmogorov}
  \textstyle{\partial_s u(s,x) + \tfrac{1}{2} \Delta u(s,x) = 0 , \qquad \langle \nabla u(s, x_0), \bfn(x_0) \rangle = 0 . }
\end{equation}
This equation is called the backward Kolmogorov equation. Using
\eqref{eq:backward_kolmogorov}, $\bar{\bfB}_t = \bfB_t - \bfk_t$ for any
$t \geq 0$ and the It\^o formula for semimartingale \citep[Chapter IV, Theorem
3.3]{revuz2013continuous} we have that for any $s \in \ooint{0,t}$
\begin{align}
  \expeLigne{u(t, \bar{\bfB}_t)\bar{\bfB}_t} &\textstyle{= \expeLigne{u(s, \bar{\bfB}_s)\bar{\bfB}_s} + \expeLigne{\tfrac{1}{2} \int_s^t \bar{\bfB}_w \Delta u(w, \bar{\bfB}_w) \rmd w}} \textstyle{+\expeLigne{\int_s^t \nabla u(w, \bar{\bfB}_w) \rmd w }} \\ 
                                             & \qquad \textstyle{- \expeLigne{\int_s^t \bar{\bfB}_w \langle \nabla u(w, \bar{\bfB}_w), \bfn(\bar{\bfB}_w) \rangle \rmd \abs{\bfk}_w}} \\
   & \qquad \textstyle{- \expeLigne{\int_s^t u(w,\bar{\bfB}_w) \bfn(\bar{\bfB}_w) \rmd \abs{\bfk}_w}} \\
  & \qquad \textstyle{+ \expeLigne{\int_s^t \bar{\bfB}_w \partial_w u(w, \bar{\bfB}_w) \rmd w} } \\
  &\textstyle{= \expeLigne{u(s, \bar{\bfB}_s)\bar{\bfB}_s} \textstyle{+\expeLigne{\int_s^t \nabla u(w, \bar{\bfB}_w) \rmd w }}} \textstyle{- \expeLigne{\int_s^t u(w,\bar{\bfB}_w) \bfn(\bar{\bfB}_w) \rmd \abs{\bfk}_w}} \label{eq:intermediate}
\end{align}
In addition, using the Fubini theorem and the definition of $\bfk_t$ we have that for any $s \in \ooint{0,t}$
\begin{equation}
  \textstyle{\expeLigne{\int_s^t u(w,\bar{\bfB}_w) \bfn(\bar{\bfB}_w) \rmd \abs{\bfk}_w} = \expeLigne{\int_s^t \CPELigne{g(\bar{\bfB}_t)}{\bar{\bfB}_w} \bfn(\bar{\bfB}_w) \rmd \abs{\bfk}_w} = \expeLigne{g(\bar{\bfB}_t)(\bfk_t - \bfk_s)} .} \label{eq:boundary_uno}
\end{equation}
Finally, using the divergence theorem and \citet[Theorem 2.6]{burdzy2004heat} we
have that for any $s \in \ooint{0,t}$
\begin{align}
  \textstyle{\expeLigne{\int_s^t \nabla u(w, \bar{\bfB}_w) \rmd w }} &= \textstyle{\int_s^t \int_\M \nabla u(w, x) p_w(x) \rmd  x \rmd w } \\
                                                                     &= - \textstyle{\int_s^t \int_\M  u(w, x) \nabla \log (p_w(x)) p_w(x) \rmd  x \rmd w + \int_s^t \int_{\partial \M} u(w,x) p_w(x) \rmd x \rmd \sigma(w)} ,
\end{align}
where $\sigma$ is the surface area measure on $\partial \M$, see
\citet{burdzy2004heat}. Using \Cref{lemma:formula_integration} and the Fubini
theorem we get that
\begin{align}  
  \textstyle{\expeLigne{\int_s^t \nabla u(w, \bar{\bfB}_w) \rmd w }} &= - \textstyle{\int_s^t \int_\M  u(w, x) \nabla \log (p_w(x)) p_w(x) \rmd  x \rmd w + \expeLigne{\int_s^t u(w, \bar{\bfB}_w) \rmd \bfk_w } }  \\
  &= - \textstyle{\expeLigne{\int_s^t g(\bar{\bfB}_t) \nabla \log (p_w(\bar{\bfB}_w))\rmd w} + 2 \expeLigne{g(\bar{\bfB}_t)(\bfk_t - \bfk_s) }} \label{eq:boundary_duo}
\end{align}
Combining \eqref{eq:intermediate}, \eqref{eq:boundary_uno} and
\eqref{eq:boundary_duo} we get that
\begin{equation}
  \textstyle{ \expeLigne{u(t, \bar{\bfB}_t) } = \expeLigne{u(s, \bar{\bfB}_s) } - \expeLigne{ g(\bar{\bfB}_t)\int_s^t \nabla \log p_w(\bar{\bfB}_w) \rmd w  }} + \textstyle{ \expeLigne{g(\bar{\bfB}_t)(\bfk_t - \bfk_s) }} .
\end{equation}
Therefore, we get \eqref{eq:second_equality} and \eqref{eq:first_equality}.
Hence, $(\tilde{\bfB}_t)_{t \in \ccint{0,T}}$ is a continuous martingale. In
addition, we have that for any $t \in \ccint{0,T}$,
$\expeLigne{\tilde{\bfB}_t \tilde{\bfB}_t^\top} = t \Id$ and therefore,
$(\tilde{\bfB}_t)_{t \in \ccint{0,T}}$ is a Brownian motion using the L\'evy
characterisation of Brownian motion \citep[Chapter IV, Theorem
3.6]{revuz2013continuous}. Denote
$(\bfj_t)_{t \in \ccint{0,T}} = (\bfk_T - \bfk_{T-t})_{t \in
  \ccint{0,T}}$. Using \eqref{eq:first_equality}, we have that for any
$t \in \ccint{0,T}$
\begin{equation}
  \textstyle{\bar{\bfB}_{T-t} = \bar{\bfY}_0 + \tilde{\bfB}_t + \int_0^t \nabla \log p_{T-s}(\bfY_s) \rmd s  - \bfj_t . }
\end{equation}
Using \eqref{eq:skorokhod_app}, we have for any $t \in \ccint{0,T}$
\begin{equation}
    \textstyle{\abs{\bfj}_t = \int_0^t \mathbf{1}_{\bfY_s \in \partial \M} \rmd \abs{\bfj}_s , \quad \bfj_t = \int_0^t \bfn(\bar{\bfY}_s) \rmd \abs{\bfj}_s ,}
  \end{equation}
  which concludes the proof.





%% file: appendices/application_psd.tex
\section{Configurational modelling of robotic arms under manipulability constraints}
\label{sec:robotic_arm}

Accurately determining and specifying the movement of a robotic arm and the forces it exerts is a fundamental problem in many real-world robotics applications. A widely-used set of descriptors for modelling the flexibility of a given joint configuration are so-called manipulability ellipsoids 
\citep{yoshikawa1985manipulability}, which are kinetostatic performance measures that quantify the ability to move or exert forces along different directions. 
\citet{jaquier2021geometry} present a geometric framework to learn trajectories of manipulability ellipsoids by making use of the fact any ellipsoid $M\in\mathbb{R}^N$ is defined by the set of points $\{\mathbf{x}\vert\mathbf{x}^T\mathbf{A}\mathbf{x}=1\}$ where $\mathbf{A}$ lies on the manifold of $N\times N$ symmetric positive definite matrices $S^N_{++}$.

In many practical settings, it is desirable to constrain the minimal or maximal volume of a manipulability ellipsoid to retain motional flexibility or limit the magnitude of the exerted force. This necessitates lower or upper limits on the determinant of $\textbf{A}$, translating into constraints on $S^N_{++}$.
To model this, we make use of one of the datasets introduced by \citet{jaquier2021geometry}, containing demonstrations of a robotic arm drawing different letters in the plane, providing the respective positional trajectories ($\mathbb{R}^2$) and velocity manipulability ellipsoids ($S^2_{++}$).

\begin{figure}[H]
\centering
\begin{subfigure}{0.4\textwidth}
  \centering
  \includegraphics[width=\linewidth]{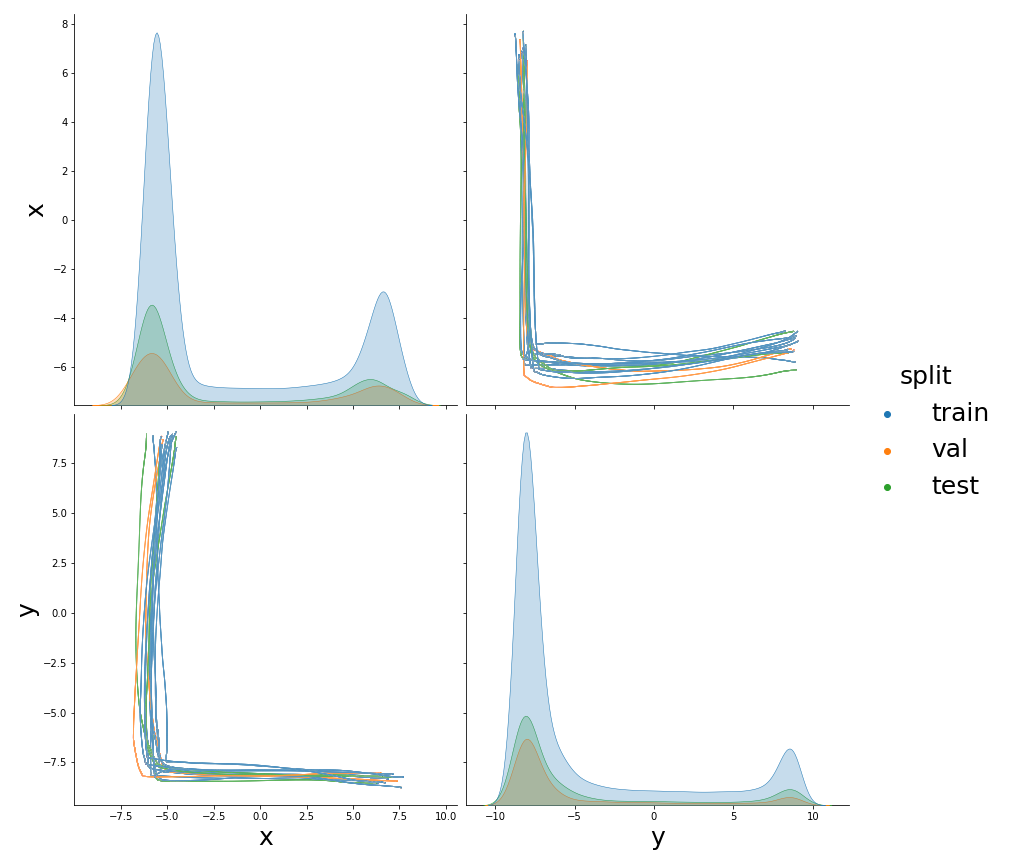}
  \caption{}
  \label{fig:spd_L_r2}
\end{subfigure}
\hspace{1.5cm}
\begin{subfigure}{0.4\textwidth}
  \centering
  \includegraphics[width=\linewidth]{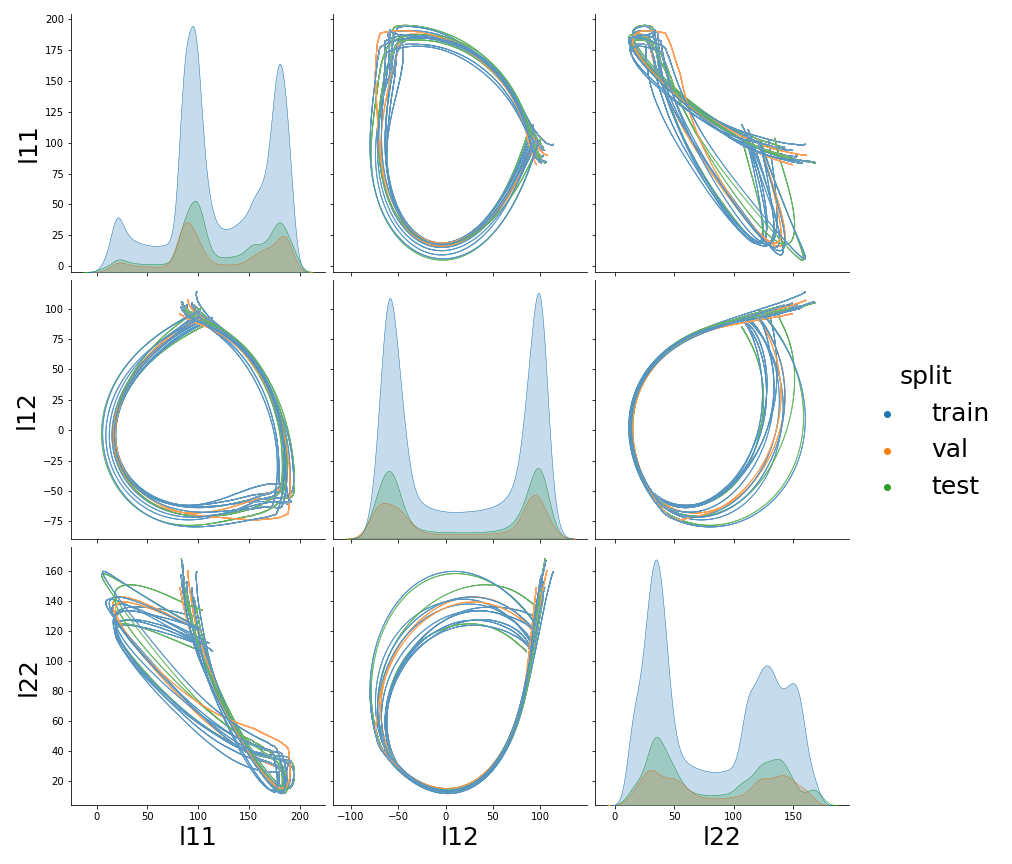}
  \caption{}
  \label{fig:spd_L_ell}
\end{subfigure}
\caption{Positional trajectories $\mathbf{x}\in\mathbb{R}^2$ (a) and the parameters $l_{11}, l_{12}, l_{22}$ of the the SPD matrix $\mathbf{A}\in S^2_{++}$ (b) for the letter \textsc{L}.}
\label{fig:spd_L}
\end{figure}

We use the processing routines provided by \citet{jaquier2021geometry} to interpolate the trajectories into $10^4$ distinct points, for each of which we derive the position $\mathbf{x}\in\mathbb{R}^2$ and the PSD matrix 
$$
\mathbf{A}=\begin{pmatrix}
a & b\\
b & c
\end{pmatrix}\in S^2_{++}.
$$
parametrising the velocity manipulability ellipsoid $M\in\mathbb{R}^2$. The resulting data is split into training, validation, and test sets by trajectory and visualised in~\Cref{fig:spd_L}. 
We add a small amount of Gaussian noise to these trajectories, which is shown as the target distribution in \ref{sec:app_spd_exp}.


%% file: appendices/application_loop.tex
\section{Conformational modelling of polypeptide backbones under end point constraints}
\label{sec:protein_background}

Polypeptides and proteins constitute an important class of biogenic macromolecules that underpin most aspects of organic life. Accurately modelling their conformational ensembles, i.e. the set of three-dimensional structures they assume under physiological conditions, is essential to both understanding the biological function of existing and designing the enzymatic or therapeutic properties of novel proteins \citep{lane2023protein}.
Motivated by the success of diffusion models in computer vision and natural language processing, there has been considerable interest in applying them to learn and sample from distributions over the conformational space of protein structures \citep{watson2022Broadly, trippe2022Diffusion, wu2022protein}.

\subsection{Problem parameterisation}
\label{sec:protein_param}

\begin{figure}[h]
\centering
\begin{subfigure}{0.45\textwidth}
  \centering
  \includegraphics[width=\linewidth]{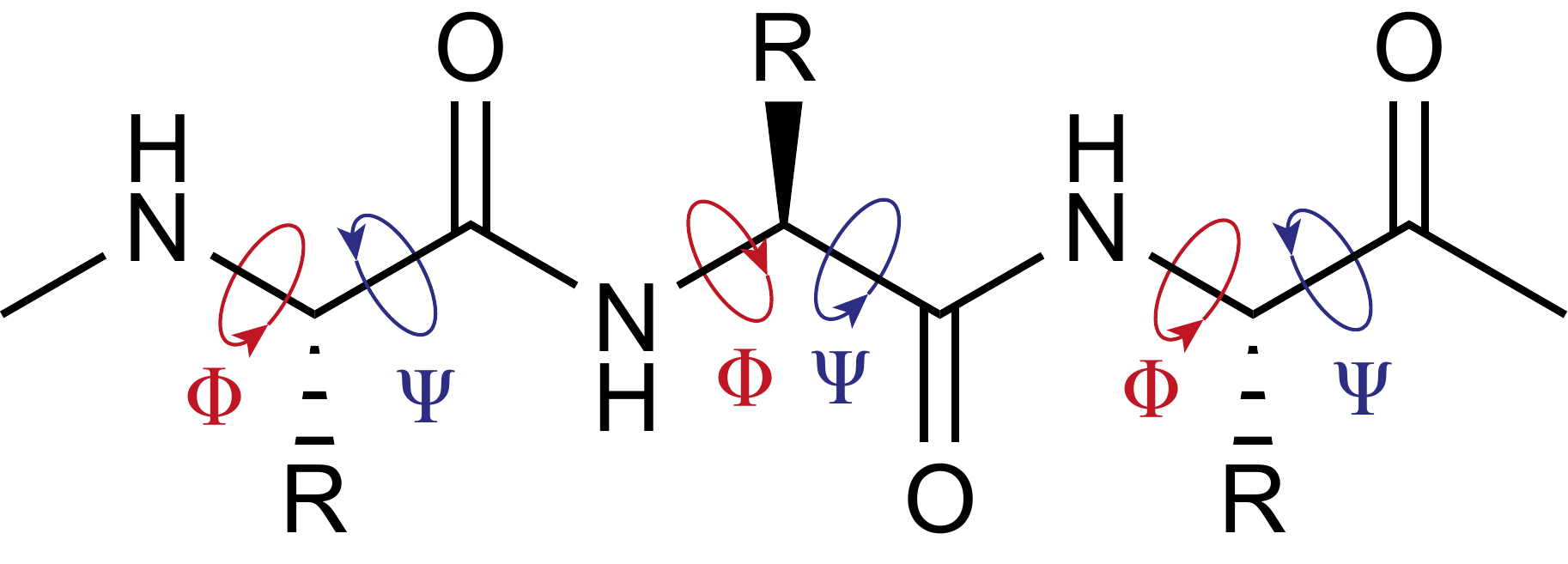}
  \caption{A commonly-used approximate parameterisation of backbone geometry only considers the $C_\alpha$ torsion angles $\Phi$ and $\Psi$.}
  \label{fig:peptide_chain}
\end{subfigure}
\hspace{0.5cm}
\begin{subfigure}{0.45\textwidth}
  \centering
  \includegraphics[width=\linewidth]{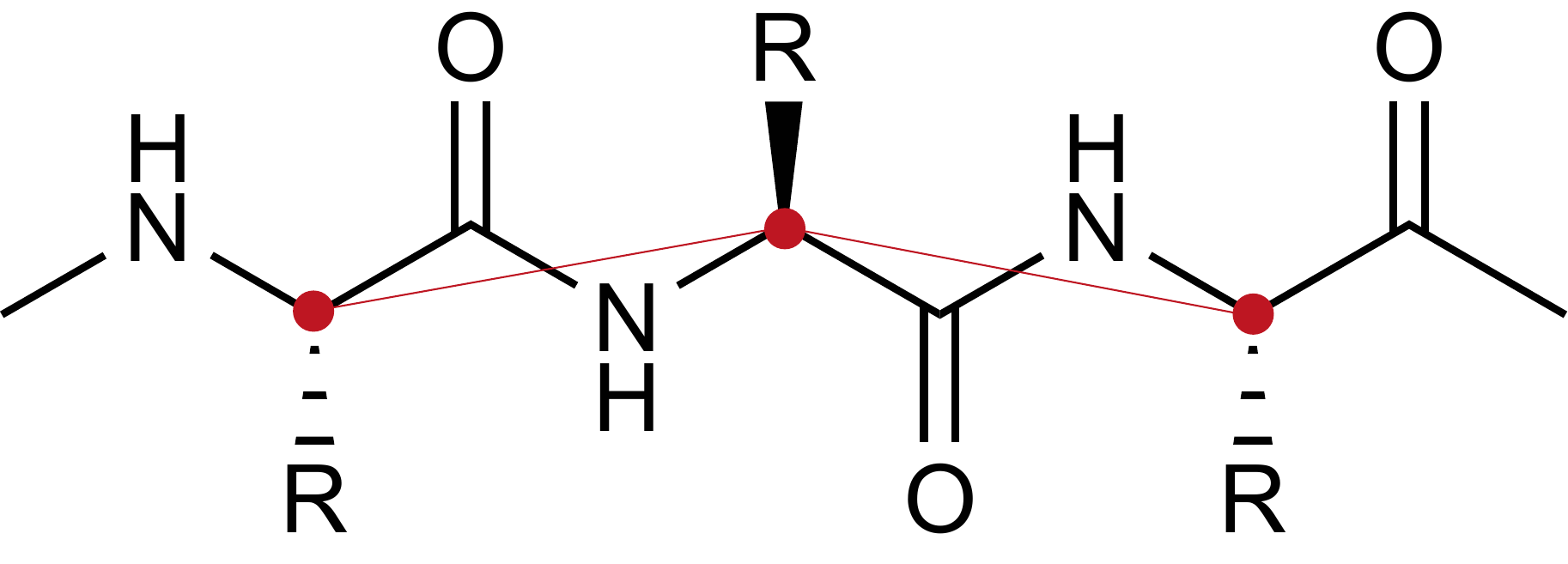}
  \caption{As peptide bond orientations can be inferred relatively reliably, researchers often only model the $C_\alpha$ traces.}
  \label{fig:peptide_trace}
\end{subfigure}
\caption{Standard approaches to modelling the conformations of polypeptide backbones.}
\label{fig:peptide}
\end{figure}

Proteins are biopolymers in which a sequence of $N$ amino acids is joined together through $N-1$ peptide bonds, resulting in a so-called polypeptide backbone with protruding amino acid residues. As the deviation of chemical bond lengths and angles from their theoretical optimum is generally negligible, the problem of modelling the three-dimensional structure of this polypeptide chain is often reframed in the space of the internal torsion angles $\Phi$ and $\Psi$ (see~\Cref{fig:peptide_chain} for an illustration), which can be modelled on a $(2N-2)$-dimensional torus $\mathbb{T}^{2N-2}$.

In many data-scarce practical settings such as antibody or enzyme design, it is often unnecessary or even undesirable to model the structure of an entire protein, as researchers are primarily interested in specific functional sites with distinct biochemical properties.
However, generating conformational ensembles for such substructural elements necessitates positional constraints on their endpoints to ensure that they can be accommodated by the remaining scaffold. While it is conceivable that a diffusion model could derive such constraints from limited experimental data, we argue that it is much more efficient and precise to encode them explicitly.

\begin{figure}[h]
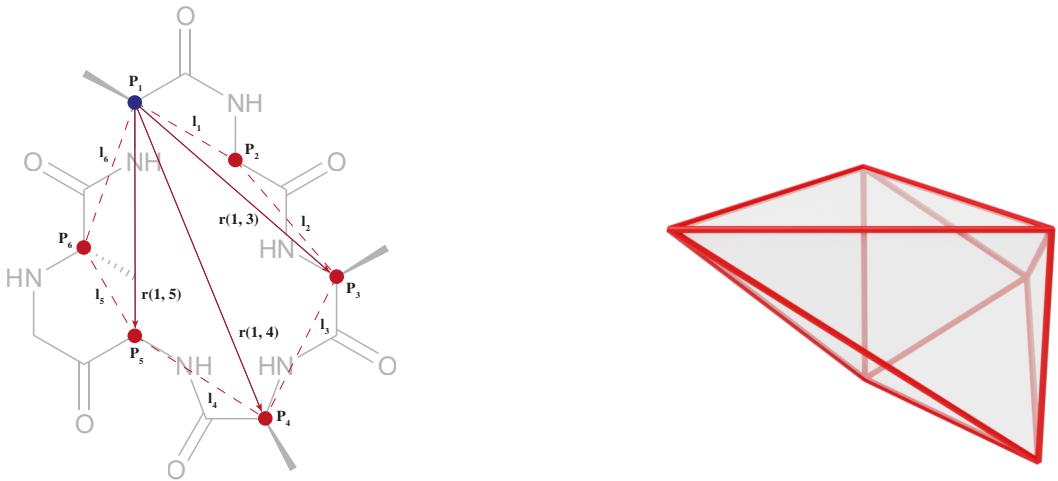

\centering
\begin{subfigure}[t]{0.45\textwidth}
  \centering
  \includegraphics[width=0.7\linewidth]{images/cyclic_peptide.pdf}
  \caption{An illustrative diagram of the proposed parameterisation for modelling the $C_\alpha$ trace geometry of the cyclic peptide \textsc{c-AAGAGG}.}
  \label{fig:cyclic_peptide}
\end{subfigure}
\hfill
\begin{subfigure}[t]{0.45\textwidth}
  \centering
  \includegraphics[width=\linewidth]{images/hull.png}
  \caption{The convex polytope constraining the diagonals of the triangles for the given bond lengths in the illustrated molecule. The total design space is the product of this polytope with the 4D flat torus.}
  \label{fig:convex_polytope_loop_illustration}
\end{subfigure}
\caption{Parameterising the conformational space of polypeptide backbones under anchor point distance constraints.}
\label{fig:han_and_rudolph_rep}
\end{figure}

For this purpose, we adopt the distance constraint formulation from \citet{han2006inverse} and interpret the backbone as a spatial chain with $N$ spherical joints and fixed-length links (see~\Cref{fig:cyclic_peptide} for an illustration). After selecting a suitable anchor point, the geometry of the polypeptide chain is fully specified by \begin{enumerate*}[label=(\alph*)] 
\item the set of link lengths $\mathbf{\ell}=\{\ell_j\}_{j=1}^{N}$, 
\item the set of vectors $\mathbf{r}=\{r(1,j)\}_{j=2}^{N}$ between the anchor point and each atom in the chain, and
\item the set of dihedral angles\end{enumerate*}
$$
\mathbf{T}=\left\{\arccos\left(\frac{\vert\langle r(1,j)\times r(1,j+1), r(1,j+1)\times r(1,j+2)\rangle\vert}{\vert r(1,j)\times r(1,j+1)\vert\vert r(1,j+1) r(1,j+2)\vert}\right)\right\}_{j=2}^{N-2}\in\mathbb{T}^{N-3}
$$ 
between any three consecutive vectors. After specifying the fixed bond lengths $\ell$, including an arbitrary anchor point distance $d_\text{anchor}=\ell_N=r(1,N)$, the set of valid vectors $\mathbf{r}$ is given by the convex polytope $\mathbb{P}\subseteq\mathbb{R}^3$ defined by the following linear constraints (see~\Cref{fig:convex_polytope_loop_illustration} for an illustration):


$$
\begin{array}{rcl}
r(1,3)  &\leq& \ell_1+\ell_2 ,\\
-r(1,3)  &\leq&-\left|\ell_1-\ell_2\right|, \\
\end{array}
$$
$$
\left.\begin{array}{rcl}
r(1, j)-r(1, j+1) &\leq& \ell_j, \\
-r(1, j)+r(1, j+1) &\leq& \ell_j, \\
-r(j)-r(j+1) &\leq&-\ell_j,
\end{array}\right\}3 \leq j \leq N-2, \\
$$
$$
\begin{array}{rcl}
r(1, N-1)  &\leq& \ell_{N-1}+d_\text{anchor}, \\
-r(1, N-1)  &\leq&-\left|\ell_{N-1}-d_\text{anchor}\right|.
\end{array}
$$

This means that the set of all valid polypeptide backbone conformations is defined by the product manifold $\mathbb{P}\times\mathbb{T}^{N-3}$, enabling us to train diffusion models that exclusively generate conformations with a fixed anchor point distance $d_\text{anchor}$.

\subsection{Data generation and model training}
\label{sec:protein_data}

As a proof-of-concept for the practicality of our methods, we chose to model the conformational distribution of the cyclic peptide \textsc{c-AAGAGG}. Cyclic peptides are an increasingly important drug modality with therapeutic uses ranging from antimicrobials to oncology, exhibiting circular polypeptide backbones (i.e. $d_\text{anchor}=0$) that confer a range of desirable pharmacodynamic and pharmacokinetic properties \citep{dougherty2019understanding}.
To reduce the dimensionality of the problem, we only consider the $C_\alpha$ traces (with fixed $C_\alpha$-$C_\alpha$ link distances of 3.6 \AA) instead of the full polypeptide backbone (see \cref{fig:peptide_trace}), although we note that our framework is fully applicable to both settings.

To derive a suitable dataset, the product manifold $\mathbb{P}\times\mathbb{T}^3$ describing the conformations of cyclic $C_\alpha$ traces of length $N=6$ was constructed (see~\Cref{fig:cyclic_peptide,fig:convex_polytope_loop_illustration} for an illustration) and used to generate $10^7$ uniform samples satisfying the anchor point distance constraint $d_\text{anchor}=0$. Subsequently, an estimate of the free energy $E_i$ of each sample $i$ was obtained by (1) reconstructing the full-atom peptide from each $C_\alpha$ trace using the \textsc{PULCHRA} algorithm \citep{rotkiewicz2008fast}, (2) relaxing all non-$C_\alpha$ backbone and side-chain atoms (keeping the $C_\alpha$ positions fixed), and (3) quantifying the potential energy of each of the resulting conformations using the \textsc{OpenMM} suite of molecular dynamics tools \citep{eastman2017openmm}, and the \textsc{AMBER} force field \citep{hornak2006comparison}. These free energy estimates were then used to approximate the Boltzmann distribution over conformational states
$$
\textstyle{
p_B(i) \propto \exp\left(-\frac{E_i}{k_BT}\right),}
$$
where temperature was set to $T=\SI{273.15}{\kelvin}$ and $k_B=\SI{1.380649e-23}{\joule\per\kelvin}$ is the Boltzmann constant. We then apply a very minor amount of smoothing to the resulting distribution by running forward Brownian motion on both the polytope and the torus for 10 steps, using a small step size of $\num{5e-3}$ and the respective metrics.
Finally, a subsample of $10^6$ $C_\alpha$ traces was drawn from this distribution and used for training and evaluating our models.

%% file: appendices/exp_details.tex
\section{Experimental details}
\label{sec:exp_details}

In what follows we describe the experimental settings used to generate results introduced in \cref{sec:experiments}.
~
The models and experiments have been implemented in Jax~\citep{jax2018github}, using a modified version of the Riemannian geometry library Geomstats~\citep{geomstats2020}.

\paragraph{Architecture.}
The architecture of the score network $\bm{s}_\theta$ is given by a multilayer
perceptron with $6$ hidden layers with $512$ units each.  We use sinusoidal activation functions.  

\paragraph{Training.}
All models are trained by the stochastic optimizer Adam \citep{kingma2014method}
with parameters $\beta_1=0.9$, $\beta_2=0.999$, batch-size of $256$ data-points.
The learning rate is annealed with a linear ramp from $0$ to $1000$ steps, reaching the maximum value of $2e-4$, and from then with a cosine schedule down to $0$ after $100k$ iterations in total.

\paragraph{Diffusion.}
Following \citet{song2020score}, the diffusion models
diffusion coefficient is parametrized as $g(t) = \sqrt{\beta(t)}$ with
$\beta: \ t \mapsto \beta_{\min} + (\beta_{\max} - \beta_{\min}) \cdot t$, 
where we found $\beta_{\min}=0.001$ and $\beta_{\max}=6$ to work best.

\paragraph{Metrics.}
We measure the performance of trained models via the Maximum Mean Discrepancy (MMD) \citep{gretton2012kernel}, which is a kernel based metric between two distributions $P$ and $Q$. 
The MMD can be empirically approximated with the following U-statistics
$\mathrm{MMD}^2(P, Q) = \tfrac{1}{m(m-1)} \sum_i \sum_{j \neq i} k(x_i, x_j) + \tfrac{1}{m(m-1)} \sum_i \sum_{j \neq i} k(y_i, y_j) - 2 \tfrac{1}{m^2} \sum_i \sum_{j} k(x_i, y_j)$
with $x_i \sim P$ and $y_i \sim Q$, where $k$ is a kernel. For synthetic experiments we use a sum of weighted RBF kernels matching the generating distributions for the Gaussian mixtures. For the other experiments we use an RBF kernel. For all experiments we use 100,000 samples to compute the MMD.

\vfill
\pagebreak

\subsection{Synthetic data on polytopes}
\label{sec:app_synthetc_data}

\paragraph{Hypercube $[-1,1]^n$.}

The hypercube is a specific instance of a convex polytope where the affine constraints are given by the following coefficients:

\begin{equation}
    A = \begin{pmatrix}
	1 & \dots & 0\\
        -1 & \dots & 0\\
        \vdots  & \ddots & \vdots \\
        0 & \dots & 1\\
        0 & \dots & -1\\
\end{pmatrix}, \;\;
b = \begin{pmatrix}
	1\\1\\\vdots\\1\\1
\end{pmatrix}.
\end{equation}

Where $A$ is a $2n \times n$ matrix and $b$ an $n$ dimensional vector.

\begin{figure}[H]
\centering
\begin{subfigure}{0.30\textwidth}
  \centering
  \includegraphics[width=\linewidth]{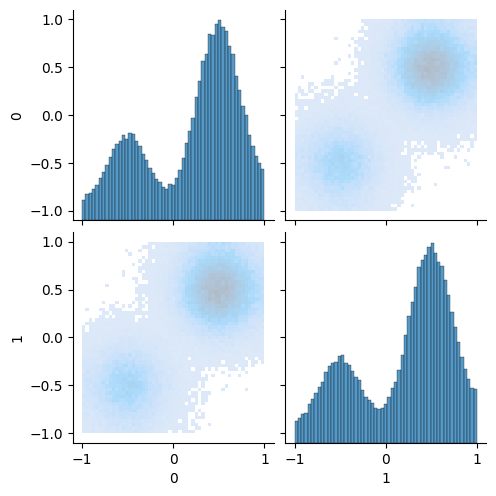}
  \caption{On the hypercube.}
  \label{fig:hypercube_data}
\end{subfigure}
\hspace{1.5cm}
\begin{subfigure}{0.30\textwidth}
  \centering
  \includegraphics[width=\linewidth]{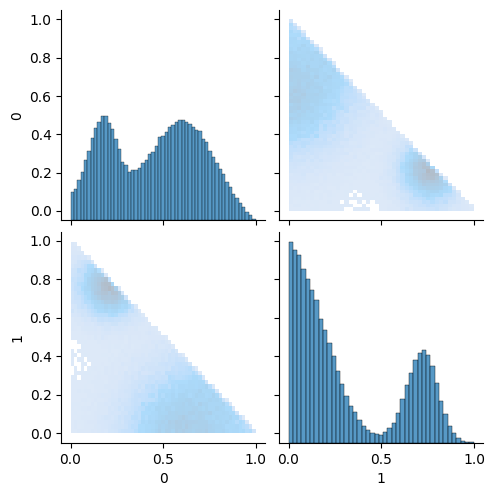}
  \caption{On the simplex.}
  \label{fig:simplex_data}
\end{subfigure}
\caption{Pairwise and marginals samples from the synthetic data distribution.}
\label{fig:synthetic_data}
\end{figure}

We construct the training and test datasets by sampling for both $100 000$ points from a mixture of `wrapped normal' distributions illustrated in \cref{fig:hypercube_data} and which density is given by
$$ p_0(x) = 0.7 ~\text{ReflectedStep}[(0.5,0.5), \cdot, \{f_i\}_{i \in \mathcal{I}}] \# \c{N}(0,0.25) + 0.3  ~\text{ReflectedStep}[(-0.5,-0.5), \cdot, \{f_i\}_{i \in \mathcal{I}}] \# \c{N}(0,0.25).$$

\paragraph{Simplex $\Delta^n$.}

Similarly, to parameterise the simplex as a convex polytope we set the matrix and constraints to be given by

\begin{equation}
    A = \begin{pmatrix}
	-1 & 0 & \dots & 0\\
        \vdots   & \vdots &  \ddots & \vdots \\
        0 & 0 & \dots & -1\\
          1 & 1 & 1& 1
\end{pmatrix}, \;\;
b = \begin{pmatrix}
	0\\\vdots\\0\\1
\end{pmatrix}.
\end{equation}

Where $A$ will be a $n-1 \times n$ matrix. Essentially we perform diffusion over the first $n-1$ components of the simplex, allowing the last component to be determined by the one minus the sum of the first $n-1$.

\begin{figure}[H]
\centering
\begin{subfigure}{0.32\textwidth}
  \centering
  \includegraphics[width=\linewidth]{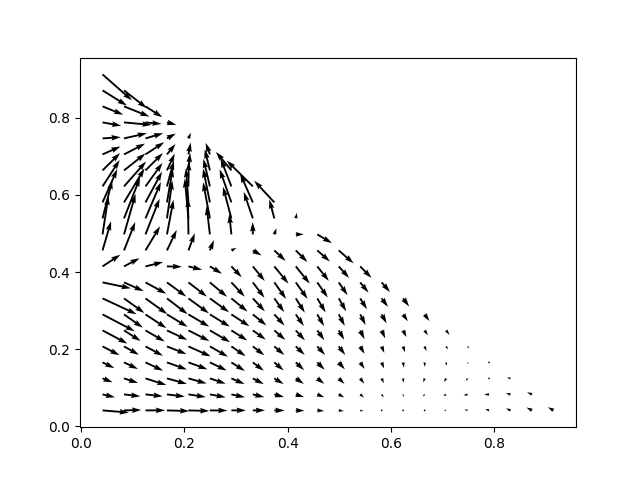}
  \caption{$t=0.01$.}
  \label{fig:001_score}
\end{subfigure}
\hfill
\begin{subfigure}{0.32\textwidth}
  \centering
  \includegraphics[width=\linewidth]{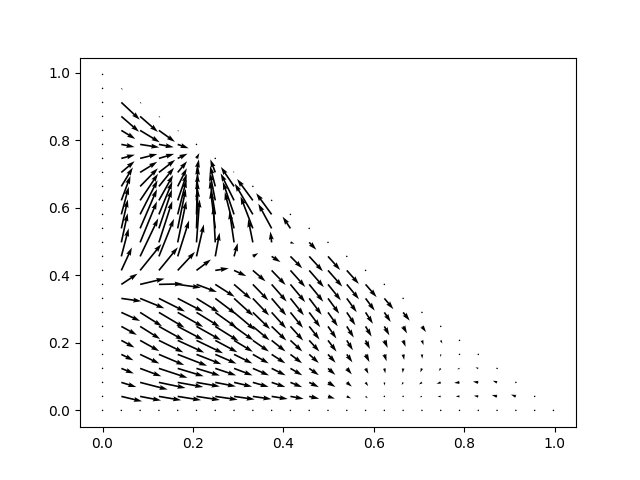}
  \caption{$t=0.5$.}
  \label{fig:05_score}
\end{subfigure}
\hfill
\begin{subfigure}{0.26\textwidth}
  \centering
  \includegraphics[width=\linewidth]{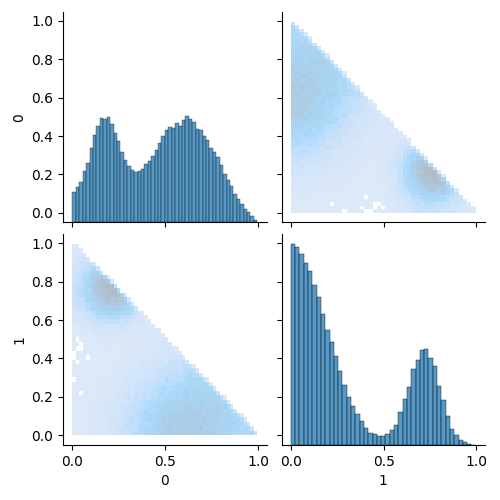}
  \caption{Generated distribution.}
  \label{fig:generated}
\end{subfigure}
\caption{Evolution of the score on the simplex and generated distribution.}
\label{fig:simplex_score}
\end{figure}

Similarly than for the hypercube, we construct the training and test datasets from generated data points which are illustrated in \cref{fig:simplex_data}. The score network at different times is illustrated in \Cref{fig:simplex_score}.

\paragraph{The Birkhoff polytope.}
The Birkhoff polytope is the space of doubly stochastic matrices, i.e.\ $B_n = \{ \mathrm{P} \in [0,1]^{n \times n} : \sum_i^n P_{i,j}=1, \sum_j^n P_{i,j}=1\}$.
It is a convex polytope in $\R^{n^2}$ and has dimension $d = (n-1)^2$.

\begin{figure}[H]
\centering
\begin{subfigure}{0.3\textwidth}
  \centering
  \includegraphics[width=\linewidth]{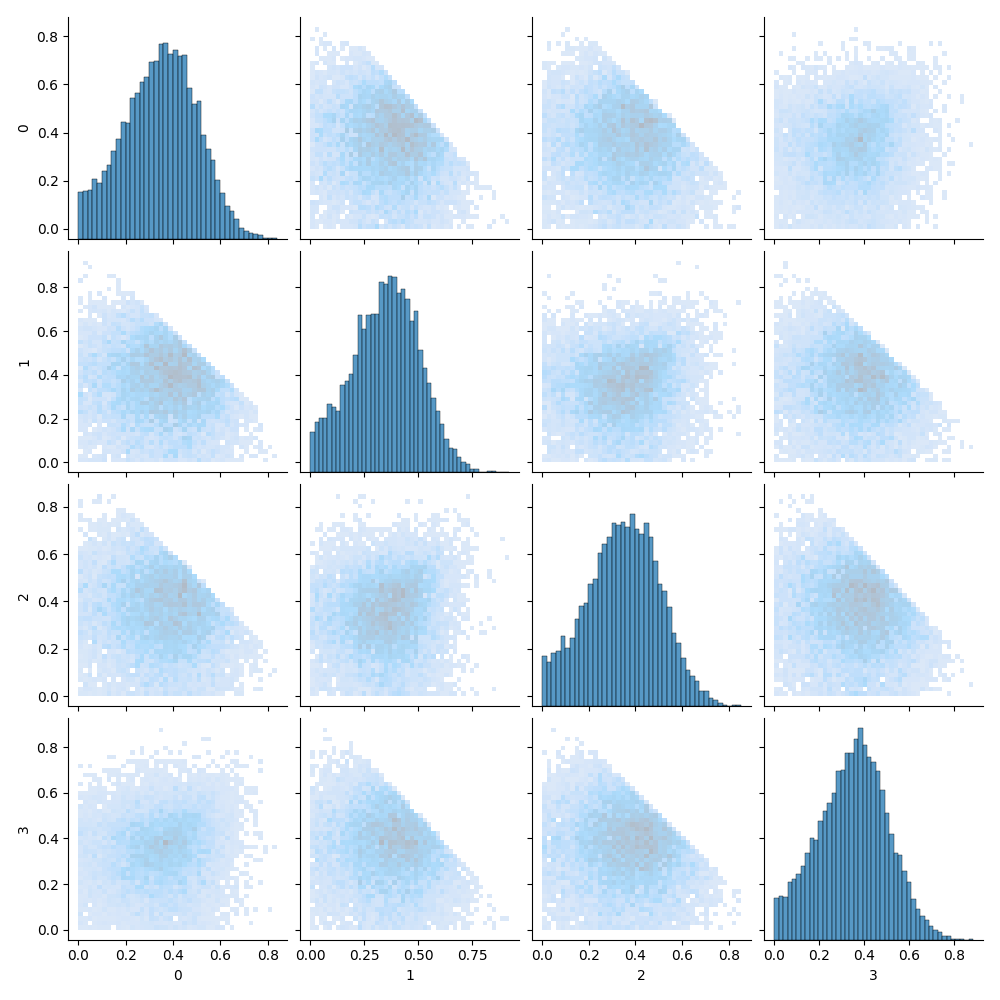}
  \caption{Data.}
  \label{fig:birkhoff_data}
\end{subfigure}
\hfill
\begin{subfigure}{0.3\textwidth}
  \centering
  \includegraphics[width=\linewidth]{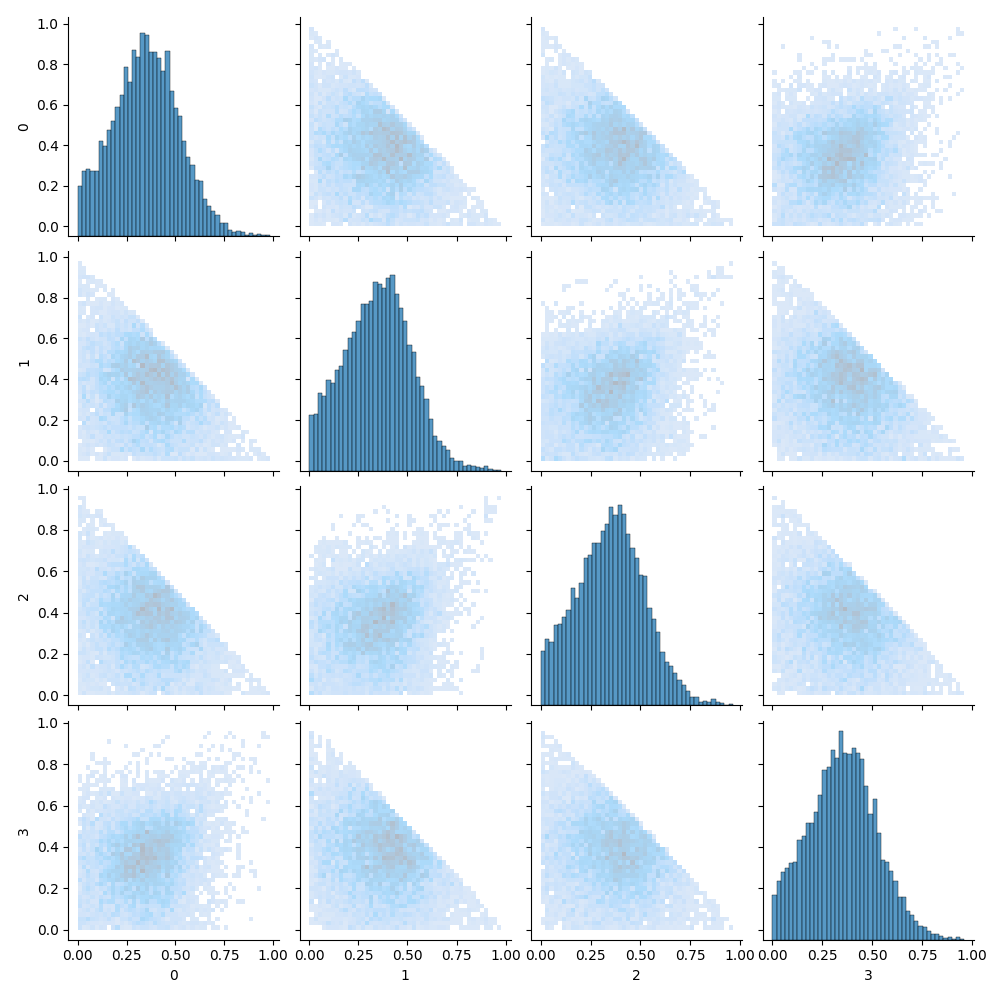}
  \caption{Log-barrier.}
  \label{fig:birkhoff_barrier}
\end{subfigure}
\hfill
\begin{subfigure}{0.3\textwidth}
  \centering
  \includegraphics[width=\linewidth]{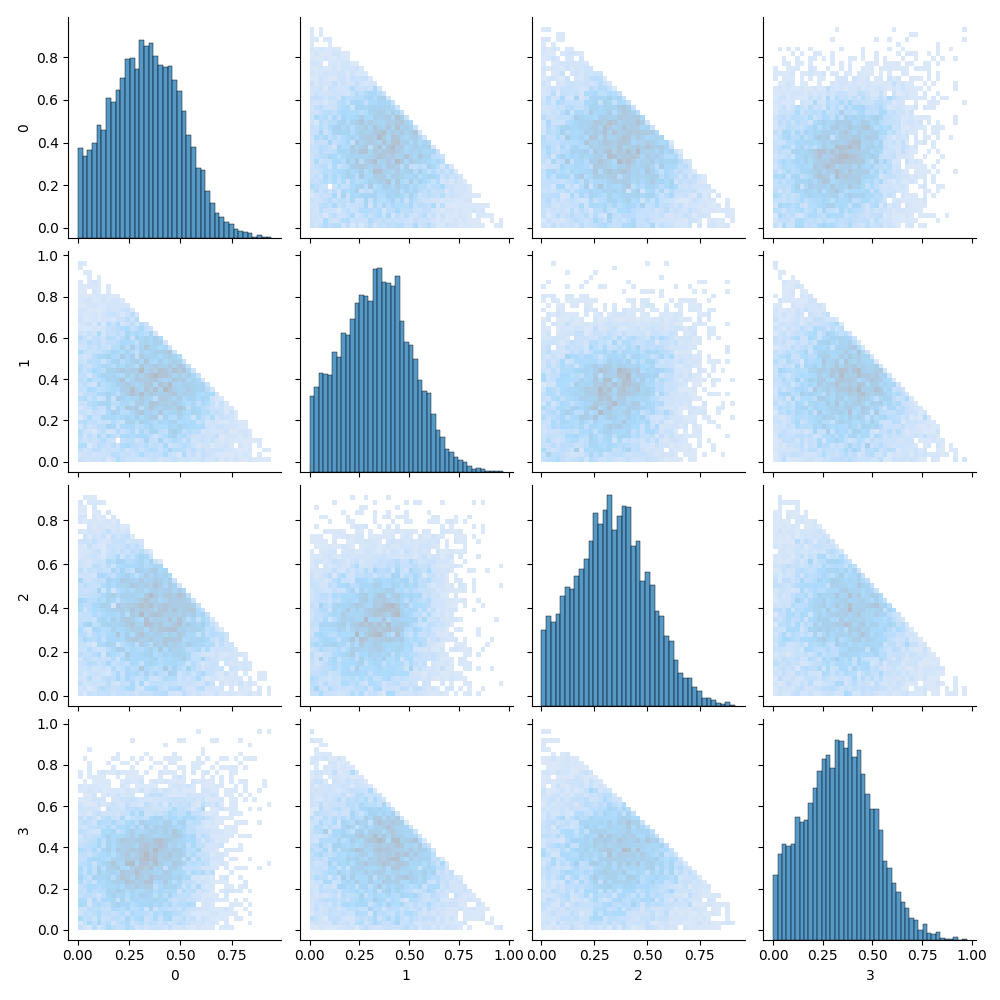}
  \caption{Reflected.}
  \label{fig:birkhoff_reflected}
\end{subfigure}
\caption{Pairwise and marginals samples on the Birkhoff polytope from synthetic data distribution and from trained constrained diffusion models.}
\label{fig:birkhoff}
\end{figure}

\vfill
\pagebreak

\subsection{Constrained SPD matrices for robotic arms modelling}
\label{sec:app_spd_exp}

\begin{figure}[H]
\centering
\begin{subfigure}{0.27\textwidth}
  \centering
  \includegraphics[width=\linewidth]{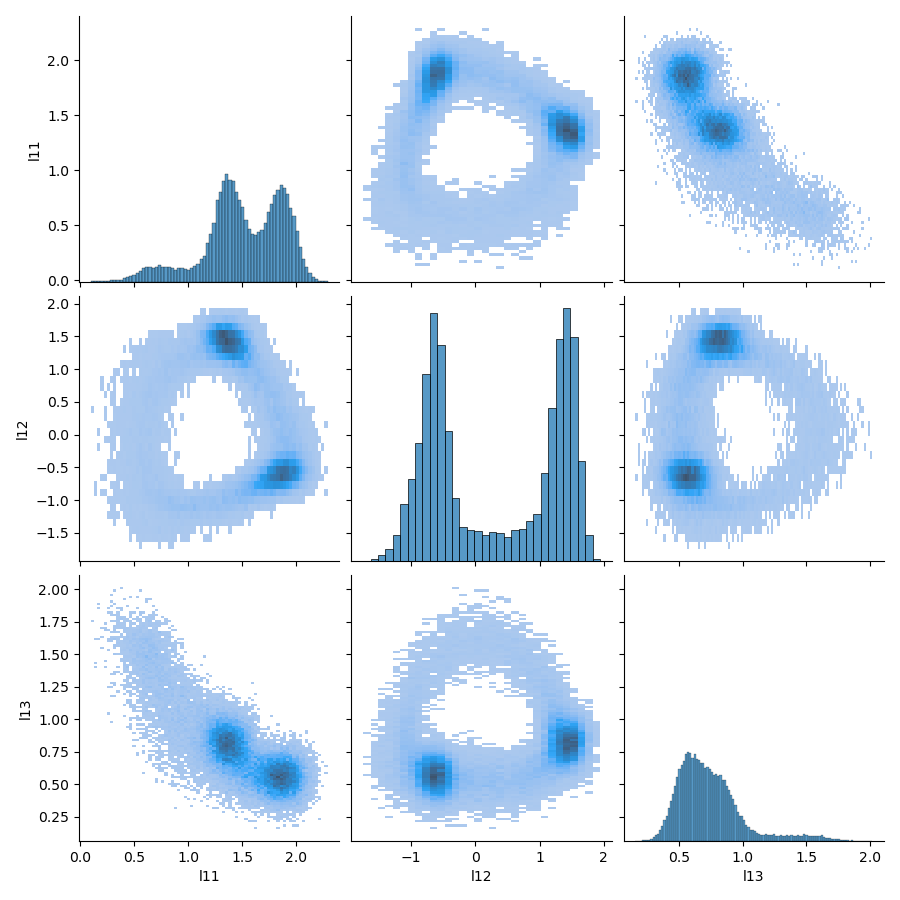}
  \caption{Data.}
  \label{fig:spd_ll_data}
\end{subfigure}
\hfill
\begin{subfigure}{0.27\textwidth}
  \centering
  \includegraphics[width=\linewidth]{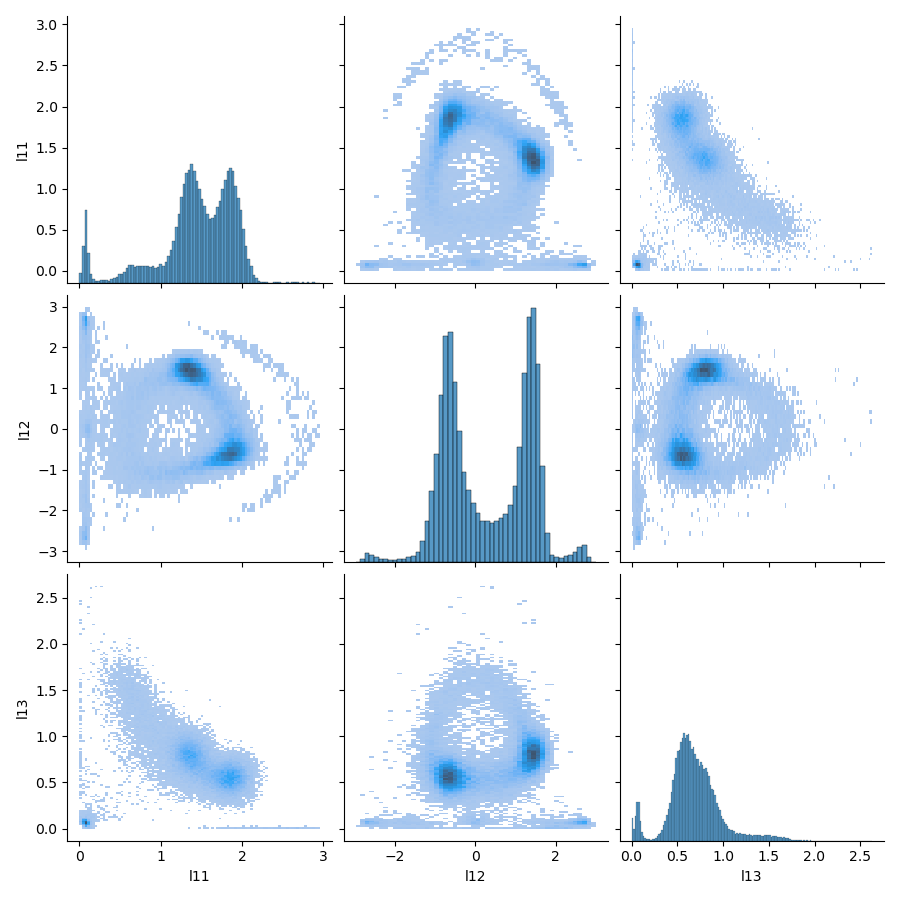}
  \caption{Log-barrier.}
  \label{fig:spd_ll_barrier}
\end{subfigure}
\hfill
\begin{subfigure}{0.27\textwidth}
  \centering
  \includegraphics[width=\linewidth]{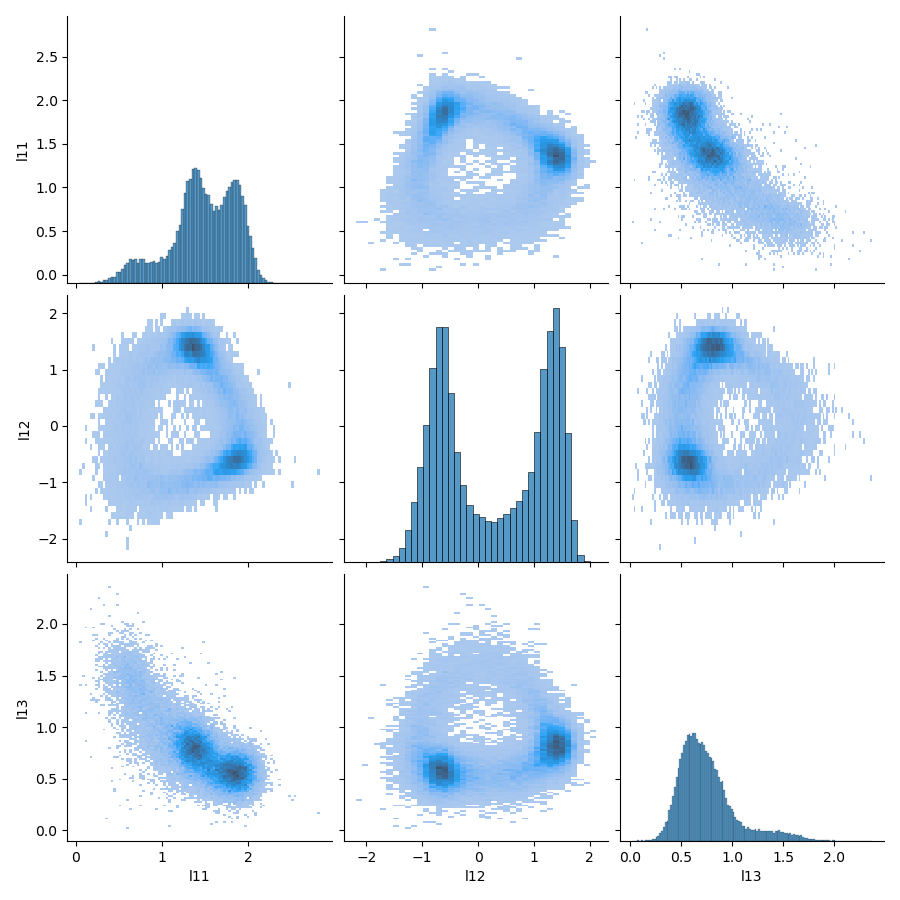}
  \caption{Reflected.}
  \label{fig:spd_ll_reflected}
\end{subfigure}
\caption{
Pairwise and marginals distributions over the coefficients $\mathrm{L}_{11}, \mathrm{L}_{21}, \mathrm{L}_{22}$ of the lower triangle matrix parameterising SPD matrices $\mathrm{M} = \mathrm{L} \mathrm{L}^\top$ (which represent the manipulability ellipsoids of the robotic arms).
}
\label{fig:spd}
\end{figure}

\begin{figure}[H]
\centering
\begin{subfigure}{0.27\textwidth}
  \centering
  \includegraphics[width=\linewidth]{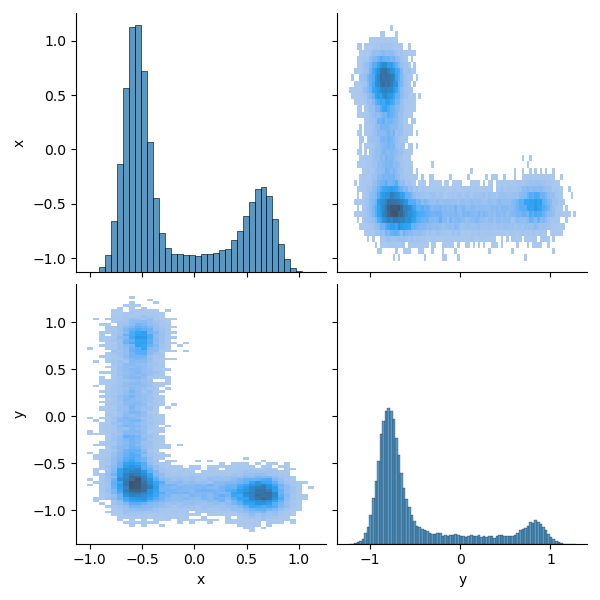}
  \caption{Data.}
  \label{fig:spd_xy_data}
\end{subfigure}
\hfill
\begin{subfigure}{0.27\textwidth}
  \centering
  \includegraphics[width=\linewidth]{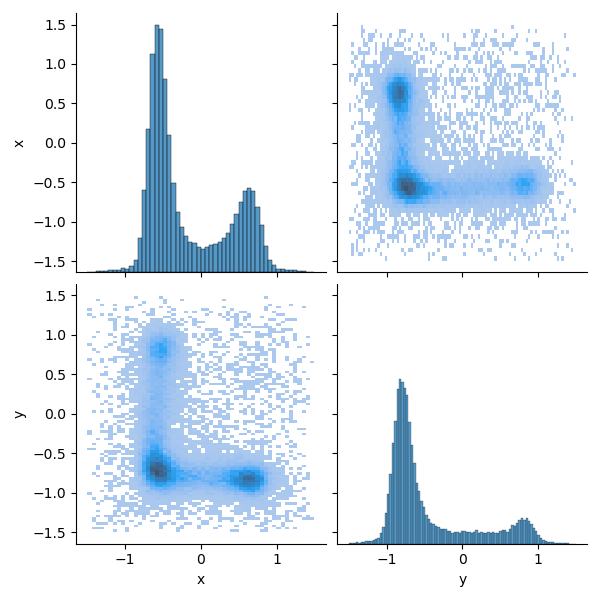}
  \caption{Log-barrier.}
  \label{fig:spd_xy_barrier}
\end{subfigure}
\hfill
\begin{subfigure}{0.27\textwidth}
  \centering
  \includegraphics[width=\linewidth]{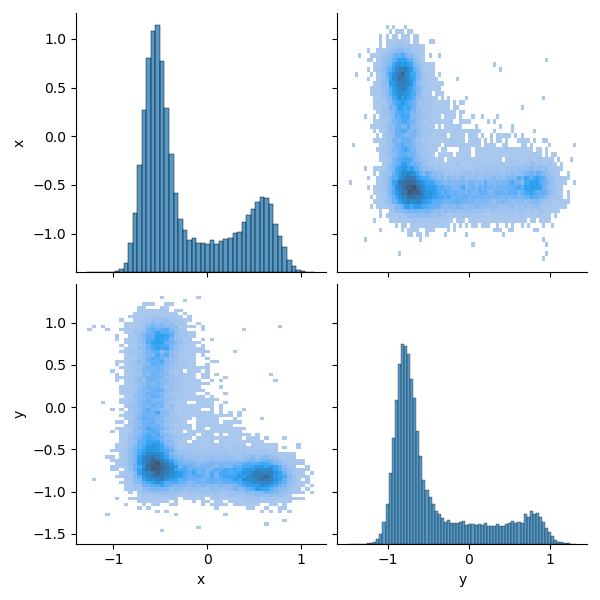}
  \caption{Reflected.}
  \label{fig:spd_xy_reflected}
\end{subfigure}
\caption{
Pairwise and marginals distributions over the $(x,y)$ locations of the robotic arms.
}
\label{fig:spd}
\end{figure}

\subsection{Conformational modelling of polypeptide backbones under anchor point constraints}
\label{sec:app_protein_exp}

\begin{figure}[h]
\centering
\begin{subfigure}{0.3\textwidth}
  \centering
  \includegraphics[width=\linewidth]{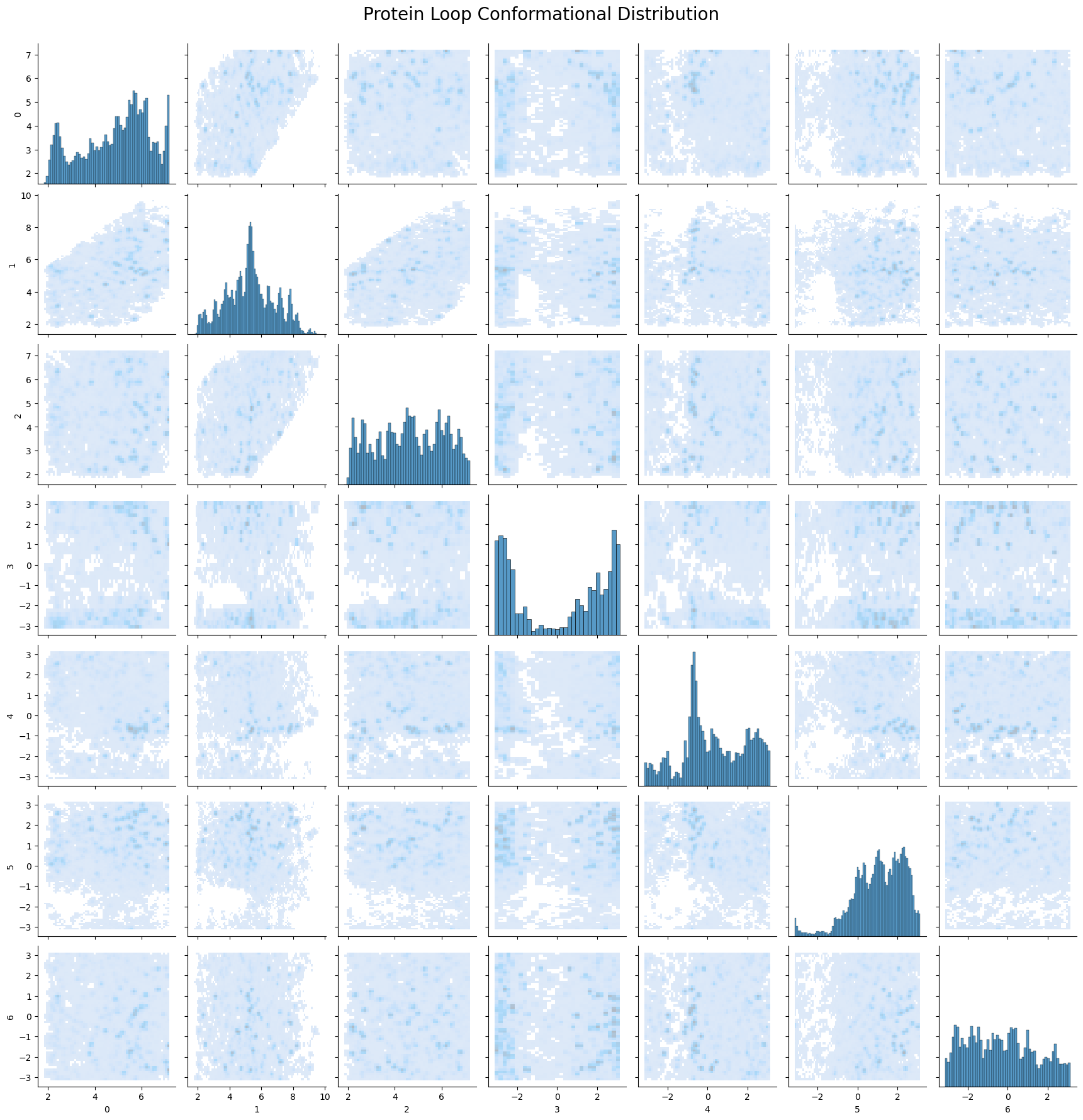}
  \caption{Data.}
\end{subfigure}
\hfill
\begin{subfigure}{0.3\textwidth}
  \centering
  \includegraphics[width=\linewidth]{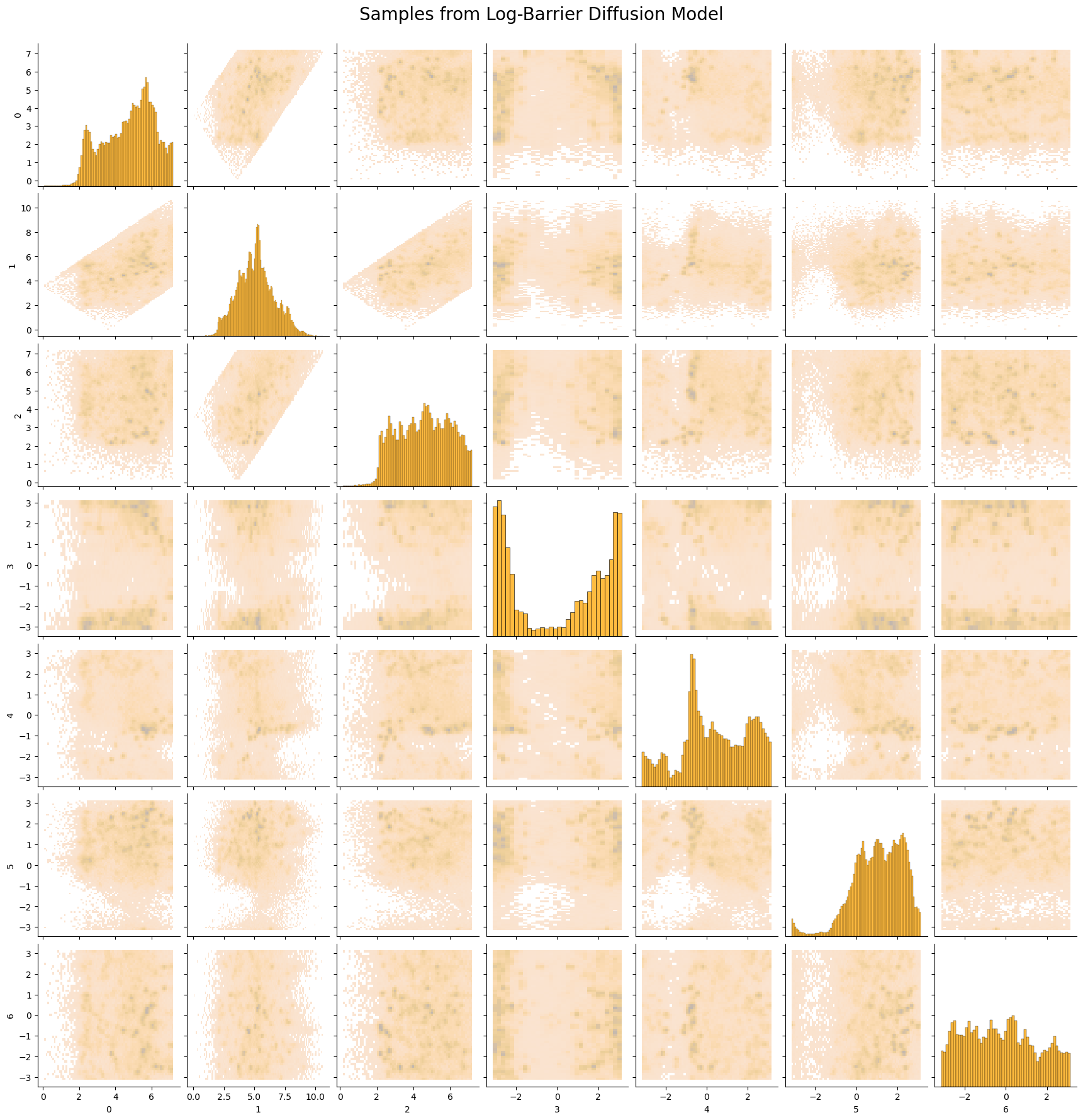}
  \caption{Log-barrier.}
  \label{fig:loops_log}
\end{subfigure}
\hfill
\begin{subfigure}{0.3\textwidth}
  \centering
  \includegraphics[width=\linewidth]{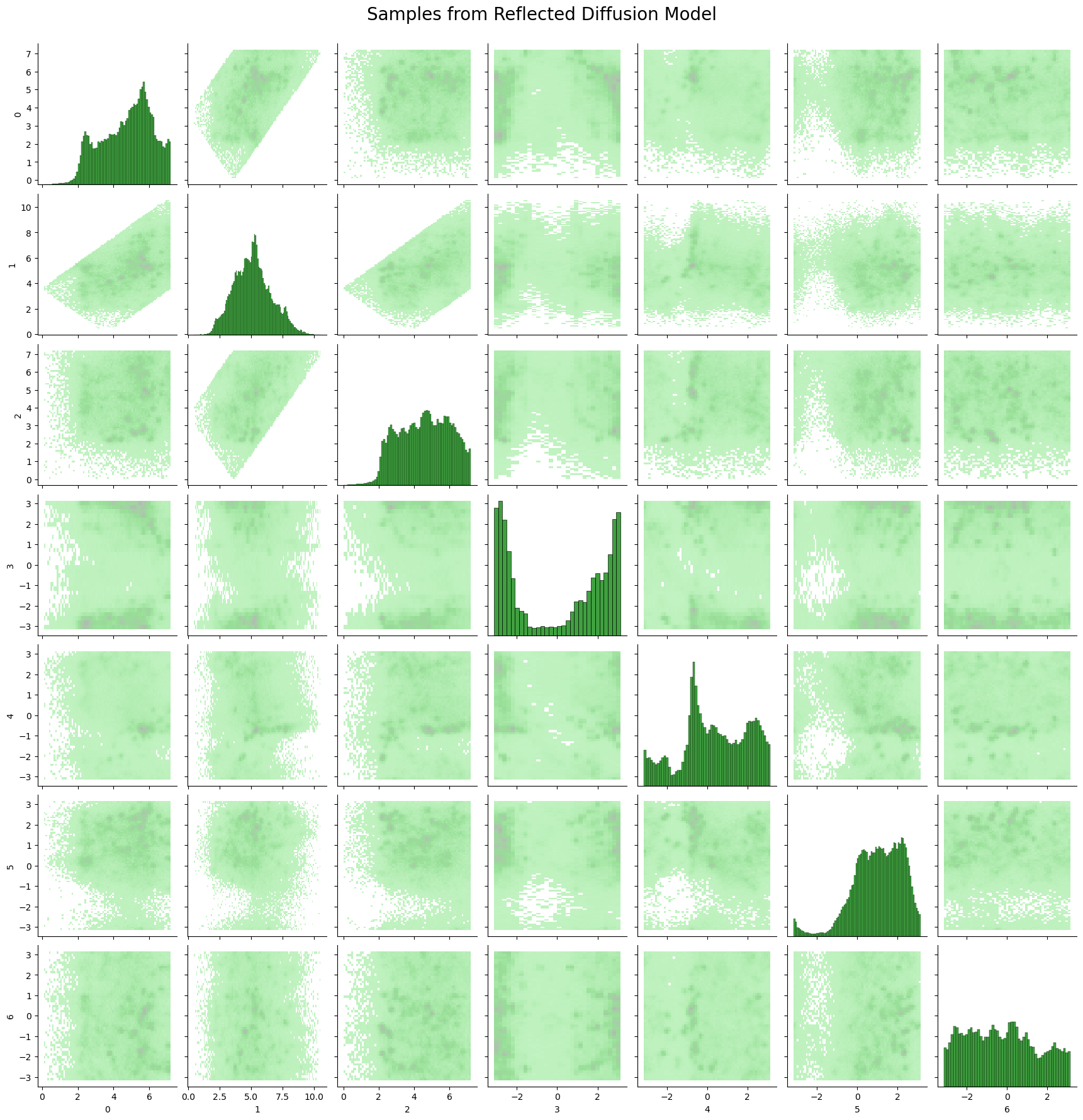}
  \caption{Reflected.}
  \label{fig:loops_reflected}
\end{subfigure}
\caption{
Pairwise and marginals distributions over the dimensions of the polytope and torus used to model the conformational ensembles of cyclic peptides generated by the reflected diffusion model.
}
\label{fig:loops_pairs}
\end{figure}

\pagebreak